\documentclass[10pt,twocolumn,letterpaper]{article}
\usepackage{cvpr} 

\usepackage[accsupp]{axessibility}
\usepackage{comment}
\usepackage[dvipsnames]{xcolor}

\renewcommand{\paragraph}[1]{\vspace{2pt}\noindent\textbf{#1}\;} 
\usepackage[skip=1pt plus2pt, indent=11pt]{parskip}
\usepackage{tabularx}
\usepackage{makecell}
\usepackage{multirow}
\usepackage{booktabs}
\usepackage{floatpag} 
\usepackage{pifont} 

\def\bk{{\boldsymbol{k}}}
\def\bx{{\boldsymbol{x}}}
\def\bX{{\mathbf{X}}} 

\def\bY{{\mathbf{Y}}} 
\def\bK{{\mathbf{K}}} 
\def\bb{{\boldsymbol{b}}}
\def\bW{{\boldsymbol{W}}}
\def\btheta{{\boldsymbol{\theta}}}
\makeatletter
\newcommand{\shorteq}{\mathrel{\mkern0.2mu\mathpalette\shorteq@\relax\mkern0.2mu}}
\newcommand{\shorteq@}[2]{\scalebox{0.5}[1]{$\m@th#1=$}}
\newcommand{\shortteq}{\!\shorteq\!}
\makeatother
\usepackage{bbold} 
\usepackage{nicefrac}
\newcommand\lldots{\ifmmode$\lldots$\else\thinspace\makebox[1em][c]{.\hfil.\hfil.}\fi} 

\usepackage[framemethod=TikZ]{mdframed}
\definecolor{lightGray}{gray}{0.97}
\definecolor{midGray}{gray}{0.40}
\definecolor{lightYellow}{RGB}{254,254,233}
\mdfdefinestyle{citationFrame}{%
  linecolor=midGray,
  linewidth=0.5pt,
  roundcorner=3pt,
  skipabove=5pt,
  skipbelow=0pt,
  innertopmargin=5pt,
  innerbottommargin=5pt,
  innerrightmargin=7pt,
  innerleftmargin=7pt,
  leftmargin=0pt,
  rightmargin=0pt,
  backgroundcolor=lightYellow
}

\definecolor{cvprblue}{rgb}{0.21,0.49,0.74}
\usepackage[pagebackref,breaklinks,colorlinks,citecolor=cvprblue]{hyperref}

\def\paperID{5942}
\def\confName{CVPR}
\def\confYear{2024}
\title{Neural Redshift: Random Networks are not Random Functions}

\author{%
    Damien Teney\\
    Idiap Research Institute\\
    {\tt\small damien.teney@idiap.ch}
    \and
    \hspace{20pt}Armand Mihai Nicolicioiu\\
    \hspace{20pt}ETH Zurich\\
    {\hspace{20pt}\tt\small armandmihai.nicolicioiu@inf.ethz.ch}
    \and
    \hspace{-23pt}Valentin Hartmann\\
    \hspace{-23pt}EPFL\\
    {\tt\small valentin.hartmann@epfl.ch\hspace{23pt}}
    \and
    \hspace{-23pt}Ehsan Abbasnejad\\
    \hspace{-23pt}University of Adelaide\\
    {\tt\small ehsan.abbasnejad@adelaide.edu\hspace{23pt}}
}

\begin{document}
\maketitle
\begin{abstract}
Our understanding of the generalization capabilities of neural networks (NNs) is still incomplete.
Prevailing explanations are based on implicit biases of gradient descent~(GD)
but they cannot account for
the capabilities of models from gradient-free methods
~\cite{chiang2022loss}
nor the simplicity bias recently observed in \emph{untrained} networks~\cite{goldblum2023no}.
This paper seeks other sources of generalization in NNs.

\vspace{1pt}
\noindent\textbf{Findings.}
To understand the inductive biases provided by architectures independently from GD, we examine untrained, random-weight networks.
Even simple MLPs show strong inductive biases:
uniform sampling in weight space yields a very biased distribution of functions in terms of complexity.
But unlike common wisdom, NNs do not have an inherent ``simplicity bias''.
This property depends on components such as ReLUs, residual connections, and layer normalizations.
Alternative architectures can be built with a bias for any level of complexity.
Transformers also inherit all these properties from their building blocks.

\vspace{1pt}
\noindent\textbf{Implications.}
We provide a fresh explanation for the success of deep learning independent from gradient-based training.
It points at promising avenues for controlling the solutions implemented by trained models.
\end{abstract}

\vspace{-8pt}    
\section{Introduction}
\label{sec:intro}

Among various models in machine learning,
neural networks (NNs) are the most successful
on a variety of tasks.
While we are pushing their capabilities with ever-larger models~\cite{sutton2019bitter},
much remains to be understood
at the level of their building blocks.
This work seeks to understand
what provides
NNs with their unique generalization capabilities.

\begin{figure}[t]
  \renewcommand{\tabcolsep}{0.35em}
  \renewcommand{\arraystretch}{1}
  \begin{tabular}{ccccc}
    &\scriptsize\textsf{ReLU} & \scriptsize\textsf{GELU} & \scriptsize\textsf{TanH} & \scriptsize\textsf{Gaussian}\\[1pt]
    \raisebox{1pt}{\rotatebox{90}{\scriptsize\textsf{$\longleftarrow$ Weights magnitude}}}&
    \includegraphics[clip, trim=15pt 15pt 0 1pt, width=.2\linewidth]{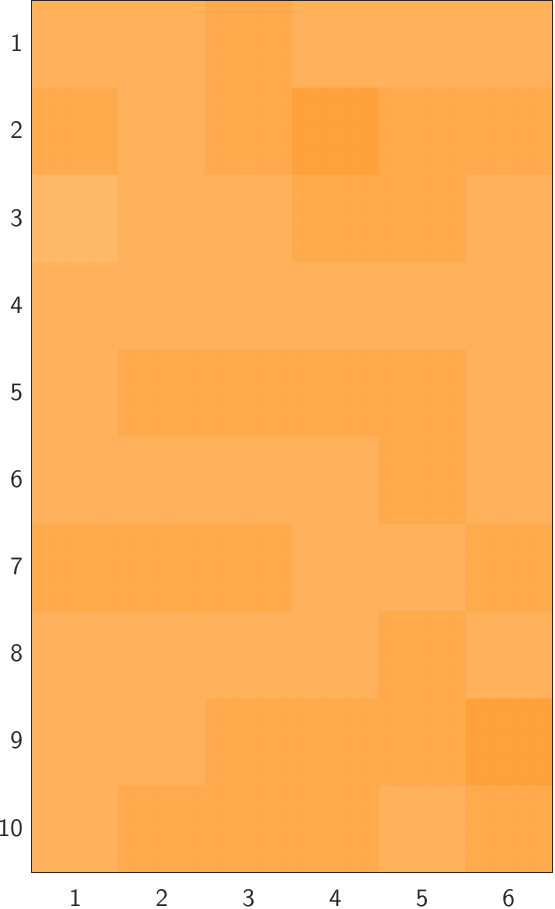}&
    \includegraphics[clip, trim=15pt 15pt 0 1pt, width=.2\linewidth]{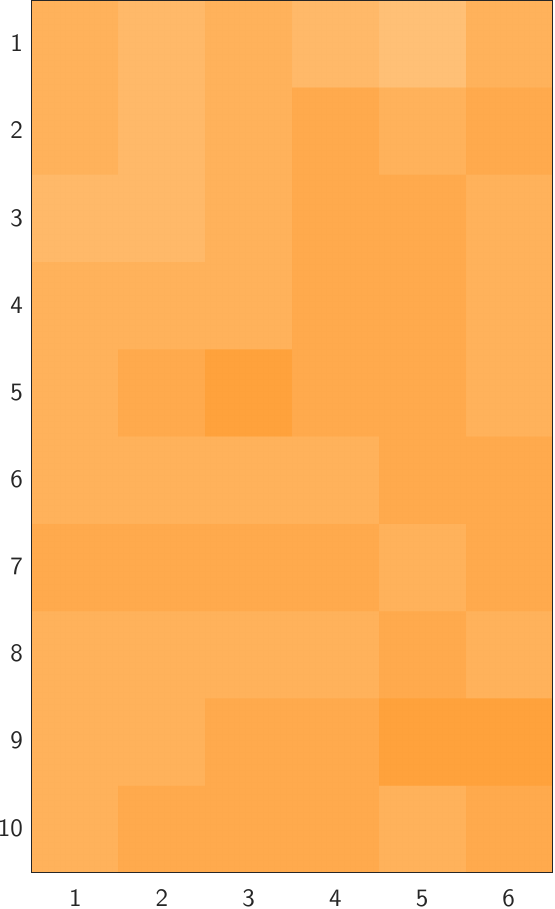}&
    \includegraphics[clip, trim=15pt 15pt 0 1pt, width=.2\linewidth]{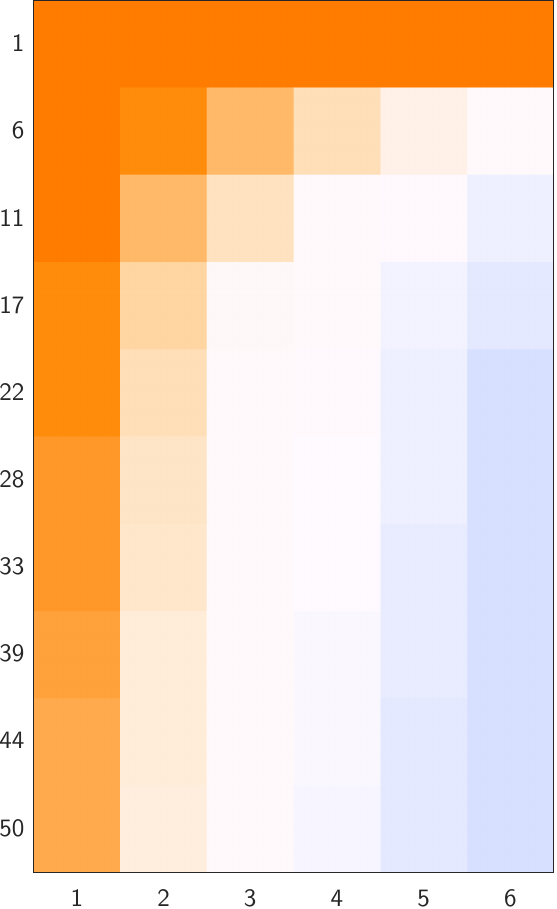}&
    \includegraphics[clip, trim=15pt 15pt 0 1, width=.2\linewidth]{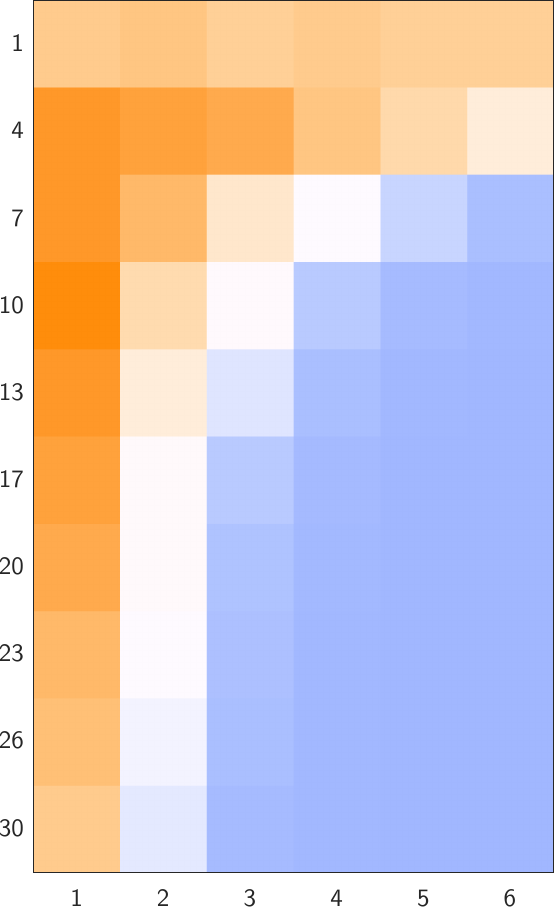}\\[-1pt]
    \multicolumn{5}{l}{\hspace{20pt}\scriptsize\textsf{Depth $\longrightarrow$}}\\[2pt]
    \multicolumn{5}{c}{\scriptsize\textbf{\textsf{Lower frequencies\raisebox{-.3\height}{\includegraphics[clip, trim=17pt 10pt 17pt 1pt, width=.4\linewidth]{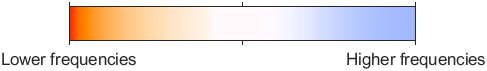}}Higher frequencies}}}
  \end{tabular}\vspace{-4pt}
  \caption{\label{figTeaser}
  We examine the complexity of the functions implemented by various MLP architectures.
  We find that much of their generalization capabilities can be understood independently from the optimization, training objective, scaling, or even data distribution.
  For example, ReLU and GELU networks~(left) overwhelmingly represent low-frequency functions
  for any \textbf{network depth} or
  \textbf{weight magnitude}.
  Other activations
  lack this property.
  \vspace{-8pt}}
\end{figure}

\paragraph{The need for inductive biases.}
The success of ML depends on using suitable inductive biases%
\footnote{\textbf{Inductive biases} are assumptions about the target function encoded in
the learning algorithm
as
the hypothesis class (\eg architecture), optimization method (\eg SGD), objective (\eg cross-entropy risk), regularizer, etc.}%
\,\cite{mitchell1980need}.
They specify how to generalize from a finite set of training examples to novel test cases.
For example, linear models allow learning from few examples but generalize correctly only
when the target function is itself linear.
Large NNs are surprising in having large representational capacity~\cite{hornik1989multilayer}
yet generalizing well across many tasks.
In other words, among all learnable functions that fit the training data,
those implemented by trained networks are often similar to the target function.

\paragraph{Explaining the success of neural networks.}
Some successful architectures are tailored to specific domains \eg CNNs
for image recognition.
But even the simplest MLP architectures (multi-layer perceptrons)
often display remarkable generalization.
Two explanations for this success prevail, although they are increasingly challenged.
\begin{itemize}[itemsep=4pt]
    \item
    \textbf{Implicit biases of gradient-based optimization}.
    A large amount of work studies the preference of (stochastic) gradient descent or (S)GD for particular solutions~\cite{frei2022implicit,pesme2021implicit}.
    But conflicting results have also appeared.
    First, full-batch GD can be as effective as SGD~\cite{geiping2021stochastic,huh2021low,mohtashami2023special,smith2021origin}.
    Second, \citet{chiang2022loss} showed that zeroth-order optimization can yield models with good generalization as frequently as GD.
    And third, \citet{goldblum2023no} showed that language models with random weights already display a preference for low-complexity sequences.
    This ``simplicity bias'' was previously thought to emerge from training~\cite{shah2020pitfalls}.
    In short, gradient descent may help with generalization but it does not seem necessary.
    
    \item
    \textbf{Good generalization as a fundamental property of all nonlinear architectures}~\cite{hermann2023foundations}.
    This vague conjecture does not account for the selection bias
    in the architectures and algorithms that researchers have converged on.
    For example, implicit neural representations (\ie a network trained to represent a specific image or 3D shape)
    show that the success of NNs is not automatic
    and requires, in that case, activations very different from ReLUs.
\end{itemize}
The success of deep learning is thus not a product primarily of GD, nor is it universal to all architectures.
This paper propose an explanation compatible with all above observations.
It builds on the growing evidence that NNs benefit from
their parametrization and the structure of their weight space~\cite{chiang2022loss,goldblum2023no,huang2020understanding,scimeca2021shortcut,theisen2021good,valle2018deep}.

\paragraph{Contributions.} 
We present experiments supporting this
three-part proposition (stated formally in Appendix~\ref{app:formal}).

\begin{mdframed}[style=citationFrame,userdefinedwidth=
\linewidth,align=center,skipabove=5pt,skipbelow=0pt]
\textbf{(1)}~NNs are biased to implement functions of a particular level of complexity (not necessarily low) determined by the architecture.
\textbf{(2)}~This preferred complexity is observable in networks
with random weights from an uninformed prior.
\textbf{(3)}~Generalization is enabled by
popular components like ReLUs setting this bias to a low complexity that often aligns with the target function.
\end{mdframed}
We name it the \textbf{Neural Redshift (NRS)} by analogy to physical effects%
\footnote{\url{https://en.wikipedia.org/wiki/Redshift}}
that bias the observations of a signal towards low frequencies.
Here, the parameter space of NNs is biased towards functions of low frequency,
one of the measures of complexity used in this work (see Figure~\ref{figTeaser}).

The NRS differs from prior work on the spectral bias~\cite{rahaman2019spectral,xu2019frequency} and simplicity bias~\cite{arpit2017closer,shah2020pitfalls} which confound the effects of architectures and gradient descent.
The spectral bias refers to the earlier learning of low-frequencies
\emph{during training} (see discussion in Section~\ref{sec:relatedWork}).
The NRS only involves the parametrization%
\footnote{
\textbf{Parametrization} refers to the mapping between a network's weights
and the function it represents.
An analogy in geometry is the parametrization of 2D points with Euclidean or polar coordinates.
Sampling uniformly from one or the other gives different distributions of points.}
of NNs
and thus shows interesting properties independent from 
optimization~\cite{vardi2023implicit}, scaling~\cite{bachmann2023scaling}, learning objectives~\cite{chiang2022loss}, or even data distributions~\cite{pezeshki2021gradient}.

Concretely, we examine various architectures 
with random weights.
We use three measures of complexity:
(1)~decompositions in Fourier series and (2)~in bases of orthogonal polynomials
(equating simplicity with low frequencies/order)
and
(3)~compressibility as an approximation of the Kolmogorov complexity~\cite{dingle2018input}.
We study how they vary across architectures
and how these properties at initialization correlate
with the performance of trained networks.

\paragraph{Summary of findings.}
\begin{itemize}[itemsep=2pt,topsep=1pt]
\item We verify the NRS with three notions
of complexity that rely on frequencies in Fourier decompositions, order in polynomial decompositions, and compressibility of the input-output mapping
(Section~\ref{sec:heatmaps}).

\item We visualize the input-output mappings of 2D networks
(Figure~\ref{figFctMapsCloseUp}).
They show intuitively the diversity of inductive biases
across architectures
that a scalar ``complexity'' cannot fully describe.
Therefore, matching the complexity of an architecture with the target function is beneficial
for generalization but hardly sufficient
(Section~\ref{sec:difficultTasks}).

\item We show that the simplicity bias is not universal but depends on common components (ReLUs, residual connections, layer normalizations).
ReLU networks also have the unique property of maintaining
their simplicity bias
for any depth and weight magnitudes.
It suggests that the historical importance of ReLUs in the development of deep learning goes beyond the common narrative about vanishing gradients~\cite{maas2013rectifier}.

\item We construct architectures where the NRS can be modulated
(with alternative activations and weight magnitudes) or entirely avoided
(parametrization in Fourier space, Section~\ref{sec:unbiasedLearner}). 
It further demonstrates that the simplicity bias is not universal 
and can be controlled to learn complex functions (\eg modulo arithmetic) or mitigate shortcut learning (Section~\ref{sec:difficultTasks}).

\item We show that the NRS is relevant to transformer sequence models.
Random-weight transformers produce sequences of low complexity
and this can also be modulated with the architecture.
This suggests that transformers inherit inductive biases from their
building blocks
via mechanisms similar to those of simple models.
(Section~\ref{sec:transformers}).
\end{itemize}

\begin{figure*}[tb]
  \centering
  \vspace{-2pt}
  \includegraphics[width=.88\linewidth]{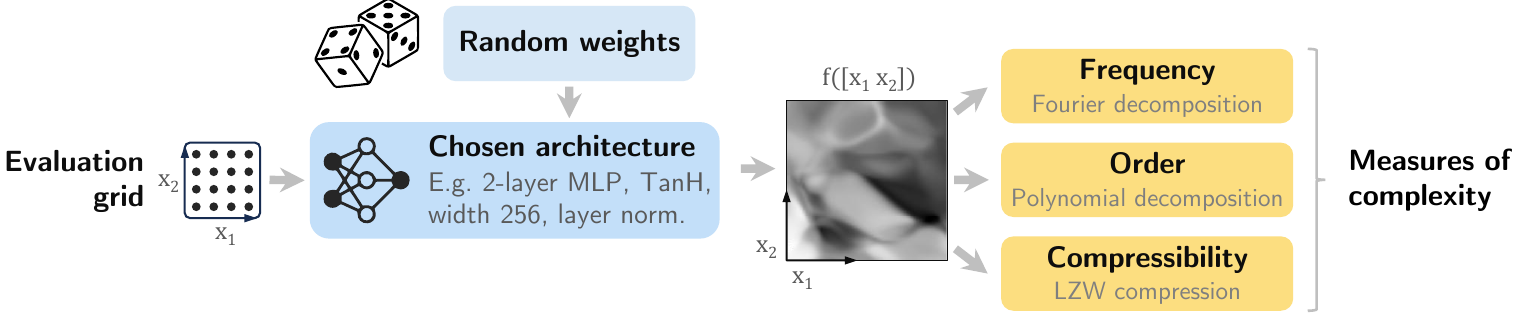}
  \vspace{-2pt}
  \caption{\label{figMethod}
  Our methodology to characterize the inductive biases of an architecture.
  We evaluate a network with random weights/biases
  on a grid of points.
  This yields a representation of the function implemented by the network, shown  here as a grayscale image for a 2D input. 
  We then characterize this function using three measures of complexity.
  \vspace{-8pt}}
\end{figure*}

\section{How to Measure Inductive Biases?}

Our goal is to understand why NNs generalize
when other models of similar capacity would often overfit.
The immediate answer is simple: \textbf{NNs have an inductive bias for functions with properties matching real-world data}. Hence two subquestions.
\begin{enumerate}[itemsep=2pt,topsep=-1pt]
\item \textbf{What are these properties?}\\
    We will show that three metrics are relevant:
    low frequency, low order, and compressibility. Hereafter, they are collectively referred to as ``simplicity''.
\item \textbf{What gives neural networks these properties?}\\
    We will show that an overwhelming fraction of their parameter space
    corresponds to functions with such simplicity.
    While there exist solutions of arbitrary complexity,
    simple ones are found by default when navigating this space
    (especially with gradient-based methods).
\end{enumerate}

\paragraph{Analyzing random networks.}
Given an architecture
to characterize,
we propose to sample random weights and biases,
then evaluate the network on a regular grid in its input space
(see Figure~\ref{figMethod}).
Let 
$f_\btheta(\bx)$
represent the function implemented by a network 
of parameters $\btheta$ (weights and biases),
evaluated on the input $\bx \in \mathbb{R}^{d}$.
The $f$ represents an architecture with a scalar output and no output activation,
which could serve for any regression or classification task.

We sample weights and biases from an uninformed prior chosen as
a uniform distribution.
Biases are sampled from $\mathcal{U}(-1,1)$,
and weights from the range proposed by~\citet{glorot2010understanding}
commonly used for initialization
\ie $\mathcal{U}(-s,s)$ with $s=\alpha \, \sqrt{6 \,/\, (\textrm{fan}_\mathrm{in} + \textrm{fan}_\mathrm{out})}$
where $\alpha$ is an extra factor ($1$ by default) to manipulate the weights' magnitude in our experiments.
These choices
are not critical.
Appendix~\ref{app:otherWeightDistributions} shows that
other distributions
(Gaussian, uniform-ball, long-tailed)
lead to similar findings.

We then evaluate
$f_\btheta(\cdot)$
on points $\bX_\mathrm{grid}\!=\!\{\bx_1, \ldots, \bx_N\}$ sampled regularly in the input space.
We restrict $\bX_\mathrm{grid}$ to the hypercube
$[-1,+1]^d$
since the data used with NNs is commonly normalized.
The evaluation of $f_\btheta(\cdot)$
on $\bX_\mathrm{grid}$
yields a $d$-dimensional grid of scalars.
In experiments of Section~\ref{sec:untrained} where $d\!\!=\!\!2$,
this is conveniently visualized as a 2D grayscale image
to provide visual intuition about the function represented by the network.
Next, we describe three quantifiable properties to extract from such representations.

\paragraph{Measures of complexity.}
\label{sec:complexityMeasures}
Applying the above procedure to various architectures with
2D inputs
yields clearly diverse visual representations
(see Figure~\ref{figFctMapsCloseUp}).
For quantitative comparisons,
we propose three functions of the form $\operatorname{C}(\bX_\mathrm{grid}, f)$ that
estimate proxies of the complexity of $f$.

\begin{itemize}[itemsep=3pt,topsep=0pt]
\item \textbf{Fourier frequency.}
A first natural choice is to use Fourier analysis
as in classical signal and image processing.
The ``image'' to analyze is the $d$-dimensional evaluation of $f$ on $\bX_\mathrm{grid}$ mentioned above.
We first compute a discrete Fourier transform
that approximates $f$ with a weighted sum of sines of various frequencies.
Formally, we have
$f(x) := (2\pi)^{\nicefrac{d}{2}} \int \tilde{f}(\bk) \, e^{i\bk\cdot \bx} d\bk$
where
$\tilde{f}(\bk) := \int f(\bx)\, e^{- i \bk\cdot \bx} \mathbf{dx}$ is the Fourier transform.
The \emph{discrete} transform means that the frequency numbers $\bk$
are regularly spaced, $\{0,1,2,\ldots,K\}$ with the maximum $K$ set according to 
the Nyquist–Shannon limit of $\bX_\mathrm{grid}$.
We use the intuition that complex functions are those with large high-frequency coefficients~\cite{rahaman2019spectral}.
Therefore, we define our measure of complexity as the average of the coefficients weighted by their corresponding frequency
\ie $\operatorname{C}_\text{Fourier}(f) ~=~ \Sigma_{k=1}^K \tilde{f}(\bk) \, k ~~/~~ \Sigma_{k=1}^K\tilde{f}(\bk)$.

For example, a smoothly varying function is likely to involve mostly low-frequency components,
and therefore give a low value to $\operatorname{C}_\text{Fourier}$.

\item \textbf{Polynomial order.}
A minor variation of Fourier analysis
uses decompositions in bases of polynomials.
The procedure is nearly identical,
except for the
sine waves of increasing frequencies
being replaced with
fixed polynomials of increasing order (details in Appendix~\ref{app:expDetails}).
We obtain an approximation of $f$ as a weighted sum of such polynomials,
and define our complexity measure
$\operatorname{C}_\text{Chebyshev}$
exactly as above,
\ie the average of the coefficients weighted by their corresponding order.
For example, the first two basis elements are a constant and a first-order polynomial,
hence the decomposition of a linear $f$ will use large coefficients on these low-order elements
and give a low complexity.
We implemented this method with several canonical bases
of orthogonal polynomials (Hermite, Legendre, and Chebyshev polynomials) and found the latter to be the most numerically stable.

\label{lz}
\item \textbf{Compressibility} has been proposed as an approximation of the Kolmogorov complexity~\cite{dingle2018input,goldblum2023no,valle2018deep}.
We apply the classical Lempel-Ziv (LZ) compression on the sequence
$\bY \!\!=\!\! \{f(\bx_i)\!: \bx_i \!\in\! \bX\}$.
We then use the size of the dictionary built by the algorithm
as our measure of complexity $\operatorname{C}_\text{LZ}(f)$.
A sequence with repeating patterns will require a small dictionary and give a low complexity.
\end{itemize}

\section{Inductive Biases in Random Networks}
\label{sec:untrained}
\label{sec:heatmaps}
\label{sec:visualizationsMappings}

We are now equipped to compare
architectures.
We will show that various components shift the inductive bias towards low or high complexity
(see Table~\ref{tabArchitecturalElements}).
In particular, ReLU activations will prove critical
for a simplicity bias insensitive to depth and weight\,/\,activation magnitude.

\begin{table}[!t]
  \renewcommand{\tabcolsep}{0.0em}
  \renewcommand{\arraystretch}{1.2}
  \vspace{-6pt}
  \caption{\label{tabArchitecturalElements}
  Components that bias NNs towards
  low/high complexity.
  \vspace{-16pt}}
  \footnotesize
  \begin{tabularx}{\linewidth}{p{.36\linewidth} p{.28\linewidth} p{.36\linewidth}}
    \toprule
    \makecell{\textbf{Lower complexity}} & \makecell{\textbf{No impact}} & \makecell{\textbf{Higher complexity}}\\
    \midrule
    \makecell{ReLU-like activations\\Small weights\,/\,activations\\Layer normalization\\Residual connections}&
    \makecell{Width\\Bias magnitudes}&
    \makecell{Other activations\\Large weights\,/\,activations\\Depth\\Multiplicative interactions}\\
    \bottomrule
  \end{tabularx}
  \vspace{-4pt}
\end{table}

\paragraph{ReLU MLPs.}
We start with a 1-hidden layer multi-layer perceptron (MLP) with ReLU activations.
We will then 
examine variations of this architecture.
Formally, each hidden layer applies a transformation on the input:
$\bx \!\leftarrow\! \phi(\bW \bx + \bb)$
with weights $\bW$, biases $\bb$, and activation function $\phi(\cdot)$.
MLPs are so simple that they are often thought as providing little or no inductive bias~\cite{bachmann2023scaling}.
On the contrary, we observe in Figures~\ref{figHeatmaps} \& \ref{figCorrelation} that MLPs have a very strong bias towards low-frequency, low-order, compressible functions.
And this simplicity bias is remarkably unaffected by the magnitude of the weights, nor increased depth.

The variance in complexity across the random networks is essentially zero:
virtually \emph{all} of them have low complexity.
This does not mean that they cannot represent complex functions,
which would violate their universal approximation property~\cite{hornik1989multilayer}.
Complex functions simply require precisely-adjusted weights and biases that are unlikely to be found by random sampling.
These can still be found by gradient descent though, as we will see in Section~\ref{sec:trained}.

\label{sec:otherActivations}
\paragraph{ReLU-like activations}%
(GELU, Swish, SELU \cite{dubey2022activation})
are also biased towards low complexity.
Unlike ReLUs, close examination in Appendix~\ref{app:untrainedFullResults}
shows that increasing depth or weight magnitudes slightly increases the complexity.

\paragraph{Others activations}%
(TanH, Gaussian, sine)
show completely different behaviour.
Depth and weight magnitude cause a dramatic increase in complexity.
Unsurprisingly, these activations
are only used in 
special applications~\cite{ramasinghe2022beyond} 
with careful initializations~\cite{sitzmann2020implicit}.
Networks with these activations have no fixed preferred complexity
independent of the weights' or  activations' magnitudes.%
\footnote{The weight magnitudes examined in Figure~\ref{figHeatmaps} are larger than
typically used for initialization,
but the same effects would result from large \emph{activation} magnitudes that occur in trained models.}
Mechanistically, the dependency on weight magnitudes is trivial to explain.
Unlike with a ReLU, the output
\eg of a GELU
is not equivariant to a rescaling of the weights and biases.

\begin{figure}[t]
  \centering
  \renewcommand{\tabcolsep}{0.2em}
  \renewcommand{\arraystretch}{1}
  \small
  \begin{tabularx}{\linewidth}{p{.18\linewidth} cc}
    &\scriptsize\textsf{ReLU} & \scriptsize\textsf{TanH}\\
    \makecell{\small Weights $\sim$\\$\mathcal{U}(-s,s)$}&
    \raisebox{-.4\height}{%
    \includegraphics[width=.19\linewidth]{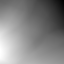}%
    \includegraphics[width=.19\linewidth]{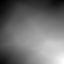}%
    }&
    \raisebox{-.4\height}{%
    \includegraphics[width=.19\linewidth]{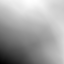}%
    \includegraphics[width=.19\linewidth]{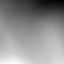}%
    }\\[18pt]
    \makecell{\small Weights $\sim$\\$\mathcal{U}(-6 s, 6 s)$}&
    \raisebox{-.8\height}{%
    \includegraphics[width=.19\linewidth]{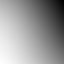}%
    \includegraphics[width=.19\linewidth]{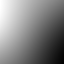}%
    }&
    \raisebox{-.8\height}{%
    \includegraphics[width=.19\linewidth]{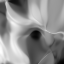}%
    \includegraphics[width=.19\linewidth]{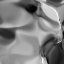}%
    }
  \end{tabularx}\vspace{-2pt}
  \caption{\label{figFctMapsCloseUp}
  Comparison of functions implemented by random MLPs (2D input, 3 hidden layers).
  ReLU and TanH architectures are biased towards different functions
  despite their universal approximation capabilities.
  ReLU architectures have the unique property of maintaining their simplicity bias
  regardless of weight magnitude.
  \vspace{-10pt}}
\end{figure}

\begin{figure*}[htb]
  \centering
  \renewcommand{\tabcolsep}{0.93em}
  \renewcommand{\arraystretch}{1.1}
  \small
  \begin{tabularx}{\linewidth}{cccccccc}
    \scriptsize\textsf{ReLU} & \scriptsize\textsf{GELU} & \scriptsize\textsf{Swish} & \scriptsize\textsf{SELU} & \scriptsize\textsf{Tanh} & \scriptsize\textsf{Gaussian} & \scriptsize\textsf{Sin} & \scriptsize\textsf{Unbiased}\\
    \includegraphics[width=.09\linewidth]{figFourier/fourier-spectrumIntegerFrequencies-signed-mlpRelu-fourier-heatmapWtNumLayers-10x6.pdf}&
    \includegraphics[width=.09\linewidth]{figFourier/fourier-spectrumIntegerFrequencies-signed-mlpGelu-fourier-heatmapWtNumLayers-10x6.pdf}&
    \includegraphics[width=.09\linewidth]{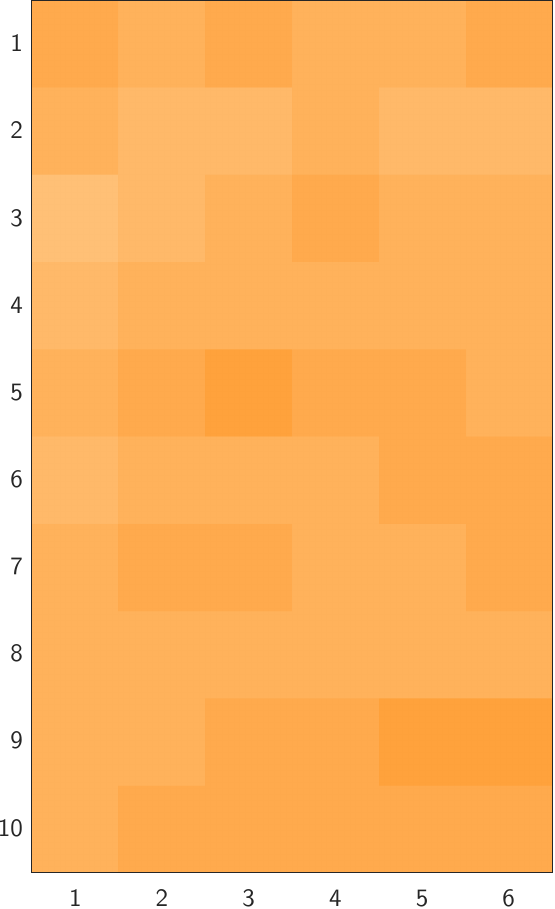}&
    \includegraphics[width=.09\linewidth]{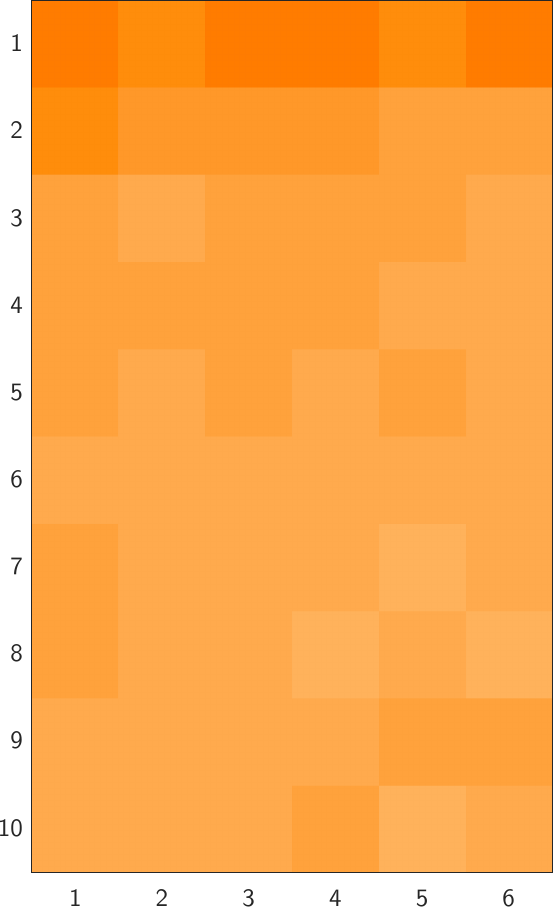}&
    \includegraphics[width=.09\linewidth]{figFourier/fourier-spectrumIntegerFrequencies-signed-mlpTanh-fourier-heatmapWtNumLayers-10x6.pdf}&
    \includegraphics[width=.09\linewidth]{figFourier/fourier-spectrumIntegerFrequencies-signed-mlpGaussian-fourier-heatmapWtNumLayers-10x6.pdf}&
    \includegraphics[width=.09\linewidth]{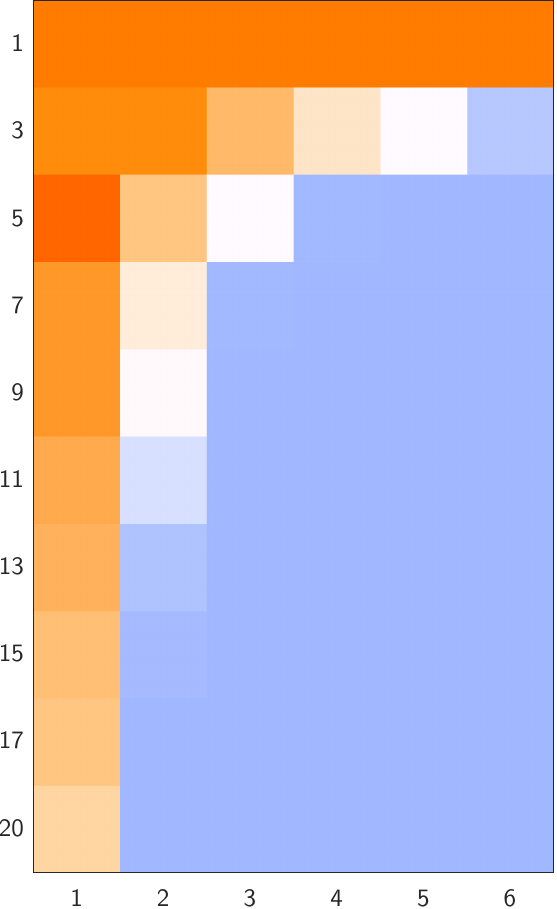}&
    \includegraphics[width=.09\linewidth]{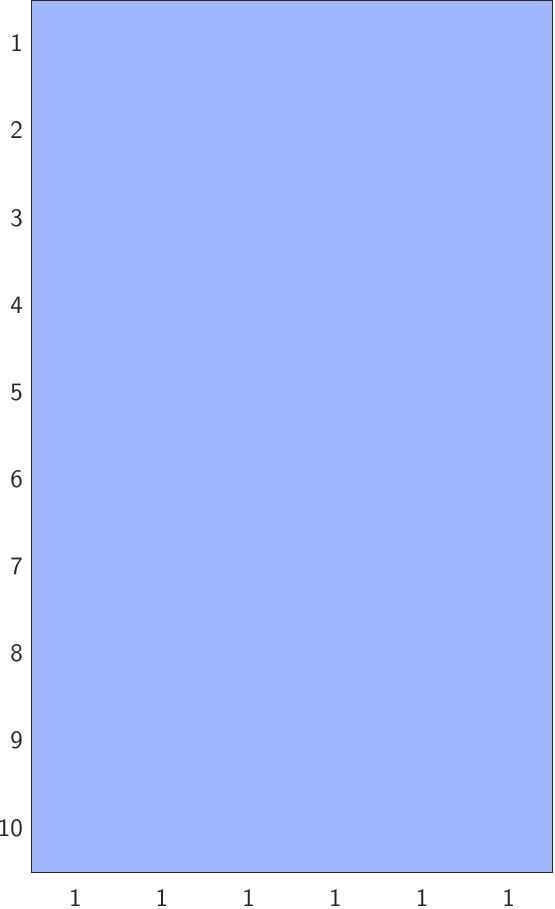}\\
    \includegraphics[width=.09\linewidth]{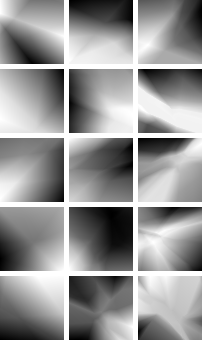}&
    \includegraphics[width=.09\linewidth]{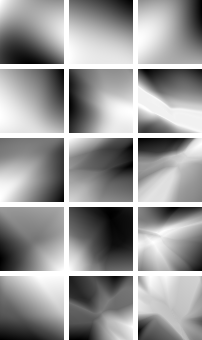}&
    \includegraphics[width=.09\linewidth]{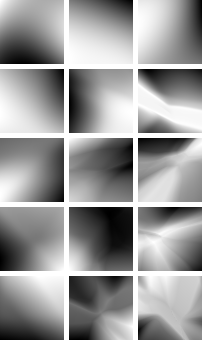}&
    \includegraphics[width=.09\linewidth]{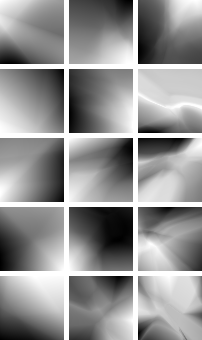}&
    \includegraphics[width=.09\linewidth]{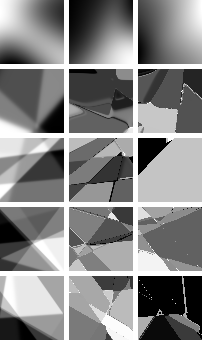}&
    \includegraphics[width=.09\linewidth]{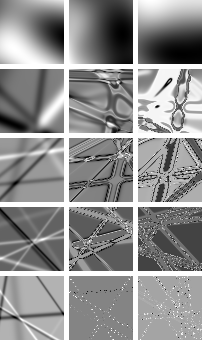}&
    \includegraphics[width=.09\linewidth]{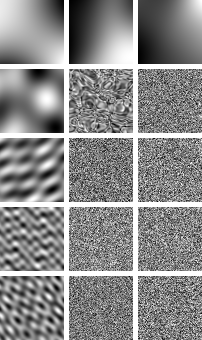}&
    \includegraphics[width=.09\linewidth]{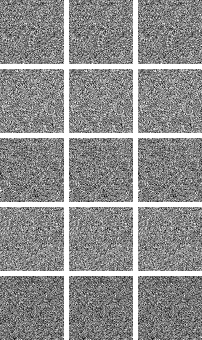}\\
  \end{tabularx}\vspace{-6pt}
  \caption{\label{figHeatmaps}
  Heatmaps of the average Fourier complexity
  of functions implemented by random-weight networks.
  Each heatmap corresponds to an activation function
  and each cell (within a heatmap) corresponds to a depth (heatmap columns) and weight magnitude (heatmap rows).
  We also show grayscale images of functions implemented by networks of an architecture
  corresponding to every other heatmap cell.
  \vspace{-10pt}}
\end{figure*}

\begin{figure*}[ht]
  \renewcommand{\tabcolsep}{0.35em}
  \renewcommand{\arraystretch}{1}
  \small
  \begin{tabularx}{\linewidth}{cccccccccc}
    & \scriptsize\textsf{ReLU} & \scriptsize\textsf{GELU} & \scriptsize\textsf{Swish} & \scriptsize\textsf{SELU} & \scriptsize\textsf{TanH} & \scriptsize\textsf{Gaussian} & \scriptsize\textsf{Sin} & \scriptsize\textsf{Unbiased}\\[-1pt]
    \raisebox{10pt}{\rotatebox{90}{\scriptsize\textsf{Complexity (Fourier)}}}&
    \includegraphics[width=.09\linewidth]{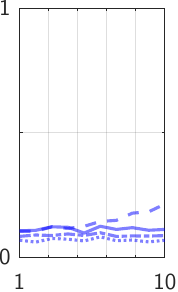}&
    \includegraphics[width=.09\linewidth]{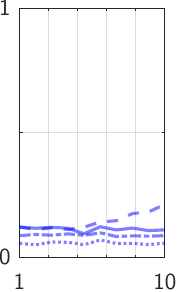}&
    \includegraphics[width=.09\linewidth]{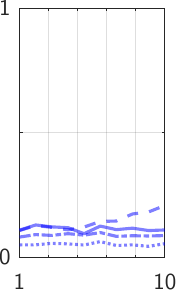}&
    \includegraphics[width=.09\linewidth]{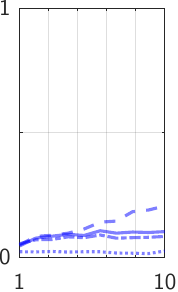}&
    \includegraphics[width=.09\linewidth]{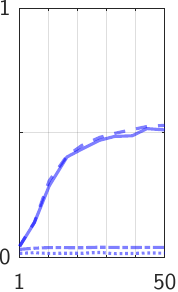}&
    \includegraphics[width=.09\linewidth]{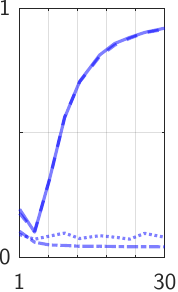}&
    \includegraphics[width=.09\linewidth]{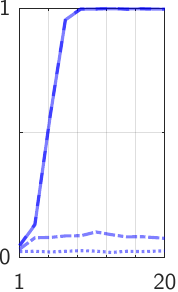}&
    \includegraphics[width=.09\linewidth]{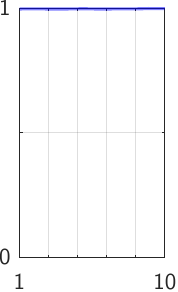}&
    \raisebox{.20\height}{\includegraphics[width=.14\linewidth]{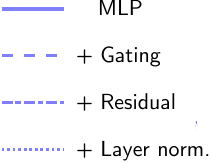}}\\
  \end{tabularx}\vspace{-7pt}
  \caption{\label{figPlotsComponents}
  The complexity of random models (Y axis)
  generally increases with weights\,/\,activations magnitudes (X axis).
  The sensitivity is however very different across activation functions.
  This sensitivity also increases with multiplicative interactions (\ie gating),
  decreases with residual connections,
  and is essentially absent with layer normalization.
  \vspace{-8pt}
  }
\end{figure*}

Figure~\ref{figCorrelation} shows close correlations between complexity measures, though they measure different proxies.
Figure~\ref{figFctMapsCloseUp}
shows that different activations make 
distinctive patterns
not captured by the complexity measures.
More work will be needed to characterize such fine-grained differences.

\paragraph{Width}%
has no impact on complexity, perhaps surprisingly.
Additional neurons change the capacity of a model (what can be represented after training) but they do not affect its inductive biases.
Indeed, the contribution of all neurons in a layer averages out to something invariant to their number.

\paragraph{Layer normalization}
is a popular component in modern architectures, including transformers~\cite{radford2019language}.
It shifts and rescales the internal representation to zero mean and unit variance~\cite{ba2016layer}.
We place layer normalizations before each activation
such that each hidden layer now applies the transformation:
$\bx \!\leftarrow\! (\bW \bx + \bb);~ 
 \bx \!\leftarrow\! \phi( (\bx - \bar{\bx}) / \operatorname{std}(\bx) )$
 where $\bar{\bx}$ and $\operatorname{std}(\bx)$
 denote the mean and standard deviation across channels.
Layer normalization has the significant effect of 
removing variations in complexity with the weights' magnitude for all activations
(Figure~\ref{figPlotsComponents}).
The weights can now vary (\eg during training) without directly affecting the
preferred complexity of the architecture.
Layer normalizations also usually apply a learnable offset ($0$ by default) and scaling ($1$ by default) post-normalization.
Given the above observations, when paired with an activation with some slight sensitivity to weight magnitude (\eg GELUs, see Appendix~\ref{app:heatmapsDetails}),
this scaling can now be interpreted as a learnable shift in complexity, modulated by a single scalar
(rather than the whole weight matrix without the normalization).

\paragraph{Residual connections}~\cite{he2016deep}.
We add these
such that each non-linearity is now described with:
$\bx \!\leftarrow\! (x + \phi(\bx))$.
This has the dramatic effect of forcing the preferred complexity to some of the lowest levels for all activations regardless of depth.
This can be explained by the fact that residual connections essentially
bypass the stacking of non-linearities that causes the increased complexity with increased depth.

\paragraph{Multiplicative interactions}
refer to multiplications of internal representations with one another~\cite{jayakumar2020multiplicative}
as in attention layers, highway networks, dynamic convolutions, etc.
We place them in our MLPs as gating operations,
such that each hidden layer corresponds to:
$\bx \!\leftarrow\! \big(\phi(W \bx + b) ~\odot~ \sigma(W' \bx + b') \big)$
where $\sigma(\cdot)$ is the logistic function.
This creates a clear increase in complexity dependent on depth and weight magnitude, even for ReLU activations.
This agrees with prior work showing that multiplicative interactions
in polynomial networks~\cite{choraria2022spectral} facilitate learning high frequencies.

\label{sec:unbiasedLearner}
\paragraph{Unbiased model.}
As a counter-example to models showing some preferred complexity,
we construct an architecture with no bias by design
in the complexity measured with Fourier frequencies.
This special architecture implements an inverse Fourier transform,
parametrized directly with the coefficients and phase shifts of the Fourier components
(details in Appendix~\ref{app:expDetails}).
The inverse Fourier transform is a weighted sum of sine waves,
so this architecture
can be implemented as 
a one-layer MLP with sine activations
and fixed input weights representing each one Fourier component.
These fixed weights prior to sine activations thus enforce a \emph{uniform prior over frequencies}.

This architecture behaves very differently from standard MLPs (Figure~\ref{figHeatmaps}).
With random weights, its Fourier spectrum
is uniform,
which gives a high complexity
for any weight magnitude (depth is fixed).
Functions implemented by this architecture look
like white noise.
Even though this architecture can be trained by gradient descent like any MLP,
we show in Appendix~\ref{app:trained} that it is practically useless because of its lack of any complexity bias.

\begin{figure}[h!]
  \centering
  \renewcommand{\tabcolsep}{0pt}
  \renewcommand{\arraystretch}{1}
  \small
  \begin{tabularx}{\linewidth}{p{.33\linewidth} p{.33\linewidth} p{.33\linewidth}}
    \makecell[l]{\includegraphics[width=.95\linewidth]{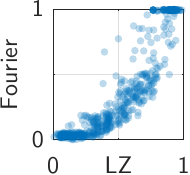}}&
    \makecell[l]{\includegraphics[width=.95\linewidth]{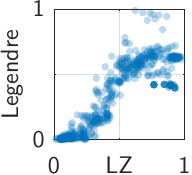}}&
    \makecell[l]{\includegraphics[width=.95\linewidth]{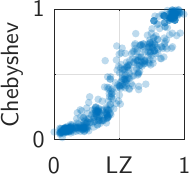}}
  \end{tabularx}\vspace{-6pt}
  \caption{\label{figCorrelation}
  Our various complexity measures are closely correlated
  despite measuring each a different proxy
  \ie frequency (Fourier), polynomial order (Legendre, Chebyshev), or compressibility (LZ).
  }
\end{figure}

\newpage 
\begin{mdframed}[style=citationFrame,userdefinedwidth=\linewidth,align=left,skipabove=6pt,skipbelow=0pt,innertopmargin=6pt,innerbottommargin=6pt]
\textbf{Importance of ReLU activations}\\[2pt]
The fact that a strong simplicity bias depends on ReLU activations suggests that their historical importance in the development of deep learning goes beyond the common narrative about vanishing gradients~\cite{maas2013rectifier}. The same may apply to residual connections and layer normalization
since they alsox contribute strongly to the simplicity bias.
This contrasts with the current literature that mostly invokes their numerical properties~\cite{balduzzi2017shattered,xiong2020layer,xu2019understanding}.
\end{mdframed}

\clearpage
\section{Inductive Biases in Trained Models}
\label{sec:trained}

We now examine how the inductive biases of an architecture
impact a model trained by standard gradient descent.
We will show that there is a strong correlation between
the complexity
at initialization (\ie with random weights as examined in the previous section)
and in the trained model.
We will also see that unusual architectures with a bias towards high complexity
can improve generalization
on tasks where the standard ``simplicity bias'' is suboptimal.

\subsection{Learning Complex Functions}
\label{sec:difficultTasks}

The NRS proposes that good generalization requires a good match between the complexity
preferred by the architecture
and the target function.
We verify this claim
by demonstrating improved generalization on complex tasks
with architectures biased towards higher complexity.
This is also a proof of concept of the potential utility of controlling inductive biases.

\begin{figure}[ht]
  \centering
  \renewcommand{\tabcolsep}{0em}
  \renewcommand{\arraystretch}{1}
  \begin{tabularx}{\linewidth}{p{.025\linewidth} p{.35\linewidth} p{.3125\linewidth} p{.3125\linewidth}}
    &\multicolumn{3}{c}{\footnotesize{\textsf{Target function (3 versions of ``modulo addition'')}}}\\[-6pt]
    &\multicolumn{3}{c}{\rule{.9\linewidth}{.5pt}}\\[1pt]
    &\makecell{\raisebox{-.3em}{\fbox{\includegraphics[height=1.05em]{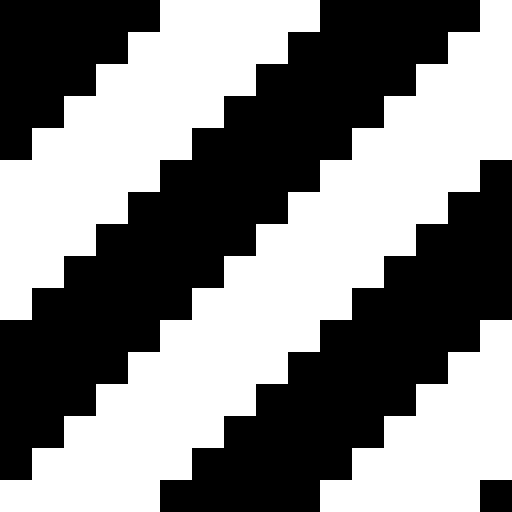}}}
    \scriptsize{\textsf{Low freq.}}}&
    \makecell{\raisebox{-.3em}{\fbox{\includegraphics[height=1.05em]{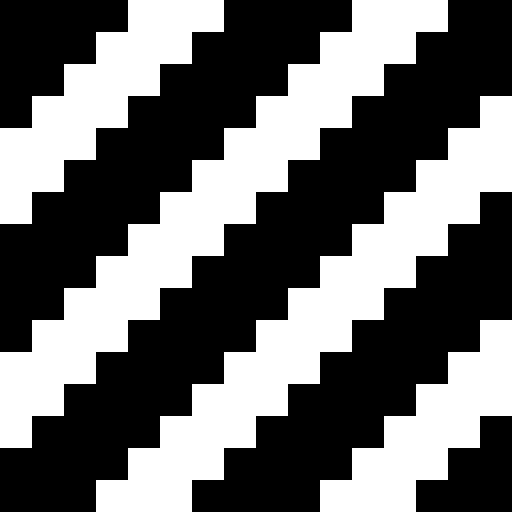}}}
    \scriptsize{\textsf{Med. freq.}}}&
    \makecell{\raisebox{-.3em}{\fbox{\includegraphics[height=1.05em]{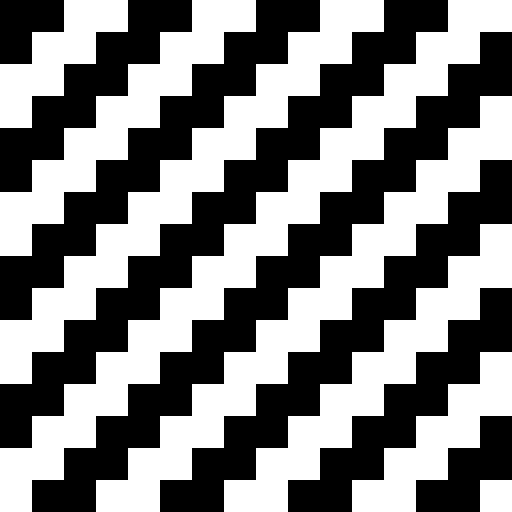}}}
    \scriptsize{\textsf{High freq.}}}\\[3pt]
    \raisebox{18pt}{\rotatebox{90}{\scriptsize\textsf{\hspace{0pt}Test accuracy}}}&
    \includegraphics[height=7em]{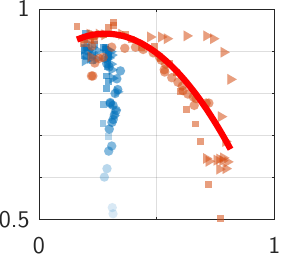}&
    \includegraphics[height=7em, clip, trim=16pt 0 0 0]{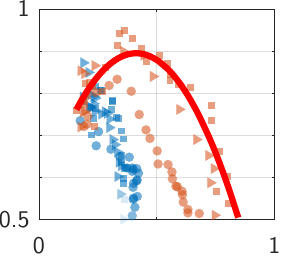}&
    \includegraphics[height=7em, clip, trim=16pt 0 0 0]{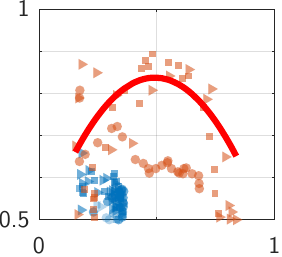}\\[-1pt]
    &\multicolumn{3}{c}{\footnotesize{\textsf{Complexity (LZ) at initialization}}}\\[-1pt]
    &\multicolumn{3}{c}{\footnotesize\textsf{(modulated by choices of activation and prefactor value)}}\\[4pt]
  \end{tabularx}
  \includegraphics[width=1\linewidth]{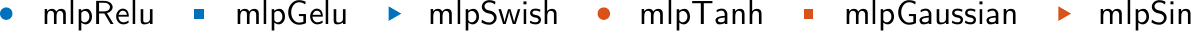}
  \\\vspace{-3pt}
  \caption{\label{figModulo}
  Results training networks on three tasks of increasing complexity.
  Each point represents a different architecture.
  \textbf{\textcolor{RoyalBlue}{ReLU-like activations}} are biased towards low complexity
  and fail to generalize on complex tasks.
  With \textbf{\textcolor{RawSienna}{other activations}},
  the complexity bias depends on the activation magnitudes,
  which we can control with a multiplicative prefactor.
  This enables generalization
  on complex tasks
  by shifting the bias to higher complexity.
  Indeed, the optimum prefactor (peak of the quadratic fit) shifts to the right
  on each task of increasing complexity.
  \vspace{-3pt}
  }
\end{figure}

\paragraph{Experimental setup.}
We consider a simple binary classification task involving modulo arithmetic. Such tasks like the parity function~\cite{shalev2017failures} are known to be challenging
for standard architectures
because they contain high-frequency patterns.
The input to our task is a vector of integers $\bx \!\in\! [0,N\!-\!1]^d$.
The correct labels are ${\mathbb{1}}(\Sigma x_i \leq (M/2) \operatorname{mod} M)$.
We consider three versions
with $N\!=\!16$ and $M\!=\!\{10,7,4\}$
that correspond to increasingly higher frequencies in the target function
(see Figure~\ref{figModulo} and Appendix~\ref{app:expDetails} for details).

\paragraph{Results.}
We see in Figure~\ref{figModulo}
that a ReLU MLP only solves the low-frequency version of the task.
Even though this model can be trained to perfect training accuracy
on the higher-frequency versions,
it then fails to generalize 
because of its simplicity bias.
We then train MLPs with other activations (TanH, Gaussian, sine) whose preferred complexity is sensitive to the activations' magnitude.
We also introduce a constant multiplicative prefactor before each activation function
to modulate this bias without changing the weights' magnitude, which could introduce optimization side effects.
Some of these models succeed in learning all versions of the task when the prefactor is correctly tuned.
For higher-frequency versions, the prefactor needs to be larger
to shift the bias towards higher complexity.
In Figure~\ref{figModulo}, we fit a quadratic approximation to the accuracy of Gaussian-activated models.
The peak then clearly shifts to the right on the complex tasks.
This agrees with the NRS proposition that \textbf{complexity at initialization relates to properties of the trained model}.

Let us also note that not all activations succeed, even with a tuned prefactor.
This shows that matching the complexity of the architecture and of the target function is beneficial but not sufficient for good generalization.
The inductive biases of an architecture are clearly not fully captured by any of our measures of complexity.

\setlength{\fboxsep}{0pt}
\setlength{\fboxrule}{.25pt}

\setlength{\fboxsep}{0pt}
\setlength{\fboxrule}{.25pt}
\begin{figure}[h]
  \centering
  \renewcommand{\tabcolsep}{.18em}
  \renewcommand{\arraystretch}{1}
  \begin{tabular}{ccccc}
    &\scriptsize{\textsf{Low frequency}}&\scriptsize{\textsf{Medium frequency}}&\scriptsize{\textsf{High frequency}}\\[-1pt]
    \rotatebox{90}{\scriptsize\textsf{\hspace{15pt}Test accuracy}} &
    \includegraphics[height=6.1em, clip, trim=0 0 0 130pt]{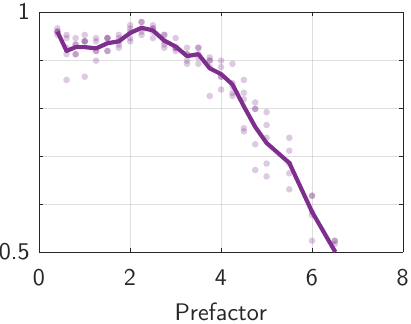}&
    \includegraphics[height=6.1em, clip, trim=16pt 0 0 130pt]{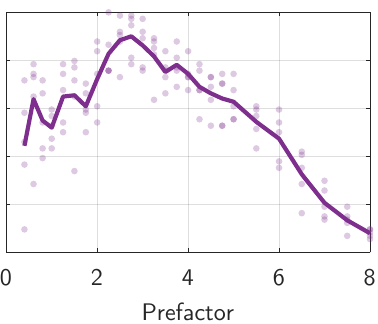}&
    \includegraphics[height=6.1em, clip, trim=16pt 0 0 130pt]{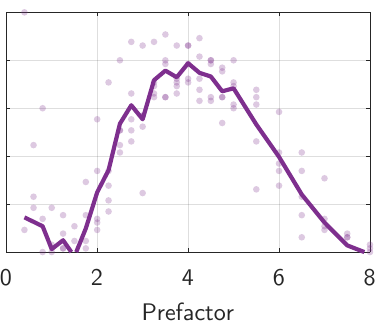}\\[-1pt]
  \end{tabular}\vspace{-7pt}
  \caption{\label{figModuloGaussian}
  Detail of Figure~\ref{figModulo} for Gaussian activations. The peak accuracy shifts to the right on tasks of increasing complexity. This corresponds to a larger prefactor that shifts the bias towards higher complexity.
  Each point represents one random seed.
  \vspace{-13pt}
  }
\end{figure}

\begin{mdframed}[style=citationFrame,userdefinedwidth=\linewidth,align=left,skipabove=0pt,skipbelow=0pt,innertopmargin=6pt,innerbottommargin=6pt]
\textbf{Reinterpretation of existing work}\\[1pt]
\textit{Loss landscapes Are All You Need} \cite{chiang2022loss}\\[1pt]
Chiang et al.\ showed that networks with random weights, as long as they fit the training data
with low training loss, are good solutions that generalize to the test data.
We find ``\textit{loss landscapes}'' slightly misleading
because the key is in the parametrization of the network
(and by extension of this landscape) and not in the loss function.
Their results can be replicated by replacing the cross-entropy loss with an MSE loss,
but not by replacing their MLP with our ``unbiased learner'' architecture.

\vspace{2pt}
The sampled solutions are good, not only because of their low training loss,
but because they are found by uniformly sampling the weight space.
Bad low-loss solutions also exist, but they are unlikely to be found by random sampling.
Because of the NRS, all random-weight networks implement simple functions,
which generalize as long as they fit the training data.
An alternative title could be ``\textit{Uniformly Sampling the Weight Space Is All You Need}''.
\end{mdframed}

\clearpage
\subsection{Impact on Shortcut Learning}
Shortcut learning refers to situations where
the simplicity bias causes
a model to rely on simple spurious features
rather than learning the more-complex target function~\cite{shah2020pitfalls}.

\paragraph{Experimental setup.}
We consider a regression task similar to Colored-MNIST.
Inputs are images of handwritten digits juxtaposed with a uniform band of colored pixels that simulate spurious features. The labels in the training data are values in $[0,1]$ proportional to the digit value as well as to the color intensity.
Therefore, a model can attain high training accuracy by relying either on the 
simple linear relation with the color, or the more complex recognition of the digits
(the target task).
To measure the reliance of a model on color or digit, we use two test sets where either the color or digit is correlated with the label while the other is randomized.
See Appendix~\ref{app:expDetails} for details.

\paragraph{Results.}
We train 2-layer MLPs with different activation functions.
We also use a multiplicative prefactor, \ie a constant $\alpha \!\in\! \mathrm{R}^{+}$ placed before each activation function such that each non-linear layer performs the following:
$\bx \!\leftarrow\! \phi\big(\alpha (W \bx + b)\big)$.
The prefactor mimics a rescaling of the weights and biases with no optimization side effects.

\begin{figure}[hb!]
  \renewcommand{\tabcolsep}{0.0em}
  \renewcommand{\arraystretch}{1}
  \small
  \begin{tabular}{ccccc}
    ~ & \scriptsize\textsf{TanH} & \scriptsize\textsf{Gaussian} & \scriptsize\textsf{Sine}\\[-1pt]
    \rotatebox{90}{\tiny\textsf{\hspace{20pt}Test accuracy\hspace{9pt}Complexity at init. (LZ)}} &
    \includegraphics[width=.32\linewidth]{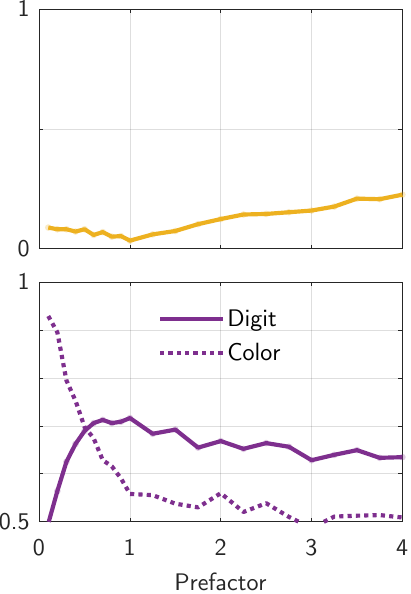}&
    \includegraphics[width=.32\linewidth]{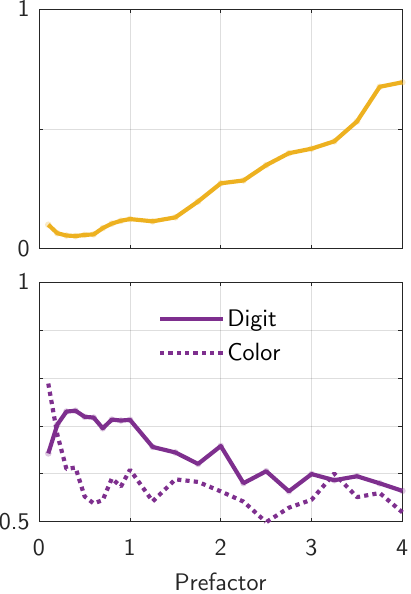}&
    \includegraphics[width=.32\linewidth]{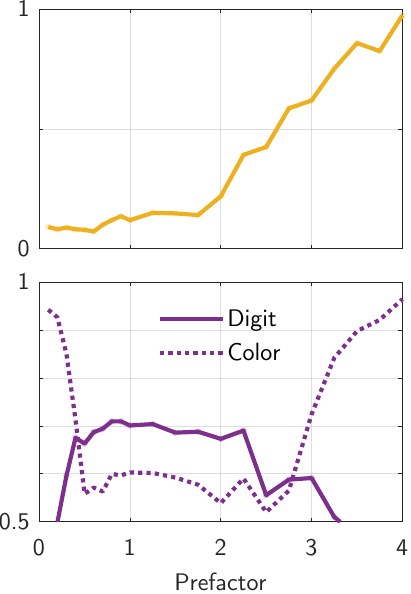}\\
  \end{tabular}\vspace{-8pt}
  \caption{\label{figColoredMnist}
  Experiments on Colored-MNIST show a clear correlation between
  complexity at initialization (top) and test accuracy (bottom).
  Models with a bias for low complexity rely on the color \ie the simpler feature.
  The accuracy on the digit peaks at a sweet spot where the models' preferred complexity matches the digits'.
  \vspace{-8pt}
  }
\end{figure}

We see in Figure~\ref{figColoredMnist}
that the LZ complexity at initialization
increases with prefactor values for TanH, Gaussian, and sine activations.
Most interestingly, the accuracy on the digit and color also varies with the prefactor.
The color is learned more easily with small prefactors
(corresponding to a low complexity at initialization)
while the digit is learned more easily at an intermediate value
(corresponding to medium complexity at initialization).
The best performance on the digit is reached at a sweet spot
that we explain as the hypothesized ``best match'' between the complexity of the target function, and that preferred by the architecture.
With larger prefactors, \ie beyond this sweet spot, the accuracy on the digit decreases,
and even more so with sine activations for which the complexity also increases more rapidly,
further supporting the proposed explanation.

\vspace{-7pt}


\begin{mdframed}[style=citationFrame,userdefinedwidth=\linewidth,align=left,skipabove=6pt,skipbelow=0pt,innertopmargin=6pt,innerbottommargin=6pt]
\textbf{Reinterpretation of existing work}\\[1pt]
\textit{How You Start Matters for Generalization} \cite{ramasinghe2022you}\\[1pt]
Ramasinghe \textit{et al.} examine implicit neural representations (\ie a network trained to represent one image).
They observe that models showing high frequencies at initialization
also learn high frequencies better.
They conclude that complexity at initialization \emph{causally} influences the solution.
But our results suggest instead that these are two effects of a common cause:
the architecture is biased towards a certain complexity, which influences both
the untrained model and the solutions found by gradient descent.
There exist configurations of weights that correspond to complex functions, but they are
unlikely to be found in either case.
Appendix~\ref{app:expFineTuning} shows that
initializing GD from such a solution
with an architecture biased toward simplicity does not yield complex solutions,
thus disproving the causal relation.
\end{mdframed}


\section{Transformers are Biased Towards\\Compressible Sequences}
\label{sec:transformers}

We now show that the inductive biases observed with MLPs
are relevant to transformer sequence models.
The experiments below confirm the bias of a transformer for generating simple, compressible sequences~\cite{goldblum2023no}
which could then explain the tendency of language models to repeat themselves~\cite{fu2021theoretical,holtzman2019curious}.
The experiments also suggest that transformers inherit this inductive bias
from the same components
as those explaining
the simplicity bias in MLPs.

\paragraph{Experimental setup.}
We sample sequences from an {untrained} GPT-2~\cite{radford2019language}.
For each sequence, we sample random weights from their default initial distribution,
then prompt the model with one random token (all of them being equivalent since the model is untrained),
then generate a sequence of 100 tokens by greedy maximum-likelihood decoding.
We evaluate the complexity of each sequence with the LZ measure (Section~\ref{lz})
and report the average over 1,000 sequences.
We evaluate variations of the architecture:
replacing activation functions in MLP blocks (GELUs by default),
varying the depth ($12$ transformer blocks by default),
and varying the activations' magnitude by modifying the scaling factor in layer normalizations ($1$ by default).

\begin{figure}[h!]
  \centering
  \renewcommand{\tabcolsep}{0.4em}
  \renewcommand{\arraystretch}{1}
  \begin{tabular}{cccc}
    \raisebox{12pt}{\rotatebox{90}{\scriptsize\textsf{Complexity (LZ)}}}&
    \includegraphics[height=.27\linewidth]{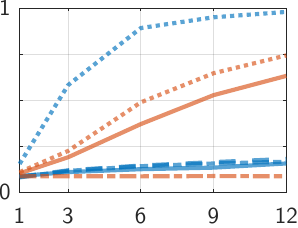}&
    \includegraphics[height=.27\linewidth]{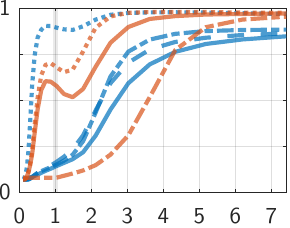}&
    \includegraphics[height=.27\linewidth]{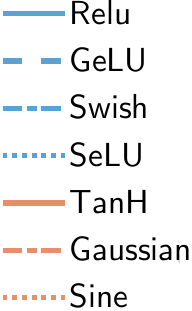}\\[-2pt]
    ~&\scriptsize\textsf{Depth (num. of transformer}&\scriptsize\textsf{Scaling in layer}&\\[-3pt]
    ~&\scriptsize\textsf{blocks, scaling=1)}&\scriptsize\textsf{normalization (depth=12)}&\\
  \end{tabular}\vspace{-5pt}
  \caption{\label{figTransformers}
  Average complexity (LZ) of sequences generated by an untrained GPT-2.
  Variations of the architecture correspond to variations in complexity comparable
  to MLPs.
  This suggests that transformers inherit a bias for simple sequences
  from their building blocks
  via mechanisms similar to those in simple models.
  \vspace{-8pt}}
\end{figure}

\paragraph{Results.}
We first observe in Figure~\ref{figTransformers} that the default architecture is biased towards relatively simple sequences.
This observation, already reported by~\citet{goldblum2023no}, is non-trivial since a random model
could as well generate completely random sequences.
Changing the \textbf{activation function} from the default GELUs has a large effect.
The complexity increases with SELU, TanH, sine, and decreases with ReLU.
It is initially low with Gaussian activations, but climbs higher than most others with larger activation magnitudes.
This is
consistent with observations made on MLPs, where ReLU induced the strongest bias for simplicity,
and TanH, Gaussian, sine for complexity.
Variations of \textbf{activations' magnitude} (via scaling in layer normalizations)
has the same monotonic effect on complexity as observed in MLPs.
However, we lack an explanation for the ``shoulders'' in the curves of SELU, Tanh, and sine.
It may relate to them being the activations that output negative values most.
Varying \textbf{depth} also has the expected effect
of magnifying the differences across activations and scales.

\paragraph{Take-away.}
These results suggest that the bias for simple sequences of transformers originates from their building blocks
via similar mechanisms to those causing the simplicity bias in other predictive models.
The building blocks of transformers also seem to balance
a shift towards higher complexity (attention, multiplicative interactions)
and lower complexity (GELUs, layer normalizations, residual connections).


\section{Related Work}
\label{sec:relatedWork}

Much effort has gone into explaining the success of deep learning
through the inductive biases of SGD~\cite{neyshabur2014search}
and structured architectures \cite{cohen2016inductive,zhou2023algorithms}.
This work rather focuses on implicit inductive biases from unstructured architectures.

The \textbf{simplicity bias}
is the tendency of NNs to fit data with simple functions
\cite{arpit2017closer,geirhos2020shortcut,poggio2018theory,teney2021evading}.
The \textbf{spectral bias} 
suggests that NNs prioritize learning low-frequency components of the target function~\cite{rahaman2019spectral,xu2019frequency}.
These studies confound architectures and optimization.
And most explanations 
invoke implicit regularization of gradient descent~\cite{soudry2018implicit,xu2019training}
and are specific to ReLU networks~\cite{hong2022activation,huh2021low,zhang2023shallow}.
In contrast, we show that some form of spectral bias exists in common architectures independently of gradient descent.

A related line of study showed that Boolean MLPs are biased towards low-entropy functions~\cite{de2019random,mingard2019neural}.
Work closer to ours~\cite{de2019random,mingard2019neural,valle2018deep} examines the simplicity bias of networks with \textbf{random weights}.
These works are limited to MLPs with binary inputs or outputs~\cite{de2019random,mingard2019neural}, ReLU activations, and simplicity measured as compressibility.
In contrast, our work examines multiple measures of simplicity and a wider set of architectures.
In work concurrent to ours, \citet{abbe2023generalization} used Walsh decompositions (analogous to Fourier series for binary functions) to characterize the simplicity of learned binary classification networks.
Their discussion is specific to classification and highly complementary to ours.

Our work also provides a new lens to explain why choices of activation functions are critical~\cite{dubey2022activation,ramasinghe2022frequency,simon2022reverse}.
See Appendix~\ref{sec:relatedWorkShort} for an extended literature review.
\section{Conclusions}
We examined inductive biases that NNs possess independently of their optimization.
We found that the parameter space of popular architectures
corresponds overwhelmingly to functions
with three quantifiable properties: low frequency, low order, and compressibility.
They correspond to the simplicity bias previously observed in \emph{trained} models
which we now explain without involving (S)GD.
We also showed that the simplicity bias is not universal to all architectures.
It results from ReLUs, residual connections, layer normalization, etc.
The popularity of these components likely reflects
the collective search for architectures that perform well on real-world data.
In short, the effectiveness of NNs is not an intrinsic property
but the result of the adequacy between key choices (\eg ReLUs)
and properties of real-world data (prevalence of low-complexity patterns).

\paragraph{Limitations and open questions.}
\begin{itemize}[itemsep=3pt,topsep=1pt]
\item
Our analysis used mostly \textbf{small models and data}
to enable visualizations (2D function maps)
and computations (Fourier decompositions).
We showed the relevance of our findings to large transformers,
but the study could be extended to other large architectures and tasks.

\item
Our analysis relies on \textbf{empirical simulations}.
It could be carried out analytically to provide theoretical insights.

\item 
Our results do not invalidate prior work on implicit biases of (S)GD.
Future work should clarify the \textbf{interplay of different sources of inductive biases}.
Even if most of the parameter space corresponds to simple functions,
GD can navigate to complex ones. 
Are they isolated points in parameter space, islands, or connected regions?
This relates to 
mode connectivity, lottery tickets~\cite{frankle2020linear},
and the hypothesis that good flat minima occupy a large volume~\cite{huang2020understanding}.

\item
We proposed \textbf{three quantifiable facets of inductive biases}.
Much is missed about the ``shape'' of functions preferred by different
activations (Figure~\ref{figFctMapsCloseUp}).
An extension 
could discover other reasons for the success of NNs
and fundamental properties shared across real-world datasets.

\item 
An application of our findings
is in the control of inductive biases
to nudge the behaviour of trained networks~\cite{zhang2023instilling}.
\textbf{Update (05-2025):} our follow-up work~\cite{teney2025we} studies the learning of task-specific activation functions.



\end{itemize}



{
    \small
    \bibliographystyle{ieeenat_fullname}
    \bibliography{main}
}

\clearpage
\setcounter{page}{1}
\appendix
\maketitlesupplementary

\section{Additional Related Work}
\label{sec:relatedWorkShort}

A vast literature examines the success of deep learning
using inductive biases of optimization methods
(\eg~SGD~\cite{neyshabur2014search})
and 
architectures
(\eg~CNNs~\cite{cohen2016inductive}, transformers~\cite{zhou2023algorithms}).
This paper instead examines \emph{implicit} inductive biases in \emph{unstructured} architectures.

\paragraph{Parametrization of NNs.}
It is challenging to understand the structure and ``effective dimensionality'' of the weight space of NNs
because multiple weight configurations and their permutations correspond to the same function~\cite{entezari2021role,stock2022property,zhou2023permutation}.
A recent study quantified the information needed to identify a model with good generalization~\cite{boopathy2023model}.
However, the estimated values are astronomical
(meaning that no dataset would ever be large enough to learn the target function).
Our work reconciles these results
with the reality (the fact that deep learning does work in practice)
by showing 
that the overlap of the set of good generalizing functions
with uniform samples in \emph{weight space}
\cite[Fig.~1]{boopathy2023model}
is much denser than its overlap with \emph{truly random} functions.
In other words, random sampling in weight space
generally yields functions
likely to generalize. Much less information is needed to pick one solution among those than estimated in~\cite{boopathy2023model}.

Some think that ``\textit{stronger inductive biases come at the cost of decreasing the universality 
of a model}''~\cite{deletang2022neural}. 
This is a misunderstanding of the role of inductive biases:
they are fundamentally necessary for machine learning
and they do not imply a restriction on the set of learnable functions.
We show in particular that MLPs have strong inductive biases yet remain universal.


\paragraph{The simplicity bias}
refers to the observed tendency of NNs to fit their training data with simple functions.
It is desirable when it prevents overparametrized networks from overfitting the training data~\cite{arpit2017closer,poggio2018theory}.
But it is a curse when it causes shortcut learning~\cite{geirhos2020shortcut,teney2021evading}.
Most papers on this topic are about trained networks,
hence they confound the inductive biases of the architectures and of the optimization.
Most explanations of the simplicity bias
involve loss functions~\cite{pezeshki2021gradient}
and gradient descent
\cite{arora2019implicit,hermann2020shapes,lyu2021gradient,tachet2018learning}.

Work closer to ours~\cite{de2019random,mingard2019neural,valle2018deep} examines the simplicity bias of networks with \textbf{random weights}.
These studies are limited to MLPs with binary inputs/outputs, ReLU activations, and/or simplicity measured as compressibility.
In contrast, we examine more architectures and other measures of complexity.
Earlier works with random-weight networks include 
\cite{raghu2017expressive,giryes2016deep,lee2017deep,matthews2018gaussian,garriga2018deep,poole2016exponential,schoenholz2016deep,pennington2018emergence}.

\citet{goldblum2023no} proposed that NNs are effective
because they combine
a simplicity bias with a flexible hypothesis space.
Thus they can represent complex functions and benefit from large datasets.
Our results also support this argument.



\paragraph{The spectral bias}~\cite{rahaman2019spectral}
or frequency principle~\cite{xu2019frequency}
is a particular form of the simplicity bias.
It refers to the observation that NNs learn low-frequency components of the target function earlier during training.%
\footnote{\textbf{Frequencies of the target function}, used throughout this paper,
should not be confused with frequencies of the input data. For example, high
frequencies in images correspond to sharp edges.
High frequencies in the target function correspond to frequent changes of label for similar images.
A low-frequency target function means that similar inputs usually have similar labels.}
Works on this topic are specific to gradient descent
\cite{soudry2018implicit,xu2019training}.
and often to ReLU networks~\cite{hong2022activation,huh2021low,zhang2023shallow}.
Our work is about properties of architectures independent of the training.


Work closer to ours~\cite{yang2019fine} has noted that the spectral bias exists with ReLUs but not with sigmoidal activations, and that it depends on weight magnitudes and depth
(all of which we also observe in our experiments).
Their analysis uses the neural tangent kernel (NTK) whereas we use a Fourier decomposition of the learned function, which is arguably more direct and intuitive. We also examine other notions of complexity, and other architectures.

In work concurrent to ours, \citet{abbe2023generalization} used Walsh decompositions (a variant of Fourier analysis suited to binary functions) to characterize learned binary classification networks.
They also propose that typical NNs preferably fit low-degree basis functions to the training data and this explains their generalization capabilities.
Their discussion, which focuses on classification tasks, is highly complementary to ours.


The ``deep image prior''~\cite{ulyanov2018deep} is an image processing method that exploit the inductive biases of an untrained network.
However it specifically relies on convolutional (U-Net) architectures, whose inductive biases have little to do with those studied in this paper.

\paragraph{Measures of complexity.}
Quantifying complexity is an open problem in the fundamental sciences.
Algorithmic information theory (AIT) and Kolmogorov complexity are
one formalization of this problem.
Kolmogorov complexity has been proposed as an explicit regularizer to train NNs by
\citet{schmidhuber1997discovering}.
\citet{dingle2018input} used AIT to
explain the prevalence of simplicity in the real-world
with examples in biology and finance.
Building on this work, \citet{valle2018deep} showed that binary ReLU networks with random weights have a similar bias for simplicity.
Our work extends this line of inquiry to continuous data, to other architectures, and to other notions of complexity.

Other measures of complexity for to machine learning models
include four related notions: sensitivity, Lipschitz constant, norms of input gradients, and Dirichlet energy~\cite{dherin2022neural}.
\citet{hahn2021sensitivity} adapted ``sensitivity'' to the discrete nature of language data to measure the complexity of language classification tasks and of models.


\paragraph{Simplicity bias in transformers.}
\citet{zhou2023algorithms} explain generalization of transformer models on toy reasoning tasks using a transformer-specific measure of complexity. They propose that the function learned by a transformer corresponds to the shortest program (in a custom programming language) that could generate the training data.
\citet{bhattamishra2022simplicity} showed that transformers are more biased for simplicity than LSTMs.

\paragraph{Controlling inductive biases.}
Recent work has investigated how to explicitly tweak the inductive biases of NNs
through learning objectives~\cite{teney2021evading,teney2022predicting}
and architectures~\cite{choraria2022spectral,tancik2020fourier}.
Our results confirms that the choice of \textbf{activation function} is critical~\cite{dubey2022activation}.
Most studies on activation functions focus on individual neurons~\cite{schoenholz2016deep}
or compare the generalization properties of entire networks~\cite{oostwal2021hidden}.
\citet{francazi2023initial} showed that some activations cause a model at initialization to have non-uniform preference over classes.
\citet{simon2022reverse} showed that the behaviour of a deep MLP
can be mimicked by a single-layer MLP with a specifically-crafted activation function.


\paragraph{Implicit neural representations} (INRs) are an
application of NNs with a need to control their spectral bias.
An INR is a regression network trained to represent \eg one specific image by mapping image coordinates to pixel intensities (they are also known as \emph{neural fields} or \emph{coordinate MLPs}).
To represent sharp image details, a network must represent a high-frequency function,
which is at odds with the low-frequency bias of typical architectures.
It has been found that replacing ReLUs with periodic functions~\cite[Sect.~5]{xie2022neural},
Gaussians~\cite{ramasinghe2022beyond}, or wavelets~\cite{saragadam2023wire}
can shift the spectral bias towards higher frequencies~\cite{ramasinghe2022frequency}.
Interestingly, such architectures (Fourier Neural Networks) were investigated
as early as 1988~\cite{gallant1988there}.
Our work shows that several findings about INRs are also relevant to general learning tasks.

\section{Why Study Random-Weight Networks?}

A motivation can be found in prior work that argued
for interpreting the inductive biases of an architecture as a prior over functions
that plays in the training of the model by gradient descent.

\citet{mingard2019neural} and \citet{valle2018deep} argued that the probability of
sampling certain functions upon random sampling in parameter space
could be treated as a prior over functions for Bayesian inference.
They then presented preliminary empirical evidence that training with SGD
does approximate Bayesian inference,
such that the probability of landing on particular solutions is proportional to their
prior probability when sampling random parameters.

\section{Formal Statement of the NRS}
\label{app:formal}

We denote with
\begin{itemize}[itemsep=5pt,topsep=3pt]

\item
$F$: the \textbf{target function} we want to learn;

\item
$f_\btheta$: a chosen \textbf{neural architecture} with parameters $\btheta$;

\item
$f^\star \!:=\! f_{\btheta^\star}$:
a \textbf{trained network}
with $\btheta^\star$ optimized s.t. $f^\star$~approximates~$F$;

\item
$\bar{f} \!:=\! f_{\bar{\btheta}}, ~~
\bar{\btheta} \!\!\sim\!\! p_{\text{prior}}(\btheta)$:
an \textbf{untrained random-weight network} with parameters drawn from an uninformed prior, such as the uniform distribution used to initialize the network prior to gradient descent.

\item
$\operatorname{C}(f)$:
a scalar estimate the of \textbf{complexity} of the function~$f$
as proposed in Section~\ref{sec:complexityMeasures};

\item
$\operatorname{perf}(f)$:
a scalar measure of \textbf{generalization performance} \ie how well $f$ approximates $F$,
for example the accuracy on a held-out test set.

\end{itemize}\vspace{11pt}

\noindent
The Neural Redshift (NRS) makes three propositions.
\begin{enumerate}[itemsep=10pt,topsep=5pt]

    \item
    \textbf{NNs are biased to implement functions of a particular level of complexity determined by the architecture.}

    \item
    \textbf{This preferred complexity is observable in networks with random weights from an uninformed prior.}
    \vspace{4pt}

    Formally, $\forall$ architecture $f$, distribution $p_{\text{prior}}(\btheta)$,\\
    $\exists$ preferred complexity $c\in\mathbb{R}$ s.t.
    \setlength{\abovedisplayskip}{3pt}
    \setlength{\belowdisplayskip}{3pt}
    \begin{align*}
        \operatorname{C}(\bar{f}) &= c ~~~~~~\textrm{with very high probability, and}\\
        \operatorname{C}(f^\star) &= g(c) ~\textrm{with}
        ~~g:\mathbb{R}\rightarrow\mathbb{R}
        ~~\textrm{a monotonic function.}
    \end{align*}
    This means that the choice of architecture shifts the complexity of the learned function up or down
    similarly as it does an untrained model's.
    The precise shift is usually not predictable because $g(\cdot)$ is unknown.

    \item
    \textbf{Generalization occurs when the preferred complexity of the architecture matches the target function's.}
    \vspace{4pt}

    Formally, given two architectures $f_1$, $f_2$
    with preferred complexities $c_1$, $c_2$,
    the one with a complexity closer to the target function's
    achieves better generalization:
    \setlength{\abovedisplayskip}{3pt}
    \setlength{\belowdisplayskip}{6pt}
    \begin{align*}
        |\operatorname{C}(F) - g(c_1)|
        \,~&<~\,
        |\operatorname{C}(F) - g(c_2)|\\
        \Longrightarrow~~
        \operatorname{perf}(f_1^\star)
        \,~&>~\,
        \operatorname{perf}(f_2^\star)
        \,.
    \end{align*}

    \textbf{For example, ReLUs are popular
    because their low-complexity bias often aligns with the target function.}
\end{enumerate}

\clearpage

\section{Technical Details}
\label{app:measures}
\label{app:expDetails}

\begin{figure*}[bhtb!]
  \renewcommand{\tabcolsep}{0.4em}
  \small
  \begin{tabularx}{\linewidth}{ccccccc}
    \includegraphics[width=.130\linewidth]{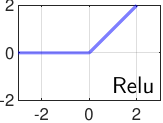}&
    \includegraphics[width=.130\linewidth]{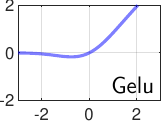}&
    \includegraphics[width=.130\linewidth]{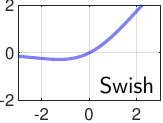}&
    \includegraphics[width=.130\linewidth]{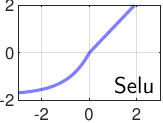}&
    \includegraphics[width=.130\linewidth]{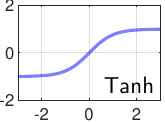}&
    \includegraphics[width=.130\linewidth]{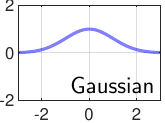}&
    \includegraphics[width=.130\linewidth]{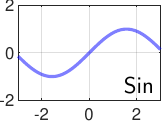}\\
    \scriptsize$\operatorname{max}(0, x)$&
    \scriptsize$x . P_\mathrm{Gaussian}(X\!\le\!x)$&
    \scriptsize$x . \sigma(x)$&
    \scriptsize$\lambda 
         \begin{cases}
            \alpha \, (e^x\!-\!1), &\!\!\!x\!\leq\!0\\
            x, &\!\!\!x\!>\!0
         \end{cases}$&
    \scriptsize$\operatorname{tanh}(x)$&
    \scriptsize$e^{(-0.5 \, x^2)}$&
    \scriptsize$\operatorname{sin}(x)$
  \end{tabularx}\vspace{-6pt}
  \caption{\label{figActivationFunctions}
  Activation functions used in our experiments.\vspace{-12pt}}
\end{figure*}

\paragraph{Activation functions.}
See Figure~\ref{figActivationFunctions} for a summary of the activations used in our experiments
and \cite{dubey2022activation} for a survey.

\paragraph{Discrete network evaluation.}
For a given network that implements the function $f(\bx)$ of input $\bx \in \mathbb{R}^d$,
we obtain obtain a discrete representation as follows.
We define a sequence of points $\bX_\mathrm{grid}\!=\!\{\bx_i\}_{i=1}^{m^d}$
corresponding to a regular grid on the $d$-dimensional hypercube $[-1,1]^d$, with $m$ values in each dimension ($m\shortteq64$ in our experiments) hence $m^d$ points in total.
We evaluate the network on every point. This gives the sequence of scalars
$\bY_f\!=\!\{
f(\bx_i)\!: \bx_i \!\in\! \bX_\mathrm{grid}\}$.

\paragraph{Visualizations as grayscale images.}
For a network $f$ with 2D inputs ($d\shortteq 2$)
we produce a visualization as a grayscale image as follows.
The values in $\bY_f$ are simply scaled and shifted to fill the range from black ($0$) to white ($1$) as:\\[2pt]
\centerline{$\tilde{\bY} = (\bY\!-\!\operatorname{min}(\bY)) \,/\, (\operatorname{max}(\bY)\!-\!\operatorname{min}(\bY))$.}\\[1pt]
We then reshape $\tilde{\bY}$ into an $m\times m$ square image.

\paragraph{Measures of complexity.}
We use our measures of complexity based on Fourier and polynomial decompositions
only with $d\shortteq 2$ because of the computational expense.
These methods first require an evaluation of the network on a discrete grid as described above ($\bY_f$)
whose size grows exponentially in the number of dimensions $d$.

\citet{xu2019frequency} proposed two 
approximations for Fourier analysis in higher dimensions.
They were not used in our experiments but could be valuable for extensions of our work to higher-dimensional settings.

\paragraph{Fourier decomposition.}
To compute the measure of complexity
$\operatorname{C}_\text{Fourier}(f)$,
we first precompute values of $f$ on a discrete grid $\bX_\mathrm{grid}$, yielding $\bY_f$  as describe above.
We then perform a standard discrete Fourier decomposition
with these precomputed values. We get:\\[2pt]
\centerline{$\tilde{f}(\bk) = \Sigma_{\bx\in\bX_\mathrm{grid}} \,\omega^{\bx^\intercal \bk} \, f(\bx)$}\\[2pt]
where $\omega\!=\!e^{-2 \pi i / m}$
and 
$\bk \in \mathbb{Z}^d$ are discrete frequency numbers.
Per the Nyquist-Shannon theorem, with an evaluation of $f$ on a grid of $m$ values in each dimension, we can reliably measure the energy for frequency numbers up to $m/2$ in each dimension
\ie for $\bk \in \bK \!=\! [ 0, \lldots, m/2 ]^d$.

The value $\tilde{f}(\bk)$ is a complex number that captures both the magnitude and phase of the $\bk$th Fourier component.
We do not care about the phase, hence our measure of complexity only uses
the real magnitude $|\tilde{f}(\bk)|$ of each Fourier component $\bk$.
We then seek to summarize the distribution of these magnitudes across frequencies
into a single value.
We define the measure of complexity:\\[2pt]
\centerline{$\operatorname{C}_\text{Fourier}(f) ~=~ \Sigma_{\bk \in \bK} \, |\tilde{f}(\bk)| \,.\, ||\bk||_2 \;~/\;~ \Sigma_{\bk \in \bK} \, |\tilde{f}(\bk)|$.}\\[2pt]
This is the average of magnitudes, weighted each by the corresponding frequency, disregarding orientation (\eg horizontal and vertical patterns in a 2D visualization of the function are treated similarly), and normalized such that magnitudes sum to~$1$.

See~\cite{rahaman2019spectral} for a technical discussion justifying Fourier analysis on non-periodic bounded functions.

\paragraph{Limitations of a scalar measure of complexity.}
The above definition is necessarily imperfect at summarizing the distributions of magnitudes across frequencies. For example,
an $f$ containing both low and high-frequencies
could receive the same value
as one containing only medium frequencies.
In practice however, we use this complexity measure on random networks,
and we verified empirically that the distributions of magnitudes are always unimodal.
This summary statistic is therefore a reasonable choice to compare distributions.

\paragraph{Polynomial decomposition.}
As an alternative to Fourier analysis,
we use decomposition in polynomial series.%
\footnote{See \eg \url{https://www.thermopedia.com/content/918/}.}
It uses a predefined set of polynomials $P_n(x)$, $n = [0, \lldots, N]$ to approximate
a function $f(x)$ on the interval $x \in [-1,1]$ as
$f(x) \!\approx\! \Sigma_{c=0}^N \, c_n P_n(x)$.
The coefficients 
are calculated as
$c_n \!=\! 0.5 \, (2n+1) \, \int_{-1}^{+1} f(x) \, P_n(x) \,dx$.
These definitions readily extends to higher dimensions.

In a Fourier decomposition, the coefficients indicate the amount 
the various frequency components in $f$.
Here, each coefficient $c_n$ indicates the amount of a component of a certain order.
In 2 dimensions ($d\shortteq2$), we have $N^2$ coefficients $c_{00}, c_{01}, \lldots, c_{NN}$.
We define our measure of complexity:\\[2pt]
\centerline{$\operatorname{C}_\text{Chebyshev}(f) ~=~ \frac{\Sigma_{n_1,n_2\shorteq0}^N \, |c_{n_1n_2}| \,.\, || \,[n_1, n_2]\, ||_2}{\Sigma_{n1,n2\shorteq0}^N \, |c_{n1,n2}|}$.}\\[2pt]
This definition is nearly identical to the Fourier one.

In practice, we experimented with Hermite, Legendre, and Chebyshev bases of polynomials.
We found the latter to be more numerically stable.
To compute the coefficients, we use
trapezoidal numerical integration
and the same
sampling of
$f$ on $\bX_\mathrm{grid}$ as described above, and a maximum order $N\shortteq100$.
To make the evaluation of the integrals more numerically stable (especially with Legendre polynomials),
we omit a border near the edges of the domain $[-1,1]^d$.
With a $64\times64$ grid, we omit 3 values on every side.

\paragraph{LZ Complexity.}
We use the compression-based measure of complexity described in~\cite{dingle2018input,valle2018deep}
as an approximation of the Kolmogorov complexity.
We first evaluate $f$ on a grid to get $\bY_f$ as described above.
The values in $\bY_f$ are reals and generally unique,
so we discretize them on a coarse scale of $10$ values regularly spaced in the range of $\bY_f$
(the granularity of $10$ is arbitrary can be set much higher with virtually no effect if $\bY_f$ is large enough).
We then apply the classical Lempel–Ziv compression algorithm on the resulting number sequence.
The measure of complexity 
$\operatorname{C}_\text{LZ}(f)$
is then defined as the size of the dictionary built by the compression algorithm.
The LZ algorithm is sensitive to the order of the sequence to compress,
but we find very little difference across different orderings of $\bY_f$ (snake, zig-zag, spiral patterns).
Thus we use a simple column-wise vectorization of the 2D grid.

In higher dimensions (Colored-MNIST experiments),
it would be computationally too expensive to define
$\bX_\mathrm{grid}$
as a dense sampling of the 
full hypercube $[-1,1]^d$ (since $d$ is large).
Instead, we randomly pick $m$ corners of the hypercube and sample $m$ points $\bx_i$ regularly between each successive pair of them. This gives a total of $m^2$ points corresponding to random linear traversals of the input space.
Instead of feeding $\bY_f$ directly to the LZ algorithm, we also feed it with successive differences between successive values, which we found to improve the stability of the estimated complexity (for example, the pixel values $10,12,15,18$ are turned into $2,3,3$).

\paragraph{LZ Complexity with transformers.}
These experiments use $\operatorname{C}_\text{LZ}(f)$
on sequences of tokens.
Each token is represented by its index in the vocabulary, and the LZ algorithm is directly applied on these sequences of integers.

\paragraph{Absolute complexity values.}
The different measures of complexity have different absolute scales and no comparable units.
Therefore, for each measure, we rescale the values such that observed values fill the range $[0,1]$.

\paragraph{Unbiased model.}
We construct an architecture
that displays no bias for any frequency in a Fourier decomposition of the functions it implements.
This architecture
$f_\btheta(\cdot)$
implements an inverse discrete Fourier transform
with learnable parameters $\btheta=\{\btheta_\textrm{mag}, \,\btheta_\textrm{phase}\}$
that correspond to the magnitude and phase of each Fourier component.
It can be implement as a one-hidden-layer MLP with sine activation, fixed input weights (each channel defining the frequency of one Fourier component), learnable input biases (the phase shifts), and learnable output weights (the Fourier coefficients).

\paragraph{Experiments with modulo addition.}
These experiments use a 4-layer MLP of width 128.
We train them with full-batch Adam, a learning rate 0.001, for 3k iterations with no early stopping.
Each experiment is run with 5 random seeds. The 
Figure~\ref{figModulo} shows the average over seeds for clarity (each point corresponds to a different architecture).
Figure~\ref{figModuloGaussian} shows all seeds (each point corresponds to a different seed).

\paragraph{Experiments on Colored-MNIST.}
The dataset is built from the MNIST digits, keeping the original separation between training and test images.
To define a regression task, we turn the original classification labels $\{0,1,\lldots,9\}$
into values in $[0,1]$.
To introduce a spurious feature, each image is concatenated with a column of pixels of uniform grayscale intensity (the ``color'' of the image).
This ``color'' is directly correlated with the label
with some added noise to simulate a realistic spurious feature:
in 3\% of the training data, the color is replaced with a random one.

The models are 2-layer MLPs of width 64.
They are trained with an MSE loss with full-batch Adam, learning rate 0.002, 10k iterations with no early stopping.
The ``accuracy'' in our plots is actually: $1\!-\!\textrm{MAE}$ (mean average error).
Since this is a regression task with test labels distributed uniformly in $[0,1]$,
this metric is indeed interpretable as a binary accuracy, with $0.5$ equivalent to random chance.


\paragraph{Experiments with transformers.}
In all experiments described above,
we directly examine the input$\,\xrightarrow{}\,$output mappings implemented by neural networks.
In the experiments with transformer sequence models, we examine
sequences generated by the models.
These models are autoregressive, which means that the function they implement is the mapping
context$\,\xrightarrow{}\,$next token.
We expect a simple function (\eg low-frequency)
to produce lots of repetitions in sequences sampled auto-regressively.
(language models are indeed known to often repeat themselves~\cite{holtzman2019curious,fu2021theoretical}).
Such sequences are highly compressible.
They should therefore give a low values of $\operatorname{C}_\text{LZ}$.


\clearpage
\section{Additional Experiments with\\Trained Models}
\label{app:trained}
\label{app:otherWeightDistributions}

This section presents experiments with models trained with standard gradient descent.
We will show that there is a correlation between the complexity of a model at initialization
(\ie with random weights)
and that of a trained model of the same architecture.

\begin{figure}[t]
  \renewcommand{\tabcolsep}{.5em}
  \renewcommand{\arraystretch}{1}
  \begin{tabular}{rcc}
  &\small Cameraman & \small Waves\\[.2em]
  \makecell[r]{\small Full\\\small data}&
  \raisebox{-0.5\height}{\includegraphics[width=7em]{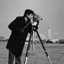}}&
  \raisebox{-0.5\height}{\includegraphics[width=7em]{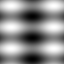}}\\
  \makecell[r]{\small Training\\\small points}&
  \raisebox{-0.5\height}{\includegraphics[width=7em]{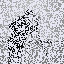}}&
  \raisebox{-0.5\height}{\includegraphics[width=7em]{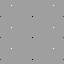}}
  \end{tabular}\vspace{-4pt}
  \caption{\label{figGt}
  Data used in the coordinate-MLP experiments.
  \vspace{-12pt}
  }
\end{figure}

\paragraph{Setup with coordinate-MLPs.}
The experiments in this section use models
trained as implicit neural representations of images (INRs), also known as coordinate-MLPs~\cite{xie2022neural}.
Such a model is trained to represent a specific grayscale image,
It takes as input 2D coordinates in the image plane
$\bx\!\in\![-1,1]^2$.
It produces as output the scalar grayscale value of the image at this location.
The ground truth data is a chosen image (Figure~\ref{figGt}).
For training, we use a subset of pixels.
For testing, we evaluate the network on a $64\!\times\!64$ grid,
which directly gives a $64\!\times\!64$ pixel representation of the function learned.

\begin{mdframed}[style=citationFrame,userdefinedwidth=\linewidth,align=left,skipabove=6pt,skipbelow=0pt]
\textbf{Why use coordinate-MLPs?}
This setup produces
interpretable visualizations
and allows comparing visually the ground truth (original image) with the learned function.
Because the ground truth is defined on a regular grid (unlike most real data)
it also facilitates the computation of 2D Fourier transforms.
We use Fourier transforms to quantitatively compare
the ground truth with the learned function
and verify the third part of the NRS
(generalization is enabled by matching of the architecture's preferred complexity with the target function's).
\end{mdframed}

\paragraph{Data.}
We use a $64\times64$ pixel version of
the well-known \textit{cameraman} image (Figure~\ref{figGt}, left)
\cite{gilkes1974photograph}.
For training, we use a random $40\%$ of the pixels.
This image contains both uniform areas (low frequencies)
and fine details with sharp transitions (high frequencies).
We also use a synthetic \textit{waves} image (Figure~\ref{figGt}, right).
It is the sum of two orthogonal sine waves, one twice the frequency of the other.
For training, we only use pixels on local extrema of the image.
They form a very sparse set of points.
This makes the task severely underconstrained.
A model can fit this data with a variety functions.
This will reveal whether a model prefers fitting low- or high-frequency patterns.

\subsection{Visualizing Inductive Biases}
\label{app:expVis}

\begin{figure*}[p]
  \thisfloatpagestyle{empty} 
  \renewcommand{\tabcolsep}{0.5em}
  \renewcommand{\arraystretch}{1.1}
  \small
  \centering
  \begin{tabular}{ccc}
    \toprule
    ~ & At init. & \raisebox{-0.5\height}{\includegraphics[width=16em]{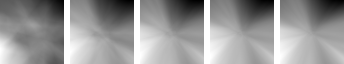}}\\
    \textbf{ReLU} & Trained & \raisebox{-0.5\height}{\includegraphics[width=16em]{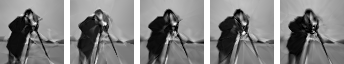}}\\
    ~ & Shuffled & \raisebox{-0.5\height}{\includegraphics[width=16em]{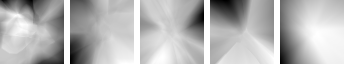}}\\
    \midrule

    ~ & At init. & \raisebox{-0.5\height}{\includegraphics[width=16em]{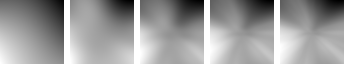}}\\
    \textbf{GELU} & Trained & \raisebox{-0.5\height}{\includegraphics[width=16em]{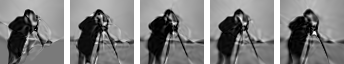}}\\
    ~ & Shuffled & \raisebox{-0.5\height}{\includegraphics[width=16em]{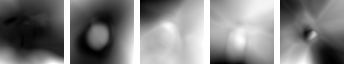}}\\
    \midrule

    ~ & At init. & \raisebox{-0.5\height}{\includegraphics[width=16em]{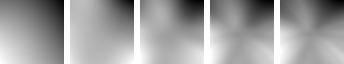}}\\
    \textbf{Swish} & Trained & \raisebox{-0.5\height}{\includegraphics[width=16em]{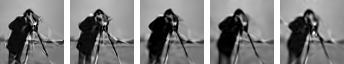}}\\
    ~ & Shuffled & \raisebox{-0.5\height}{\includegraphics[width=16em]{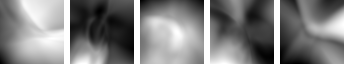}}\\
    \midrule

    ~ & At init. & \raisebox{-0.5\height}{\includegraphics[width=16em]{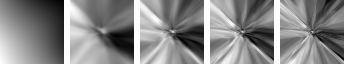}}\\
    \textbf{TanH} & Trained & \raisebox{-0.5\height}{\includegraphics[width=16em]{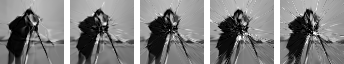}}\\
    ~ & Shuffled & \raisebox{-0.5\height}{\includegraphics[width=16em]{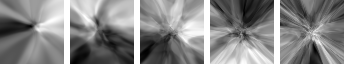}}\\
    \midrule

    ~ & At init. & \raisebox{-0.5\height}{\includegraphics[width=16em]{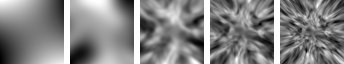}}\\
    \textbf{Gaussian} & Trained & \raisebox{-0.5\height}{\includegraphics[width=16em]{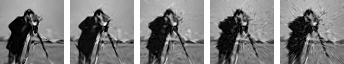}}\\
    ~ & Shuffled & \raisebox{-0.5\height}{\includegraphics[width=16em]{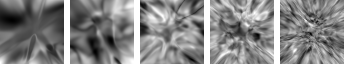}}\\
    \midrule

    ~ & At init. & \raisebox{-0.5\height}{\includegraphics[width=16em]{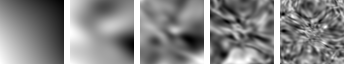}}\\
    \textbf{Sin} & Trained & \raisebox{-0.5\height}{\includegraphics[width=16em]{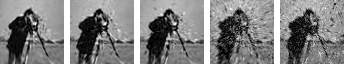}}\\
    ~ & Shuffled & \raisebox{-0.5\height}{\includegraphics[width=16em]{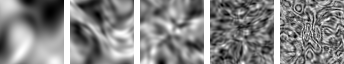}}\\
    \midrule

    ~ & At init. & \raisebox{-0.5\height}{\includegraphics[width=16em]{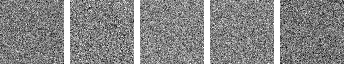}}\\
    \textbf{Unbiased} & Trained & \raisebox{-0.5\height}{\includegraphics[width=16em]{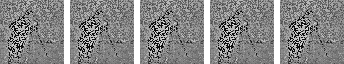}}\\
    ~ & Shuffled & \raisebox{-0.5\height}{\includegraphics[width=16em]{figTrained/v-cameraman-reconstructFromCoefs-fctMaps2.png}}\\
    \bottomrule

    ~&~& \scriptsize (Default) \hspace{.85em} Magnitude of initial weights $\rightarrow$ \hspace{.85em}(Larger)
  \end{tabular}\vspace{-4pt}
  \caption{\label{figFctMapsLearnedCameraman}
  Coordinate-MLPs trained to represent the \textit{cameraman}
  with various activations and initial weight magnitudes.
  The model is trained on $40\%$ pf pixels and evaluated on a $64\times64$ grid.
  The images provide intuitions about the inductive biases of each architecture.
  The differences across models with random weights
  (at init.) and with shuffled trained weights (shuffled)
  show that the increase in complexity in non-ReLU models is
  realized by changes in weight magnitudes (which are maintained through the shuffling).
  In contrast, ReLU networks revert to a low complexity
  after shuffling, suggesting that complexity is encoded in the precise weight values,
  not their magnitudes.
  \vspace{-8pt}}
  \thispagestyle{empty}
\end{figure*}

We first perform experiments to get a visual intuition
of the inductive biases provided by different activation functions.
We train 3-layer MLPs of width 64 with full-batch Adam and a learning rate of 0.02
on the cameraman and waves data.
Figure~\ref{figFctMapsLearnedCameraman} (next page)
shows very different functions across architectures.
The cameraman image contains fine details with sharp edges.
Their presence in the reconstruction indicate whether the model learned high-frequency components.

\begin{figure*}[t!]
  \thisfloatpagestyle{empty} 
  \renewcommand{\tabcolsep}{0.3em}
  \renewcommand{\arraystretch}{1}
  \small
  \centering
  \begin{tabular}{rc}
    \toprule
    \multicolumn{2}{c}{
    \textbf{Full data}~
    \raisebox{-0.4\height}{\includegraphics[width=4em]{figTrained/gt-shapes-T3.png}}
    \hspace{0.7em}
    \textbf{Sparse training points}~
    \raisebox{-0.4\height}{\includegraphics[width=4em]{figTrained/shapes-T3-fctMapsTr2.png}}
    }\\
    \midrule
    \textbf{ReLU} & \raisebox{-0.4\height}{\includegraphics[width=20em]{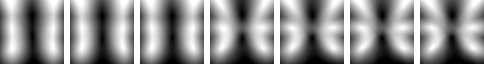}}\\[1.0em]
    \textbf{GELU} & \raisebox{-0.4\height}{\includegraphics[width=20em]{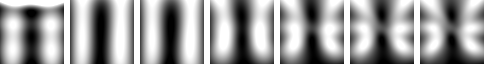}}\\[1.0em]
    \textbf{Swish} & \raisebox{-0.4\height}{\includegraphics[width=20em]{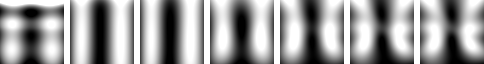}}\\[1.0em]
    \textbf{TanH} & \raisebox{-0.4\height}{\includegraphics[width=20em]{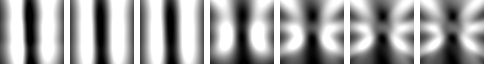}}\\[1.0em]
    \textbf{Gaussian} & \raisebox{-0.4\height}{\includegraphics[width=20em]{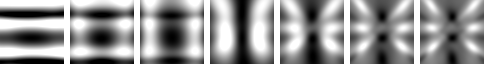}}\\[1.0em]
    \textbf{Sine} & \raisebox{-0.4\height}{\includegraphics[width=20em]{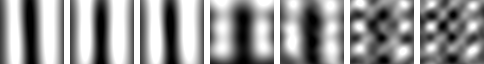}}\\
    \bottomrule
    ~& \scriptsize ~0.4 \hspace{1.6em} ~0.8 \hspace{1.6em} ~1.0 \hspace{1.6em} ~7.0 \hspace{1.6em} 13.0 \hspace{1.6em} 19.0 \hspace{1.6em} 22.0\\[-1pt]
    ~& Magnitude of initial weights (fraction of standard magnitude)\\
  \end{tabular}\vspace{-4pt}
  \caption{\label{figFctMapsLearnedWaves}
  Coordinate-MLPs trained on sparse points
  of the \textit{waves} data.
  Variations across learned functions
  show how architectures are biased towards low (ReLU) or high frequencies (Sine).
  ReLU activations give the most consistent behaviour across weight magnitudes.
  \vspace{-8pt}}
  \thispagestyle{empty}
\end{figure*}

\paragraph{Differences across architectures.}
The \textbf{ReLU-like activations} are biased for simplicity,
hence the learned functions tend to smooth out image details,
favor large uniform regions and smooth variations.
Yet, they can also represent sharp transitions, when these are necessary to fit the training data.
The decision boundary with ReLUs, which is a polytope~\cite{montufar2014number} is faintly discernible
as criss-crossing lines in the image.
Surprisingly, we observe differences across different initial weight magnitudes with ReLU,
even though our experiments on random networks did not show any such effect (Section~\ref{sec:untrained}).
We believe that this is a sign of optimization difficulties when the initial weights are large
(\ie difficulty of reaching a complex solution).

With \textbf{other activations} (TanH, Gaussian, sine) the bias for low or high frequencies is
much more clearly modulated by the initial weight magnitude.
With large magnitudes, the images contain high-frequency patterns.
Similar observations are made with the waves data (Figure~\ref{figFctMapsLearnedWaves}).

The \textbf{unbiased model} is useless, as expected. It reaches perfect accuracy on the training data,
but the predictions on other pixels look essentially random.

\paragraph{With random weights.}
We also examine in Figure~\ref{figFctMapsLearnedCameraman} the function represented by each model at initialization (with random weights).
As expected, we observe a strong correlation between the amount of high frequencies at initialization and in the trained model.
We also examine models at the end of training, after shuffling the trained weights (within each layer).
This is another random-weight model, but its distribution of weight magnitudes matches exactly the trained model.
Indeed, the shuffling preserves the weights within each layer
but destroys the precise connections across layers.
This enables a very interesting observation.
With non-ReLU-like architectures, there is a clear increase in complexity between the functions at initialization and with shuffled weights.
This means that the learned increase in \textbf{complexity
in non-ReLU networks is
partly encoded by changes in the distribution of weight magnitudes} (the only thing preserved through shuffling).
In contrast, ReLU networks revert to a low complexity after shuffling.
This suggests that \textbf{complexity in ReLU networks is encoded
in the weights' precise values and connections across layers, not in their magnitude}.

\begin{figure*}[thp]
  \thisfloatpagestyle{empty} 
  \renewcommand{\tabcolsep}{0.5em}
  \renewcommand{\arraystretch}{1}
  \small
  \centering
  \begin{tabular}{ccc}
    \textbf{Weights}&
    \textbf{Biases}&
    \textbf{Activations}\\[3pt]
    \includegraphics[width=.18\linewidth]{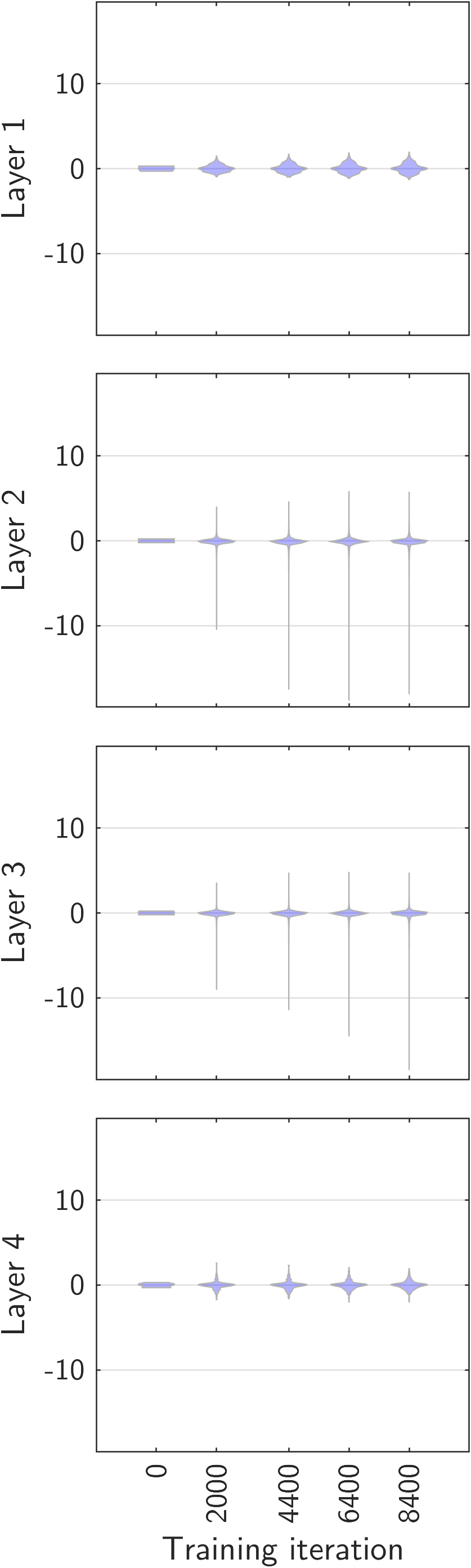}~~~~~~~~&
    \includegraphics[width=.18\linewidth]{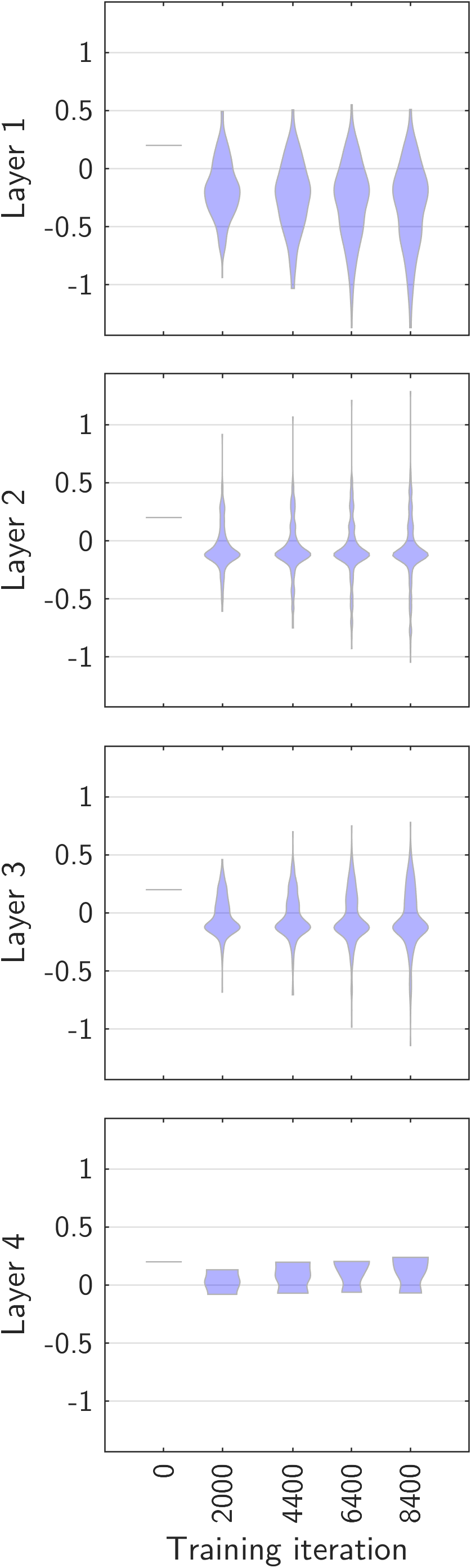}~~~~~~~~&
    \includegraphics[width=.18\linewidth]{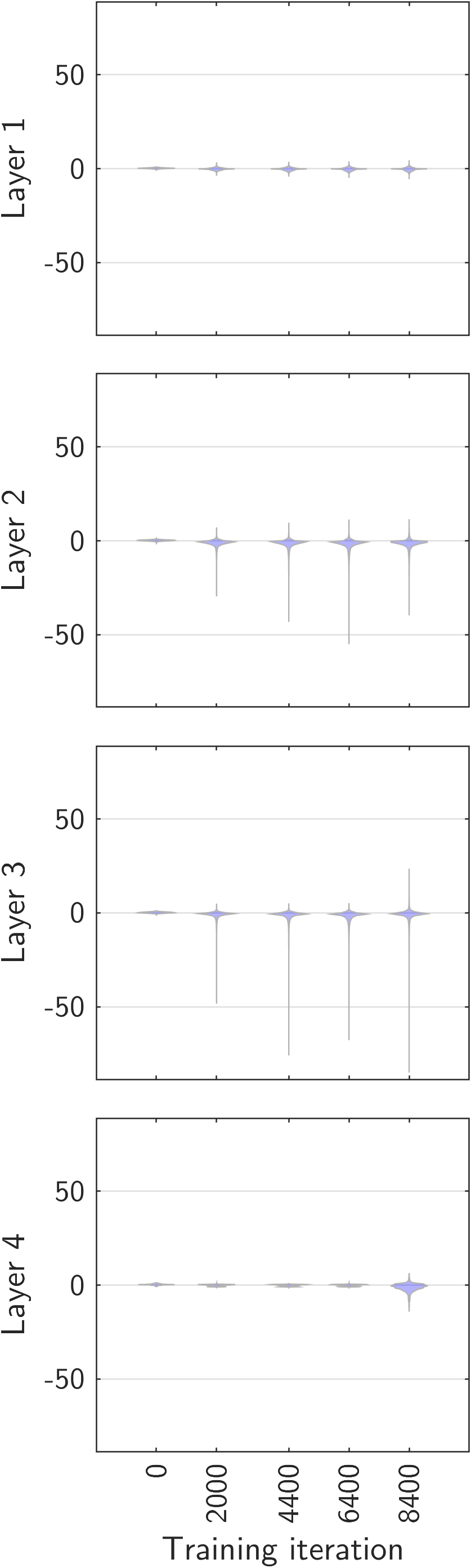}~~~~~
  \end{tabular}\vspace{-4pt}
  \caption{\label{figViolin}
  Distributions of the magnitudes of the weights, biases, and activations during training
  of a 3-layer MLP (the 4th row is the output layer) on the cameraman data.
  Weights and biases are initialized from a uniform distribution and zero, respectively.
  The distributions become very long-tailed as training progresses.
  The occurrence of large values is the reason why
  the dependence of the ``preferred complexity'' of certain architecture on weight magnitudes is important (it would not matter if the distribution of magnitudes remained constant throughout training).
  \vspace{-8pt}}
\end{figure*}

\begin{figure*}[thp]
  \thisfloatpagestyle{empty} 
  \renewcommand{\tabcolsep}{0.5em}
  \renewcommand{\arraystretch}{1}
  \small
  \centering
  \begin{tabular}{cccc}
    
    &\scriptsize\textsf{ReLU} & \scriptsize\textsf{GELU} & \scriptsize\textsf{Swish}\\
\multirow{4}{*}{\rotatebox{90}{\scriptsize \textsf{Complexity (Fourier) $\longrightarrow$}}}
    &\raisebox{-0.5\height}{\includegraphics[width=.18\linewidth]{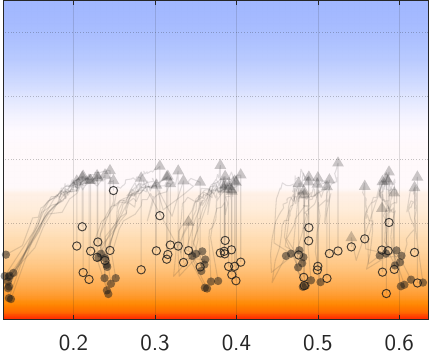}}&
    \raisebox{-0.5\height}{\includegraphics[width=.18\linewidth]{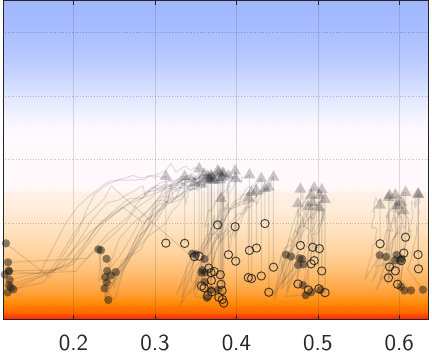}}&
    \raisebox{-0.5\height}{\includegraphics[width=.18\linewidth]{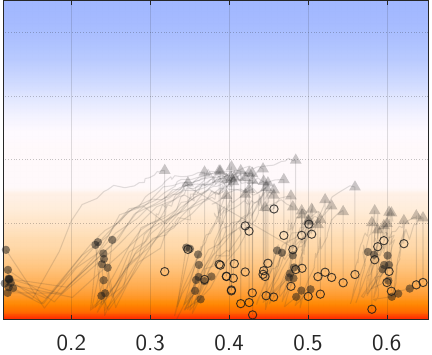}}\\[4.4em]

    &\scriptsize\textsf{TanH} & \scriptsize\textsf{Gaussian} & \scriptsize\textsf{Sine}\\[1pt]
    &\raisebox{-0.5\height}{\includegraphics[width=.18\linewidth]{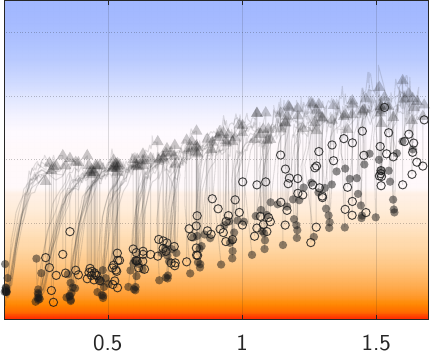}}&
    \raisebox{-0.5\height}{\includegraphics[width=.18\linewidth]{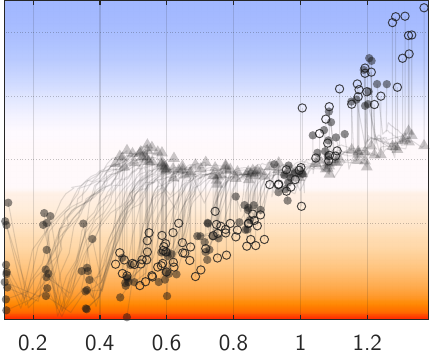}}&
    \raisebox{-0.5\height}{\includegraphics[width=.18\linewidth]{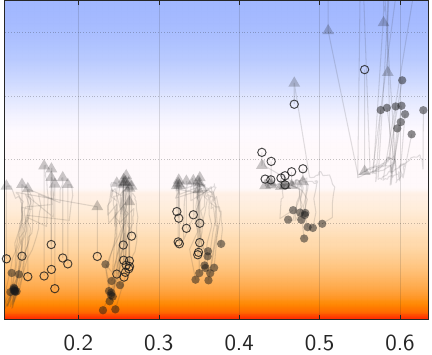}}\\[4.4em]

    &\multicolumn{3}{c}{\scriptsize\textsf{Average magnitude of weights during training}}\\[.6em]
    &\multicolumn{3}{c}{\scriptsize\textsf{
    $\bullet$ At initialization \hspace{1em}
    \textcolor{black!30}{\ding{115}} At last epoch\hspace{1em}
    $\circ$ With shuffled trained weights}}\\[3pt]
    
    \multicolumn{4}{c}{~~~~~~~~\scriptsize\textsf{\textsf{Lower frequencies\raisebox{-.3\height}{\includegraphics[clip, trim=17pt 10pt 17pt 1pt, width=.20\linewidth]{fig/heatmap-legendHorizontal.png}}Higher frequencies}}}
  \end{tabular}\vspace{-4pt}
  \caption{\label{figCameramanTraining}
  Training trajectories of MLP models trained on the cameraman data.
  Each line corresponds to one training run (with a different seed or initial weight magnitude).
  With ReLU-like activations, the models at initialization have low complexity regardless of the initialization magnitude.
  As training progresses, the complexity increases to fit the training data.
  This increased complexity is encoded in the weights precise values and connections across layers, since at the end of training, shuffling the weights reverts models to the initial low complexity.
  With other activations, the initial weight magnitude impacts the complexity at initialization and of the trained model.
  Some of the additional complexity in the trained model seems to be partly encoded by increases in the weight magnitudes, since shuffling the trained weights does seem to retain some of this additional complexity.
  \vspace{-8pt}}
\end{figure*}

\subsection{Training Trajectories}

We will now show that NNs can represent any function,
but complex ones require
precise weight values and connections across layers
that are unlikely through random sampling
but that can be found through gradient-based training.

Unlike prior work~\cite{ramasinghe2022you} that claimed that the complexity at initialization \emph{causally} influences the solution,
our results indicate instead they are two effects of a common cause
(the ``preferred complexity'' of the architecture).
The architecture is biased towards a certain complexity, and this influences both
the randomly-initialized model
and those found by gradient descent.
There exist weight values for other functions (of complexity much lower or higher than the preferred one) but they are less likely to occur in either case.

For example, ReLU networks are biased towards simplicity but can represent complex functions.
Yet, contrary to~\cite{ramasinghe2022you}, initializing gradient descent
with such a complex function does not yield a complex solutions after training on simple data.
In other words, the architecture's bias prevails over the exact starting point of the training.

\paragraph{Experimental setup.}
We train models with
different activations and initial magnitudes
on the cameraman data, using $\nicefrac{1}{9}$ pixels for training.
We plot in Figure~\ref{figCameramanTraining}
the training trajectory of each model.
Each point of a trajectory represents the 
average weight magnitude vs. the Fourier complexity of the function represented by the model.

\paragraph{Changes in weight magnitudes during training.}
The first observation is that
the average weight magnitude
changes surprisingly little.
However, further examination (Figure~\ref{figViolin})
shows that the distribution shifts from uniform to long-tailed.
The trained models contain more and more large-magnitude weights.

\paragraph{Changes in complexity during training.}
In Figure~\ref{figCameramanTraining},
we observe that
models with ReLU-like activations at initialization have low complexity regardless of the initialization magnitude.
As training progresses, the complexity increases to fit the training data.
This increased complexity is encoded in the weights' precise values and connections across layers, since at the end of training, shuffling the weights reverts models back to the initial low complexity.
With other activations, the initial weight magnitudes impact the complexity
at initialization and of the trained model.
Some of the additional complexity in the trained model seems to be partly encoded by increases
in weight magnitudes, since shuffling the trained weights does seem to retain some of this additional complexity.

\paragraph{Summary.}
The choice of activation function and initial weight magnitude
affect the ``preferred complexity'' of a model.
This complexity is visible both at initialization (with random weights)
and after training with gradient descent.
The complexity of the learned function can depart from the ``preferred level''
just enough to fit the training points.
Outside the interpolated training points,
the shape of the learned function  is very much affected by the preferred complexity.

With ReLU networks, this effect usually drives the complexity downwards (the \textbf{simplicity bias}).
With other architectures and large weight magnitudes, this often drives the complexity upwards.
Both can be useful: Section~\ref{sec:trained}
showed that sine activations can
enable learning the parity function from sparse training points, and reduce shortcut learning
by shifting the preferred complexity upwards.

Our observations also explain why 
the coordinate-MLPs with sine activations proposed in~\cite{sitzmann2020implicit} (SIREN)
require a careful initialization.
This adjusts the preferred complexity to the typical frequencies found in natural images.

\subsection{Pretraining and Fine-tuning}
\label{app:expFineTuning}
We outline preliminary results from additional experiments.

\paragraph{Why study fine-tuning?}
We have seen that the preferred complexity of an architecture can be observed with random weights.
The model can then be trained by gradient descent to represent data with a different level of complexity.
For example, a ReLU network, initially biased for simplicity, can represent a complex function
after training on complex data.
Gradient descent finely adjusts the weights
to represent a complex function.
We will now see how pretraining then fine-tuning on data with different levels of complexity ``blends'' the two in the final fine-tuned model.

\paragraph{Experimental setup.}
We pretrain an MLP with ReLU or TanH activations on high-frequency data
(high-frequency 2D sine waves).
We then fine-tune it on lower-frequency 2D sine waves of a different random orientation.

\paragraph{Observations.}
During early fine-tuning iterations, TanH models retain a high-frequency bias
much more than ReLU models.
This agrees with the proposition in~\ref{app:expVis}
that the former encode high frequencies in weight magnitudes,
while ReLU models encode them in precise weight values, which are quickly lost during fine-tuning.

We further verify this explanation by shuffling the pretrained weights (within each layer) before fine-tuning.
The ReLU models then show no high-frequency bias at all (since the precise arrangement of weights is completely lost through the shuffling).
TanH models, however, do still show high-frequency components in the fine-tuned solution.
This confirms that TanH models encode high frequencies partly in weight magnitudes
since this is the only property preserved by the shuffling.

Finally, we do not find evidence for the prior claim~\cite{ramasinghe2022you}
that complexity at initialization persists \emph{indefinitely} throughout fine-tuning.
Instead, with enough  iterations of fine-tuning,
any pretraining effect on the preferred complexity eventually vanishes.
For example, a ReLU model pretrained on high frequencies
initially contains high-frequency components in the fine-tuned model.
But with enough iterations, they eventually disappear
\ie the simplicity bias of ReLUs eventually takes over.
We believe that the experiments in~\cite{ramasinghe2022you}
were simply not run for long enough.
This observation also disproves the causal link proposed in~\cite{ramasinghe2022you}
between the complexity at initialization and in the trained model.

\clearpage
\onecolumn
\section{Full Results with Random Networks}
\label{app:heatmapsDetails}
\label{app:untrainedFullResults}

On the next pages (Figures~\ref{figHeatmapsFourierAppendix}--\ref{figHeatmapsLzAppendix}), we present heatmaps showing 
the average complexity of functions implemented by neural networks of various architectures with random weights and biases.
Each heatmap corresponds to one architecture with varying number of layers (heatmap columns)
and weight magnitudes (heatmap rows).
For every other cell of a heatmap, we visualize, as a 2D grayscale image,
a function implemented by one such a network
with 2D input and scalar output.

\begin{figure*}[h!]
  \renewcommand{\tabcolsep}{0.35em}
  \renewcommand{\arraystretch}{2}
  \small
  \begin{tabularx}{\linewidth}{cccccccccc}
    & \scriptsize\textsf{ReLU} & \scriptsize\textsf{GELU} & \scriptsize\textsf{Swish} & \scriptsize\textsf{SELU} & \scriptsize\textsf{TanH} & \scriptsize\textsf{Gaussian} & \scriptsize\textsf{Sin} & \scriptsize\textsf{Unbiased}\\[-3pt]

    \raisebox{28pt}{\rotatebox{90}{\scriptsize\textsf{Fourier}}}&
    \includegraphics[width=.09\linewidth]{figFourier/fourier-spectrumIntegerFrequencies-signed-Relu-plots.pdf}&
    \includegraphics[width=.09\linewidth]{figFourier/fourier-spectrumIntegerFrequencies-signed-Gelu-plots.pdf}&
    \includegraphics[width=.09\linewidth]{figFourier/fourier-spectrumIntegerFrequencies-signed-Swish-plots.pdf}&
    \includegraphics[width=.09\linewidth]{figFourier/fourier-spectrumIntegerFrequencies-signed-Selu-plots.pdf}&
    \includegraphics[width=.09\linewidth]{figFourier/fourier-spectrumIntegerFrequencies-signed-Tanh-plots.pdf}&
    \includegraphics[width=.09\linewidth]{figFourier/fourier-spectrumIntegerFrequencies-signed-Gaussian-plots.pdf}&
    \includegraphics[width=.09\linewidth]{figFourier/fourier-spectrumIntegerFrequencies-signed-Sin-plots.pdf}&
    \includegraphics[width=.09\linewidth]{figFourier/fourier-spectrumIntegerFrequencies-signed-reconstructFromCoefs-plots.pdf}&
    \raisebox{.20\height}{\includegraphics[width=.14\linewidth]{figUntrained/plots-legend.pdf}}\\

    \raisebox{26pt}{\rotatebox{90}{\scriptsize\textsf{Chebyshev}}}&
    \includegraphics[width=.09\linewidth]{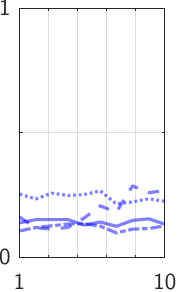}&
    \includegraphics[width=.09\linewidth]{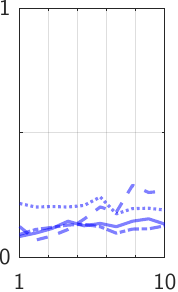}&
    \includegraphics[width=.09\linewidth]{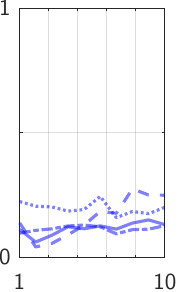}&
    \includegraphics[width=.09\linewidth]{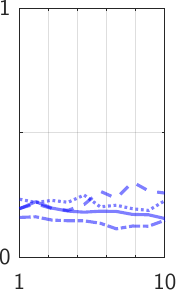}&
    \includegraphics[width=.09\linewidth]{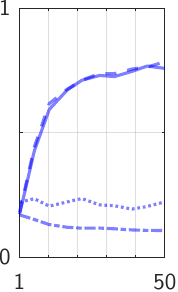}&
    \includegraphics[width=.09\linewidth]{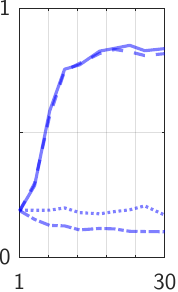}&
    \includegraphics[width=.09\linewidth]{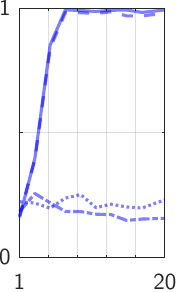}&
    \includegraphics[width=.09\linewidth]{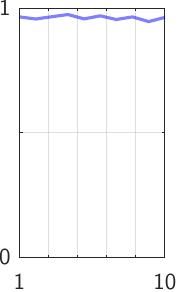}&\\

    \raisebox{26pt}{\rotatebox{90}{\scriptsize\textsf{Legendre}}}&
    \includegraphics[width=.09\linewidth]{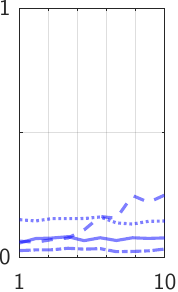}&
    \includegraphics[width=.09\linewidth]{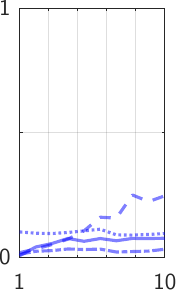}&
    \includegraphics[width=.09\linewidth]{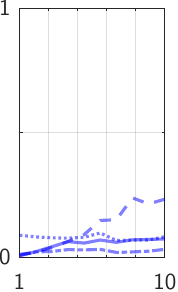}&
    \includegraphics[width=.09\linewidth]{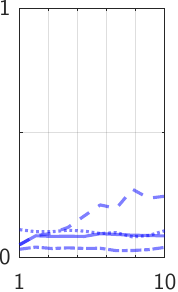}&
    \includegraphics[width=.09\linewidth]{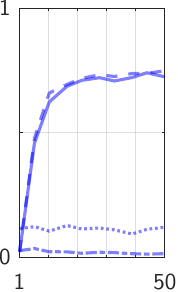}&
    \includegraphics[width=.09\linewidth]{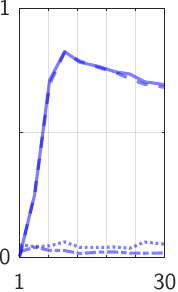}&
    \includegraphics[width=.09\linewidth]{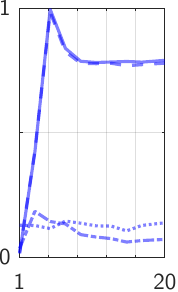}&
    \includegraphics[width=.09\linewidth]{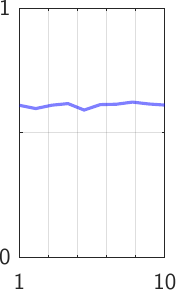}&\\

    \raisebox{34pt}{\rotatebox{90}{\scriptsize\textsf{LZ}}}&
    \includegraphics[width=.09\linewidth]{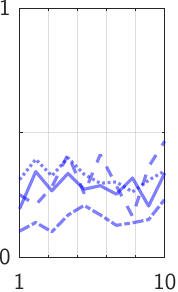}&
    \includegraphics[width=.09\linewidth]{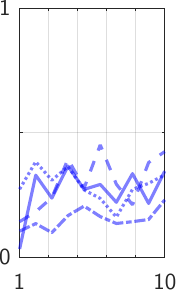}&
    \includegraphics[width=.09\linewidth]{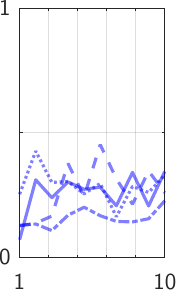}&
    \includegraphics[width=.09\linewidth]{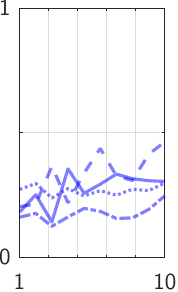}&
    \includegraphics[width=.09\linewidth]{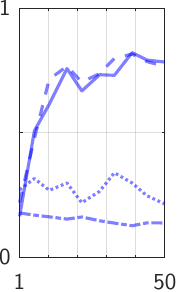}&
    \includegraphics[width=.09\linewidth]{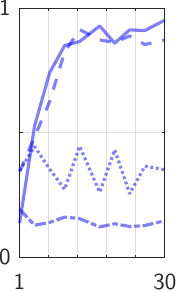}&
    \includegraphics[width=.09\linewidth]{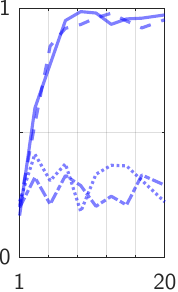}&
    \includegraphics[width=.09\linewidth]{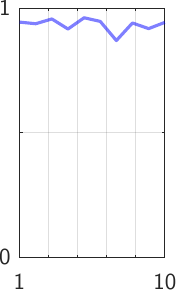}&\\

  \end{tabularx}\vspace{-6pt}
  \caption{All \textbf{measures of complexity (Y axes)} of random networks generally increase with \textbf{weight/activation magnitudes (X axis)}. The sensitivity
  is however very different across activation functions (columns).
  This sensitivity also increases with multiplicative interactions (\ie gating),
  decreases with residual connections,
  and is essentially absent with layer normalization.
  These effects are also visible on the heatmaps (see next pages), but faint hence visualized here as line plots.
  }
\end{figure*}

\begin{figure*}[h!]
  \centering
  \renewcommand{\tabcolsep}{2em}
  \renewcommand{\arraystretch}{1}
  \small
  \begin{tabular}{ccc}
    \includegraphics[width=.17\linewidth]{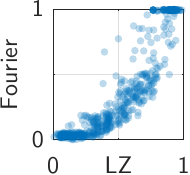}&
    \includegraphics[width=.17\linewidth]{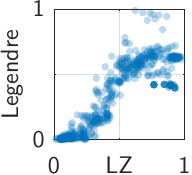}&
    \includegraphics[width=.17\linewidth]{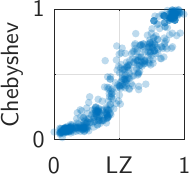}
  \end{tabular}\vspace{-6pt}
  \caption{\label{figCorrelationAppendix}
  Correlations between our measures of complexity on random networks.
  They are based on frequency (Fourier), polynomial order (Legendre, Chebyshev), or compressibility (LZ).
  They capture different notions, yet they are closely correlated.\vspace{-12pt}}
\end{figure*}

\clearpage
\begin{figure*}[h!]
  \centering
  \renewcommand{\tabcolsep}{0.93em}
  \renewcommand{\arraystretch}{1.1}
  \small
  \begin{tabularx}{\linewidth}{cccccccc}
    \scriptsize\textsf{ReLU} & \scriptsize\textsf{GELU} & \scriptsize\textsf{Swish} & \scriptsize\textsf{SELU} & \scriptsize\textsf{TanH} & \scriptsize\textsf{Gaussian} & \scriptsize\textsf{Sin} & \scriptsize\textsf{Unbiased}\\

    \includegraphics[width=.09\linewidth]{figFourier/fourier-spectrumIntegerFrequencies-signed-mlpRelu-fourier-heatmapWtNumLayers-10x6.pdf}&
    \includegraphics[width=.09\linewidth]{figFourier/fourier-spectrumIntegerFrequencies-signed-mlpGelu-fourier-heatmapWtNumLayers-10x6.pdf}&
    \includegraphics[width=.09\linewidth]{figFourier/fourier-spectrumIntegerFrequencies-signed-mlpSwish-fourier-heatmapWtNumLayers-10x6.pdf}&
    \includegraphics[width=.09\linewidth]{figFourier/fourier-spectrumIntegerFrequencies-signed-mlpSelu-fourier-heatmapWtNumLayers-10x6.pdf}&
    \includegraphics[width=.09\linewidth]{figFourier/fourier-spectrumIntegerFrequencies-signed-mlpTanh-fourier-heatmapWtNumLayers-10x6.pdf}&
    \includegraphics[width=.09\linewidth]{figFourier/fourier-spectrumIntegerFrequencies-signed-mlpGaussian-fourier-heatmapWtNumLayers-10x6.pdf}&
    \includegraphics[width=.09\linewidth]{figFourier/fourier-spectrumIntegerFrequencies-signed-mlpSin-fourier-heatmapWtNumLayers-10x6.pdf}&
    \includegraphics[width=.09\linewidth]{figFourier/fourier-spectrumIntegerFrequencies-signed-reconstructFromCoefs-fourier-heatmapWtNumLayers-10x6.pdf}\\
    \includegraphics[width=.09\linewidth]{figUntrained/mlpRelu-biasScale1.0-5x6.png}&
    \includegraphics[width=.09\linewidth]{figUntrained/mlpGelu-biasScale1.0-5x6.png}&
    \includegraphics[width=.09\linewidth]{figUntrained/mlpSwish-biasScale1.0-5x6.png}&
    \includegraphics[width=.09\linewidth]{figUntrained/mlpSelu-biasScale1.0-5x6.png}&
    \includegraphics[width=.09\linewidth]{figUntrained/mlpTanh-biasScale1.0-5x6.png}&
    \includegraphics[width=.09\linewidth]{figUntrained/mlpGaussian-biasScale1.0-5x6.png}&
    \includegraphics[width=.09\linewidth]{figUntrained/mlpSin-biasScale1.0-5x6.png}&
    \includegraphics[width=.09\linewidth]{figUntrained/reconstructFromCoefs-biasScale1.0-5x6.png}\\

    \multicolumn{7}{l}{\textbf{With residual connections:}}\\

    \includegraphics[width=.09\linewidth]{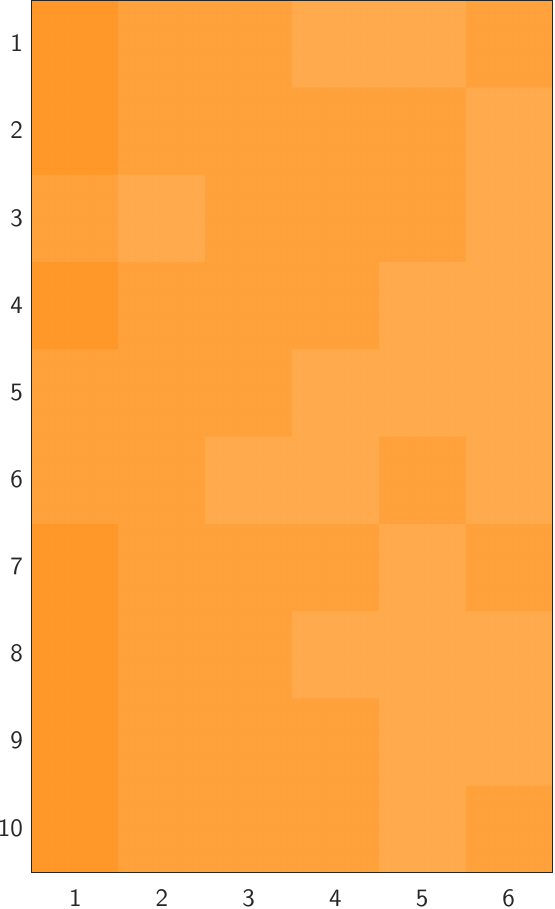}&
    \includegraphics[width=.09\linewidth]{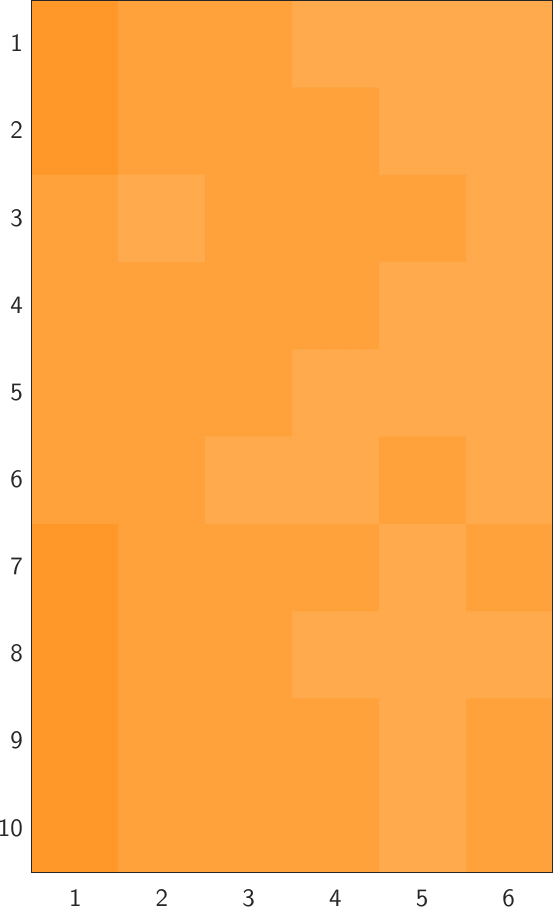}&
    \includegraphics[width=.09\linewidth]{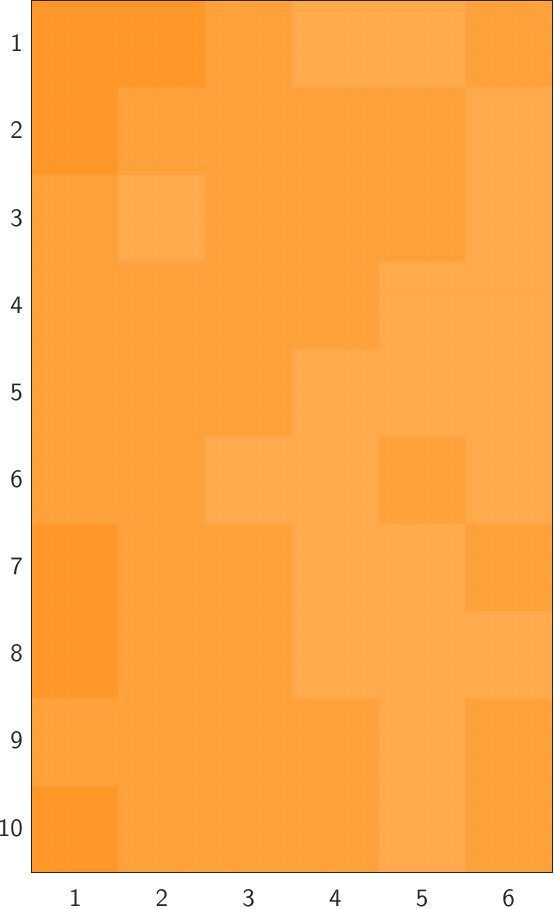}&
    \includegraphics[width=.09\linewidth]{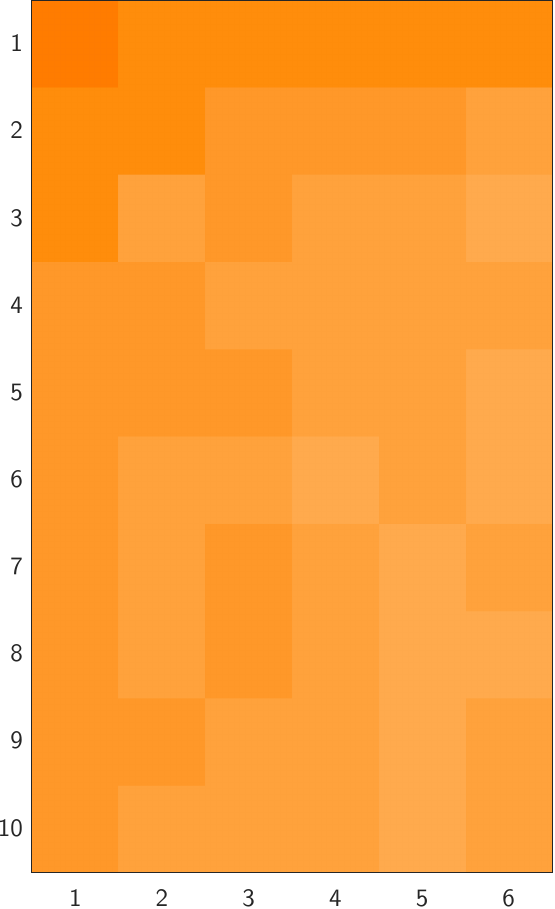}&
    \includegraphics[width=.09\linewidth]{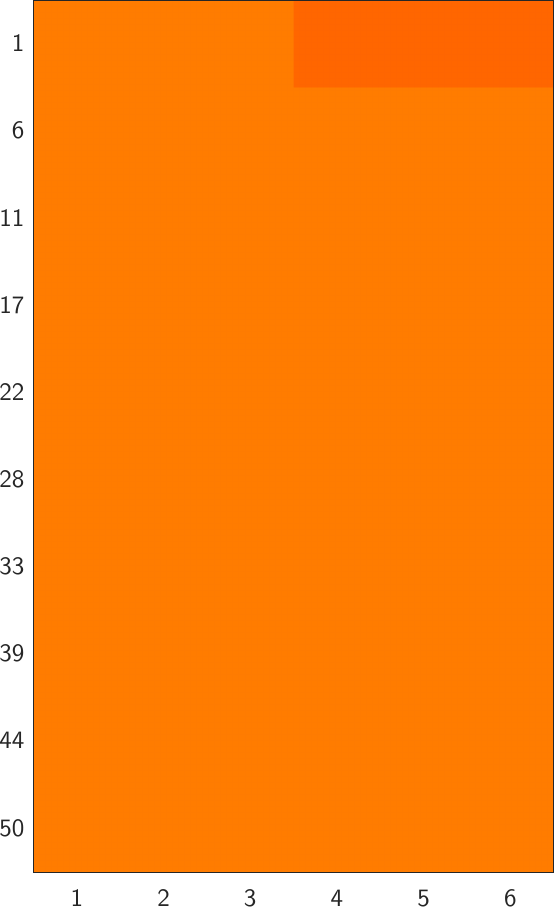}&
    \includegraphics[width=.09\linewidth]{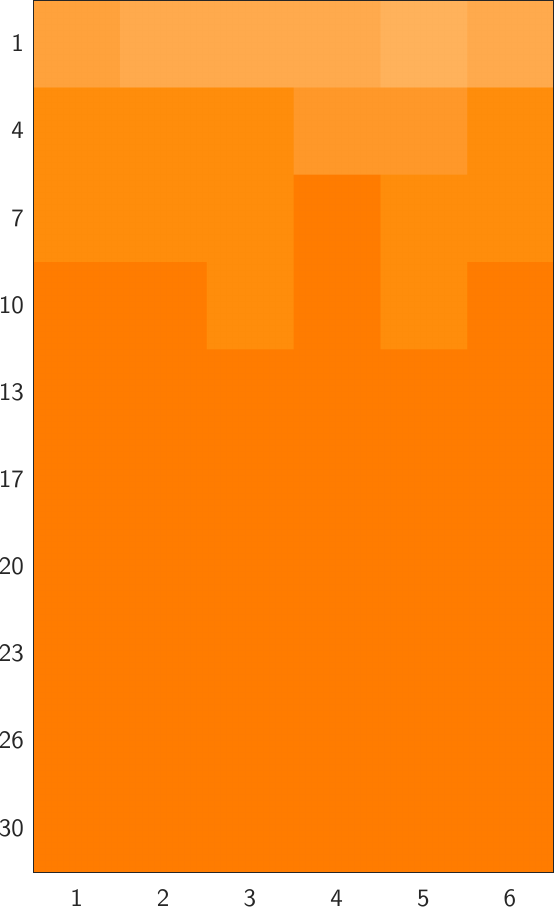}&
    \includegraphics[width=.09\linewidth]{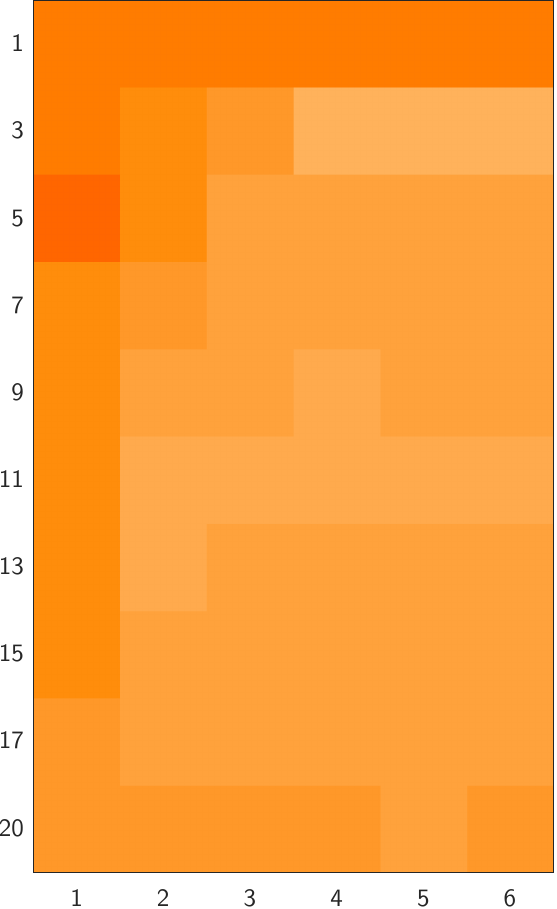}&
    \includegraphics[width=.09\linewidth]{figFourier/fourier-spectrumIntegerFrequencies-signed-reconstructFromCoefs-fourier-heatmapWtNumLayers-10x6.pdf}\\
    \includegraphics[width=.09\linewidth]{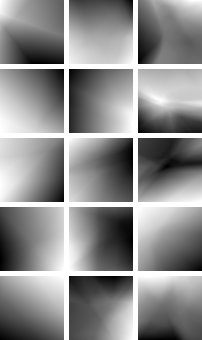}&
    \includegraphics[width=.09\linewidth]{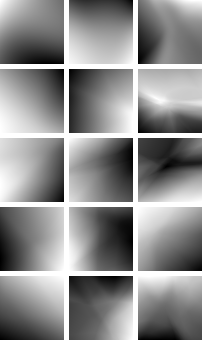}&
    \includegraphics[width=.09\linewidth]{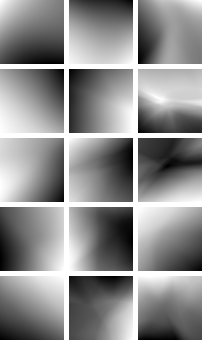}&
    \includegraphics[width=.09\linewidth]{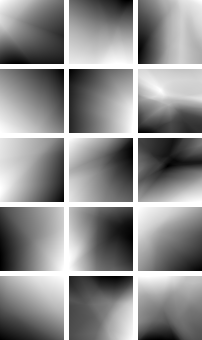}&
    \includegraphics[width=.09\linewidth]{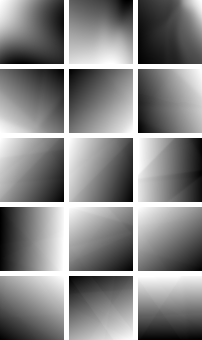}&
    \includegraphics[width=.09\linewidth]{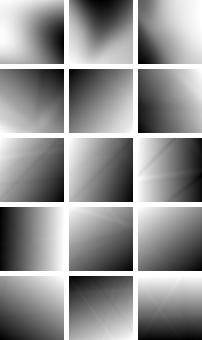}&
    \includegraphics[width=.09\linewidth]{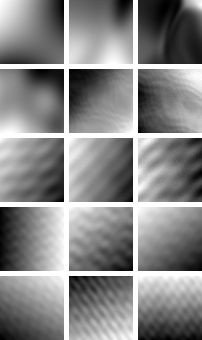}&
    \includegraphics[width=.09\linewidth]{figUntrained/reconstructFromCoefs-biasScale1.0-5x6.png}\\

    \multicolumn{7}{l}{\textbf{With layer normalization:}}\\

    \includegraphics[width=.09\linewidth]{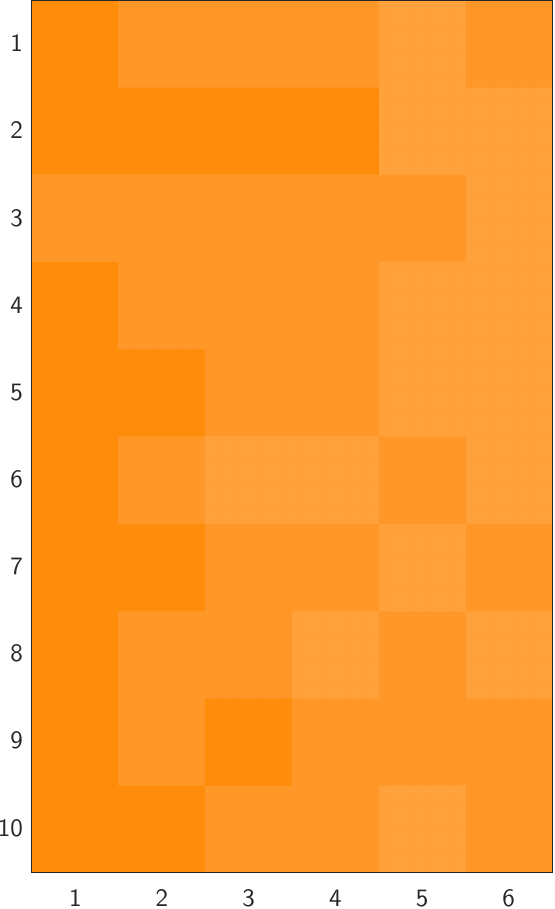}&
    \includegraphics[width=.09\linewidth]{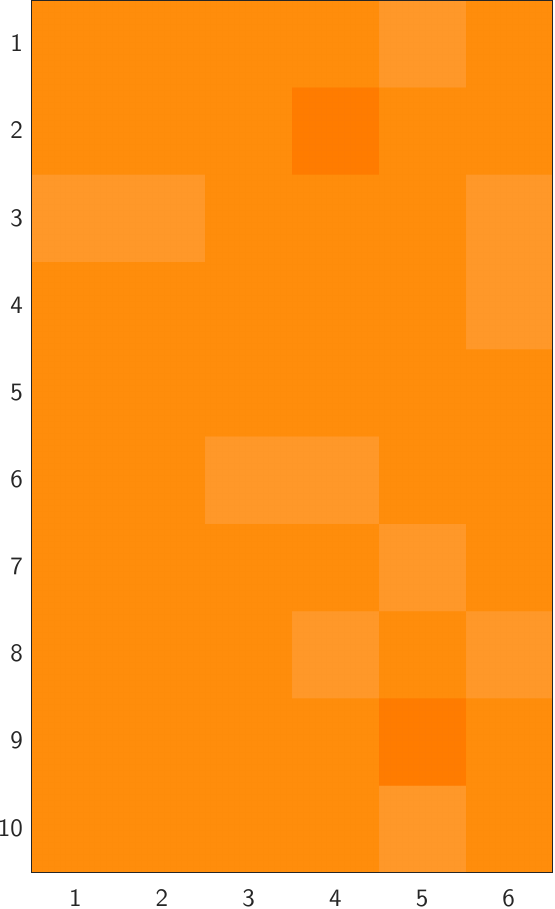}&
    \includegraphics[width=.09\linewidth]{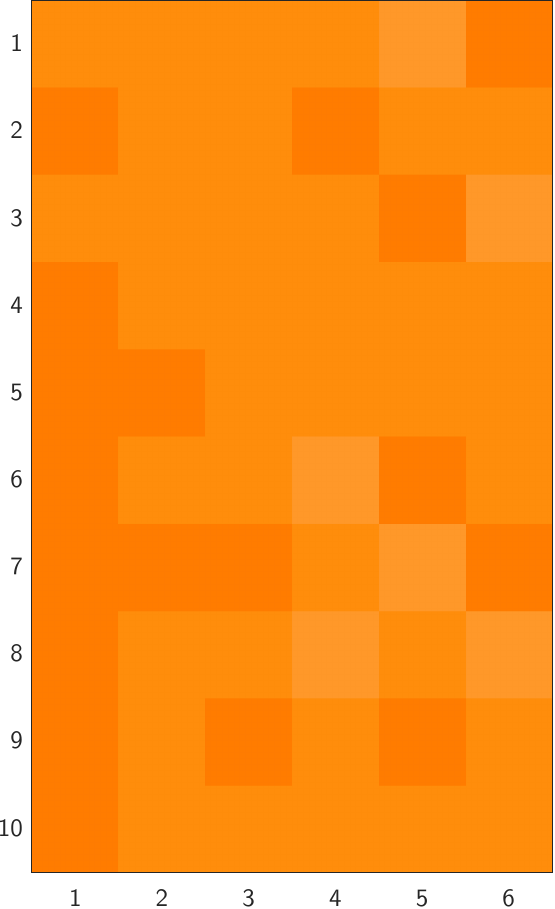}&
    \includegraphics[width=.09\linewidth]{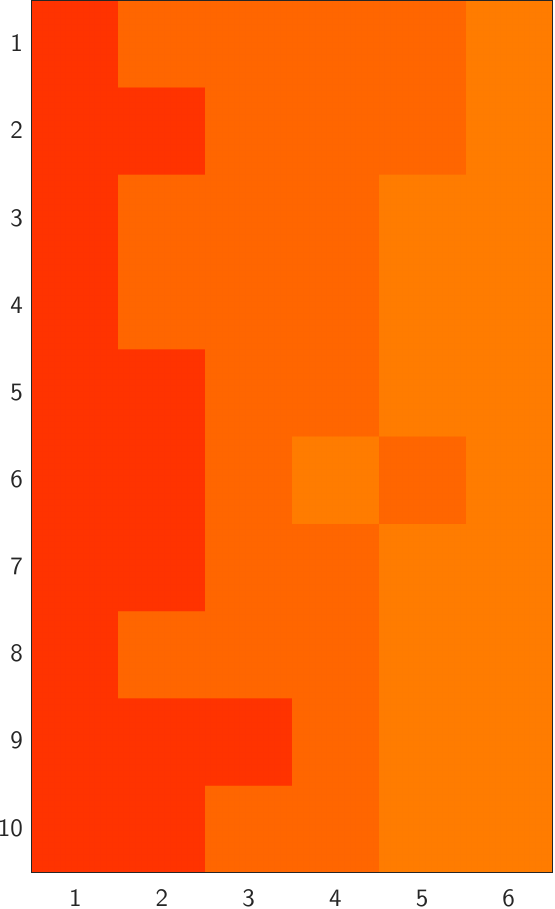}&
    \includegraphics[width=.09\linewidth]{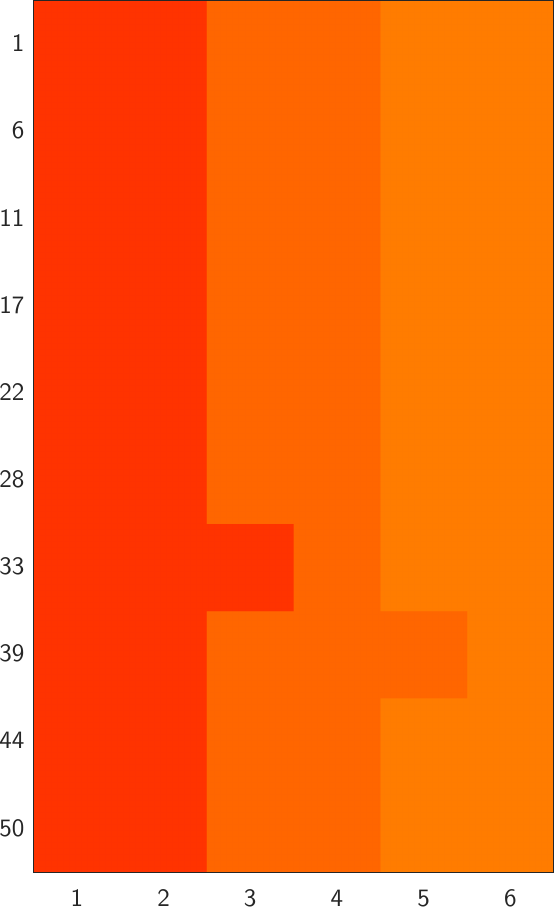}&
    \includegraphics[width=.09\linewidth]{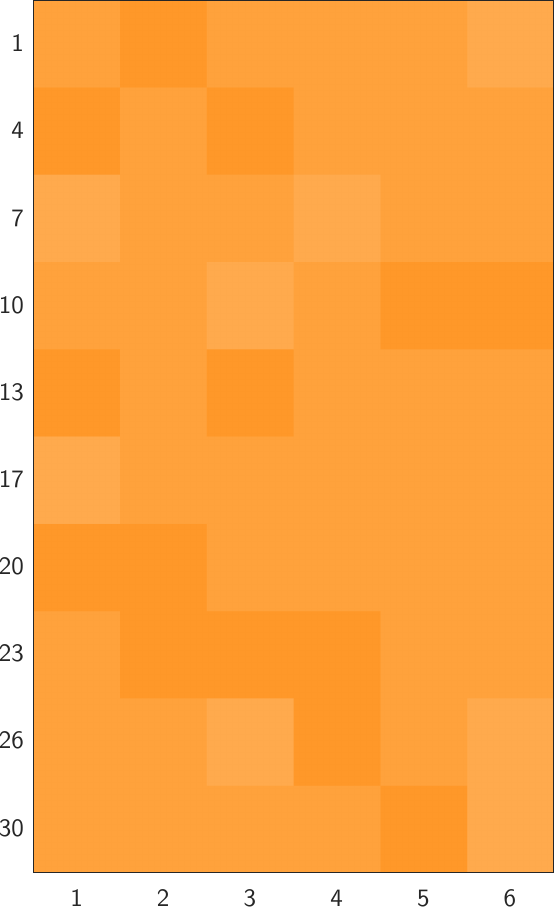}&
    \includegraphics[width=.09\linewidth]{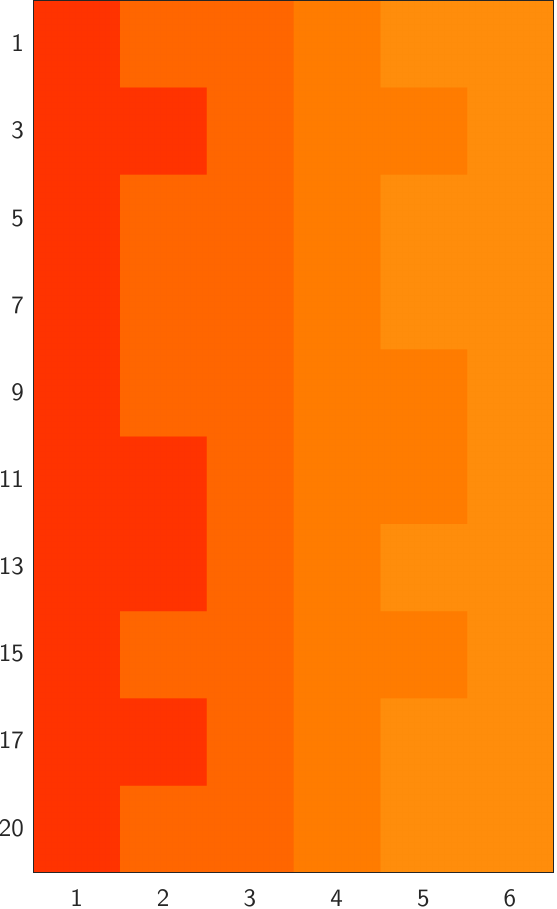}&
    \includegraphics[width=.09\linewidth]{figFourier/fourier-spectrumIntegerFrequencies-signed-reconstructFromCoefs-fourier-heatmapWtNumLayers-10x6.pdf}\\
    \includegraphics[width=.09\linewidth]{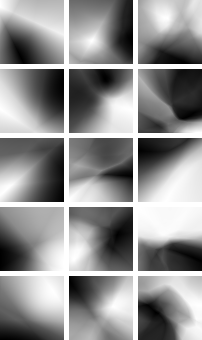}&
    \includegraphics[width=.09\linewidth]{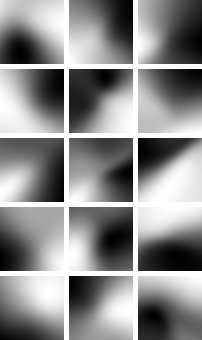}&
    \includegraphics[width=.09\linewidth]{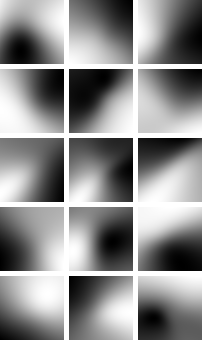}&
    \includegraphics[width=.09\linewidth]{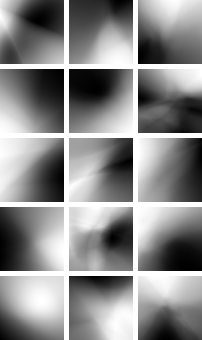}&
    \includegraphics[width=.09\linewidth]{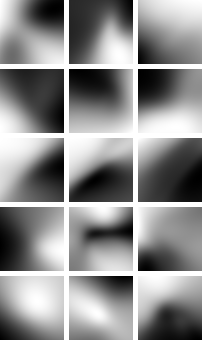}&
    \includegraphics[width=.09\linewidth]{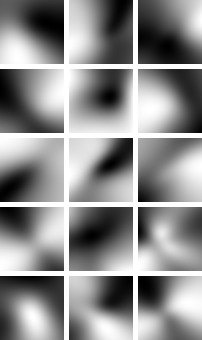}&
    \includegraphics[width=.09\linewidth]{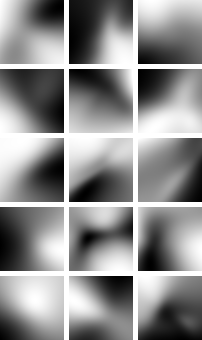}&
    \includegraphics[width=.09\linewidth]{figUntrained/reconstructFromCoefs-biasScale1.0-5x6.png}\\

    \multicolumn{7}{l}{\textbf{With gating:}}\\

    \includegraphics[width=.09\linewidth]{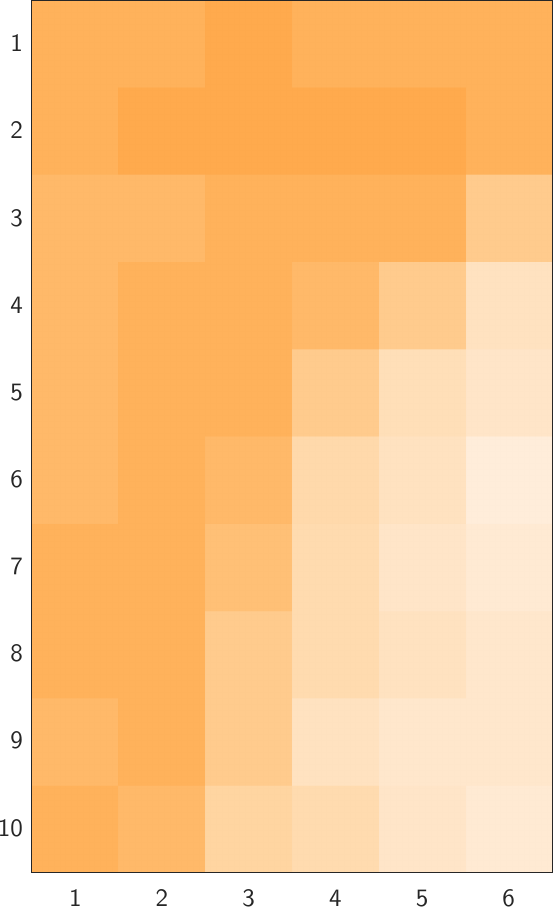}&
    \includegraphics[width=.09\linewidth]{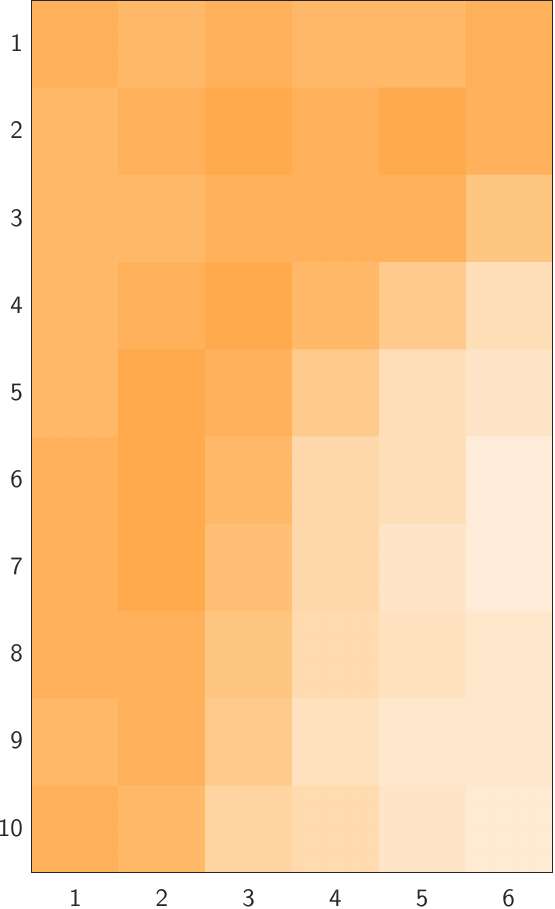}&
    \includegraphics[width=.09\linewidth]{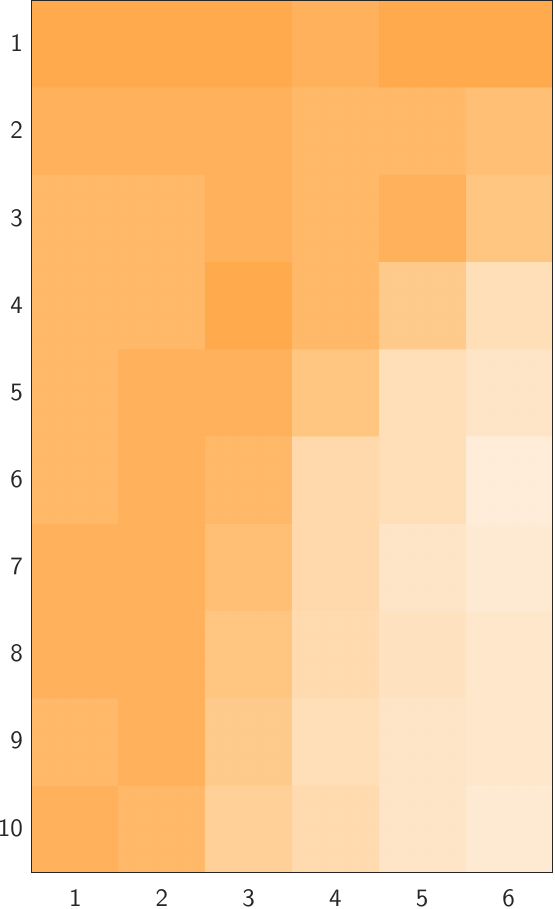}&
    \includegraphics[width=.09\linewidth]{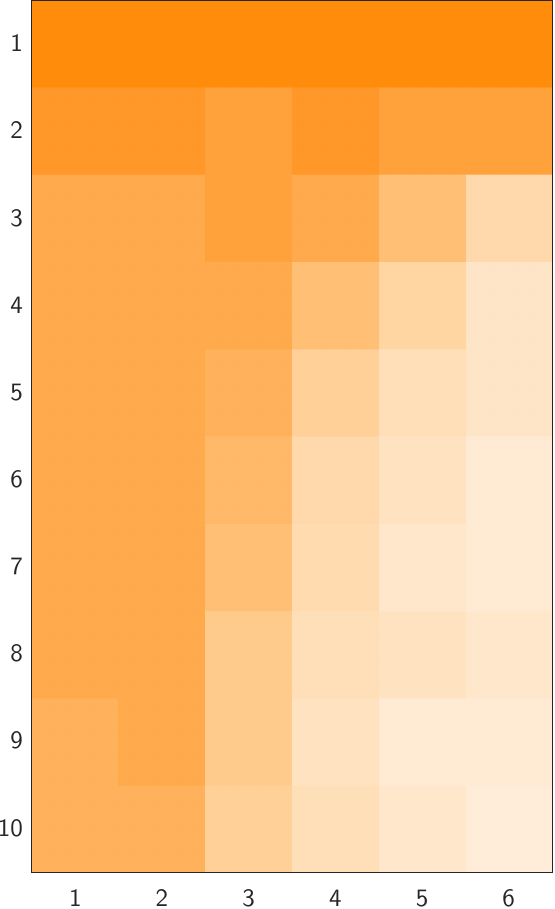}&
    \includegraphics[width=.09\linewidth]{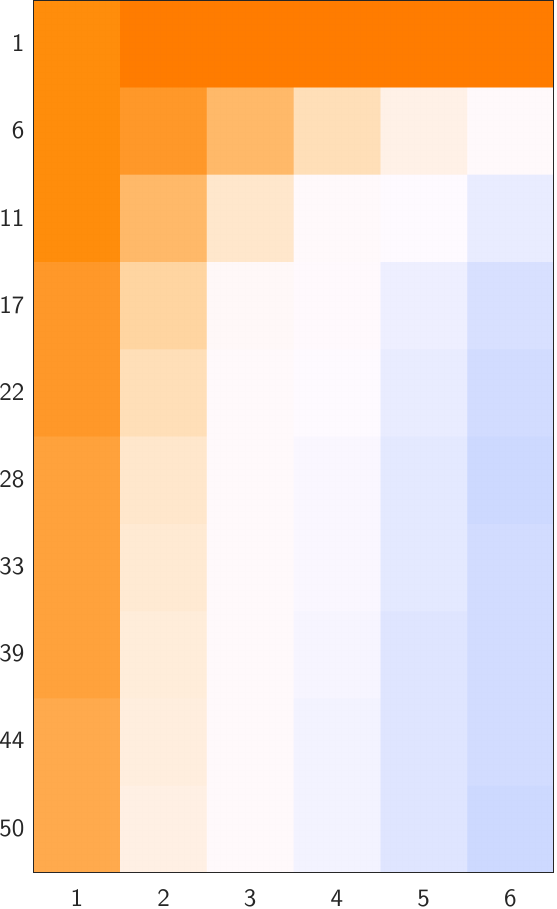}&
    \includegraphics[width=.09\linewidth]{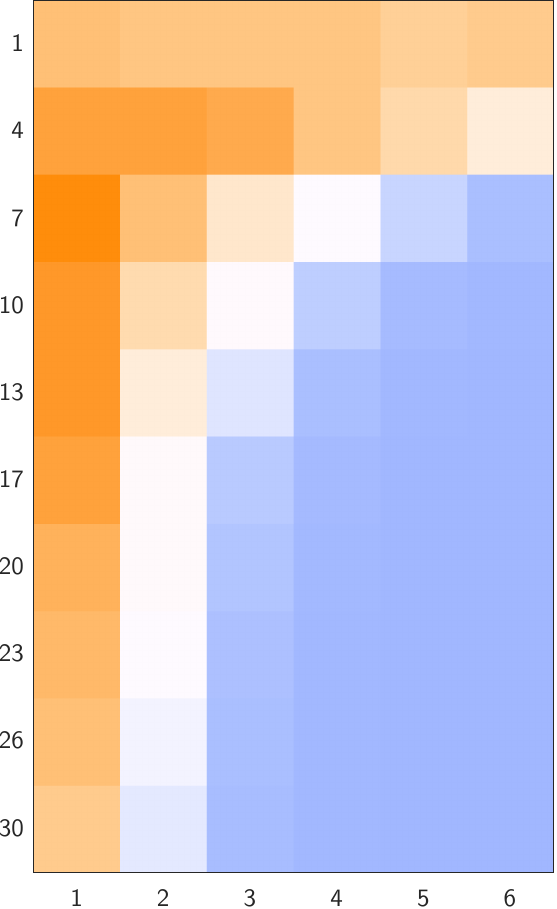}&
    \includegraphics[width=.09\linewidth]{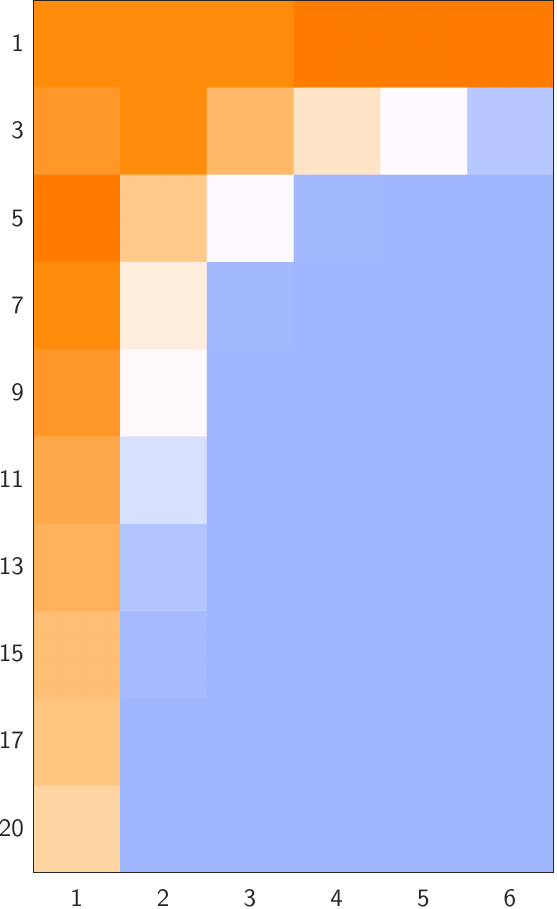}&
    \includegraphics[width=.09\linewidth]{figFourier/fourier-spectrumIntegerFrequencies-signed-reconstructFromCoefs-fourier-heatmapWtNumLayers-10x6.pdf}\\
    \includegraphics[width=.09\linewidth]{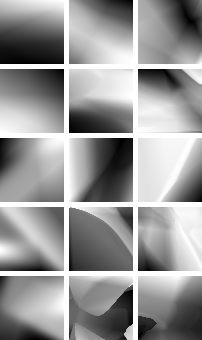}&
    \includegraphics[width=.09\linewidth]{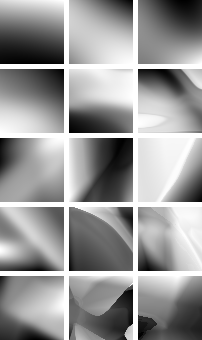}&
    \includegraphics[width=.09\linewidth]{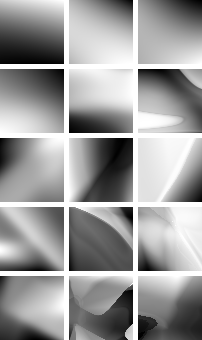}&
    \includegraphics[width=.09\linewidth]{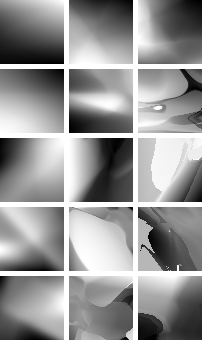}&
    \includegraphics[width=.09\linewidth]{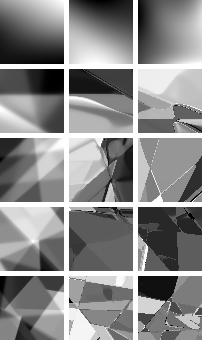}&
    \includegraphics[width=.09\linewidth]{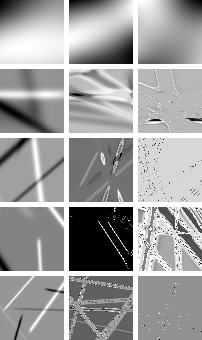}&
    \includegraphics[width=.09\linewidth]{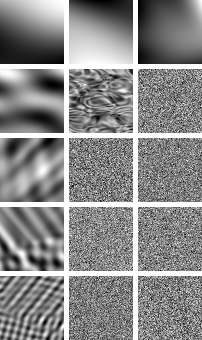}&
    \includegraphics[width=.09\linewidth]{figUntrained/reconstructFromCoefs-biasScale1.0-5x6.png}\\
    
  \end{tabularx}\vspace{-6pt}
  \caption{\label{figHeatmapsFourierAppendix}
  Heatmaps of the average complexity (\textbf{Fourier})
  of various architectures with random weights, and example functions.
  \vspace{-8pt}}
\end{figure*}

\clearpage
\begin{figure*}[h!]
  \centering
  \renewcommand{\tabcolsep}{0.93em}
  \renewcommand{\arraystretch}{1.1}
  \small
  \begin{tabularx}{\linewidth}{cccccccc}
    \scriptsize\textsf{ReLU} & \scriptsize\textsf{GELU} & \scriptsize\textsf{Swish} & \scriptsize\textsf{SELU} & \scriptsize\textsf{TanH} & \scriptsize\textsf{Gaussian} & \scriptsize\textsf{Sin} & \scriptsize\textsf{Unbiased}\\

    \includegraphics[width=.09\linewidth]{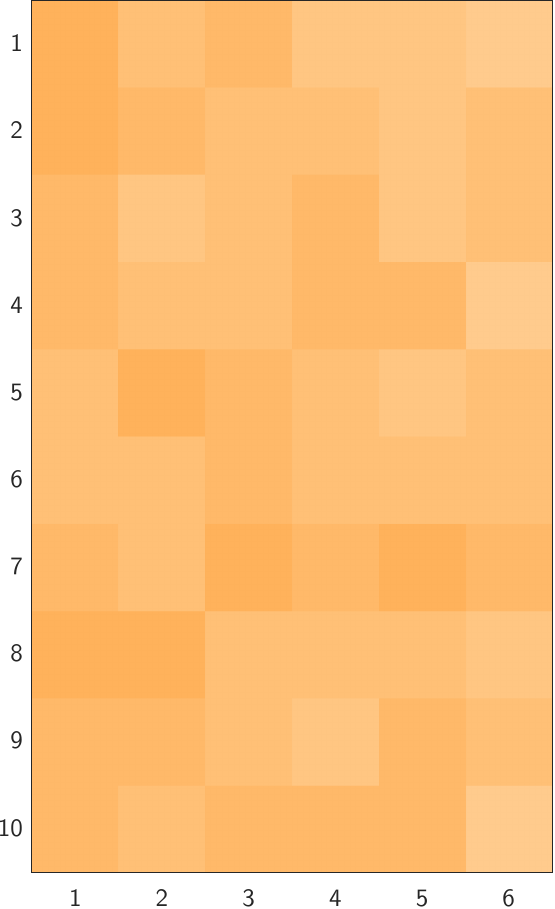}&
    \includegraphics[width=.09\linewidth]{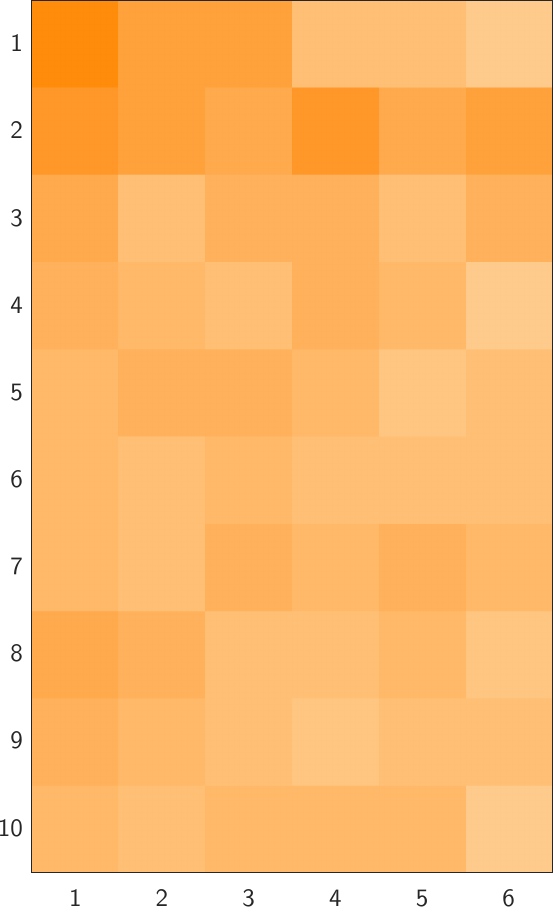}&
    \includegraphics[width=.09\linewidth]{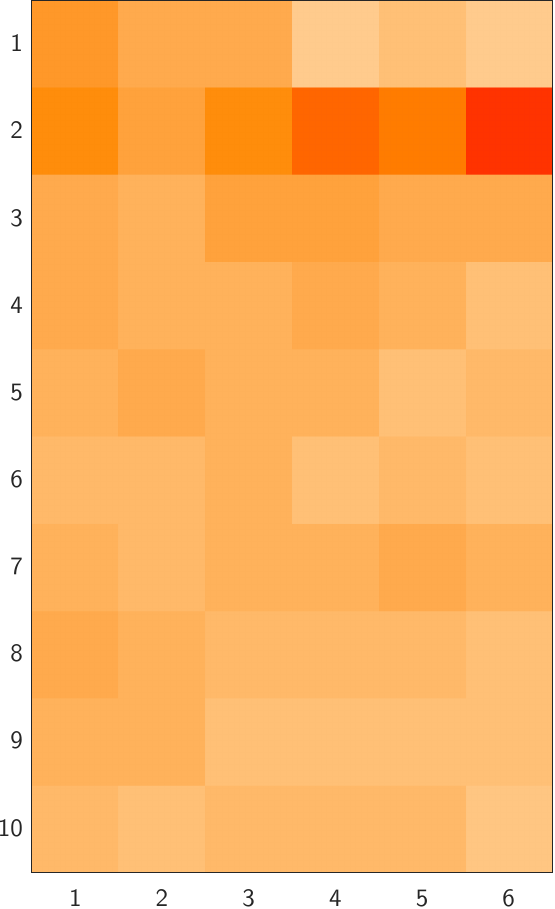}&
    \includegraphics[width=.09\linewidth]{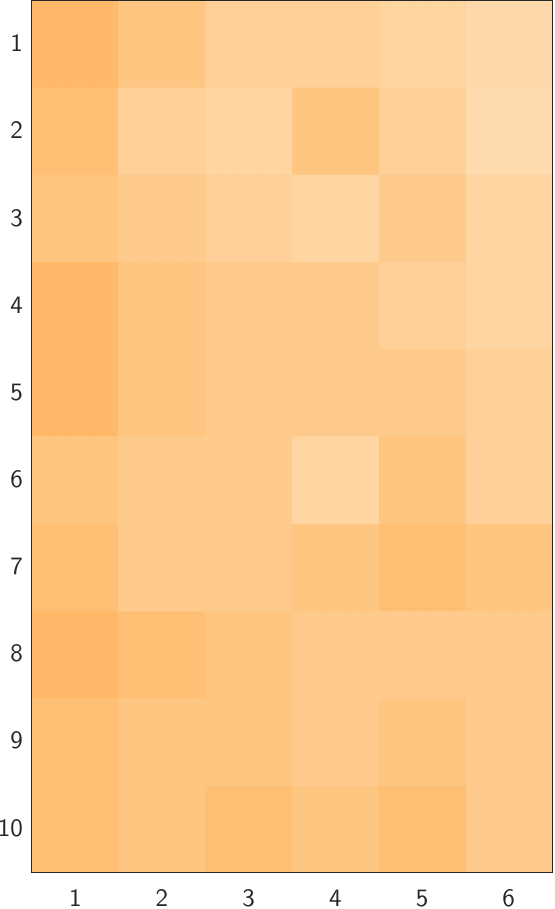}&
    \includegraphics[width=.09\linewidth]{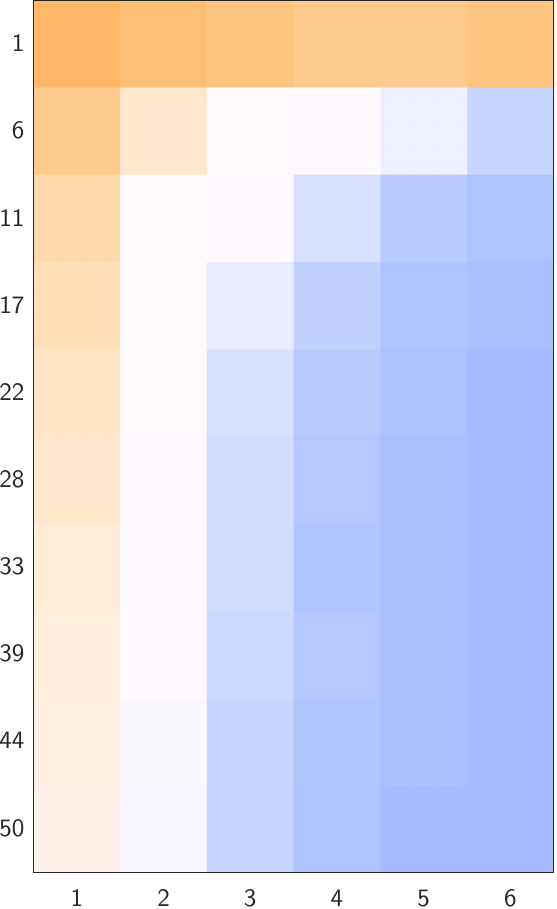}&
    \includegraphics[width=.09\linewidth]{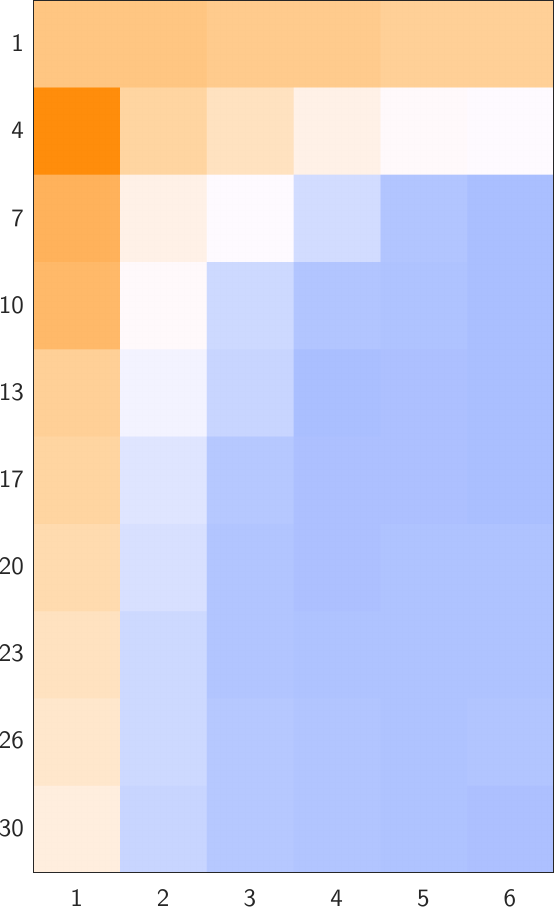}&
    \includegraphics[width=.09\linewidth]{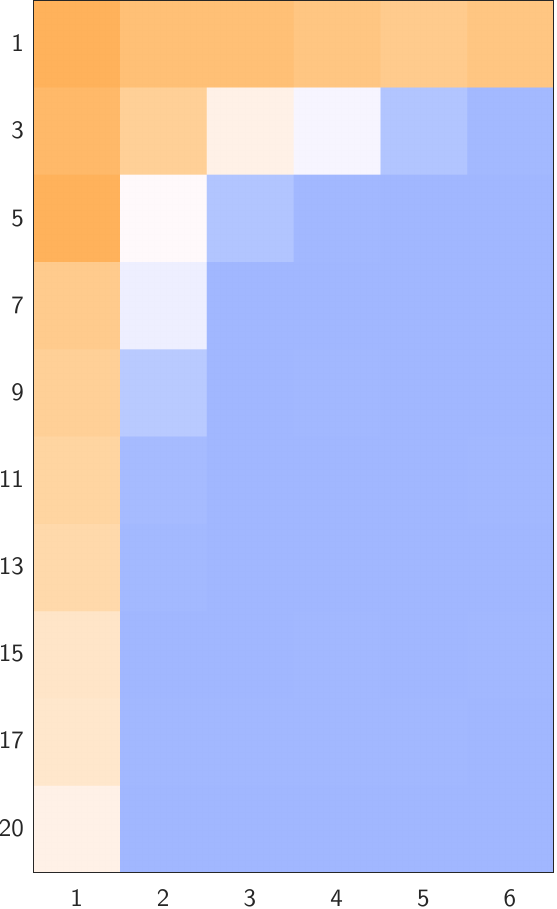}&
    \includegraphics[width=.09\linewidth]{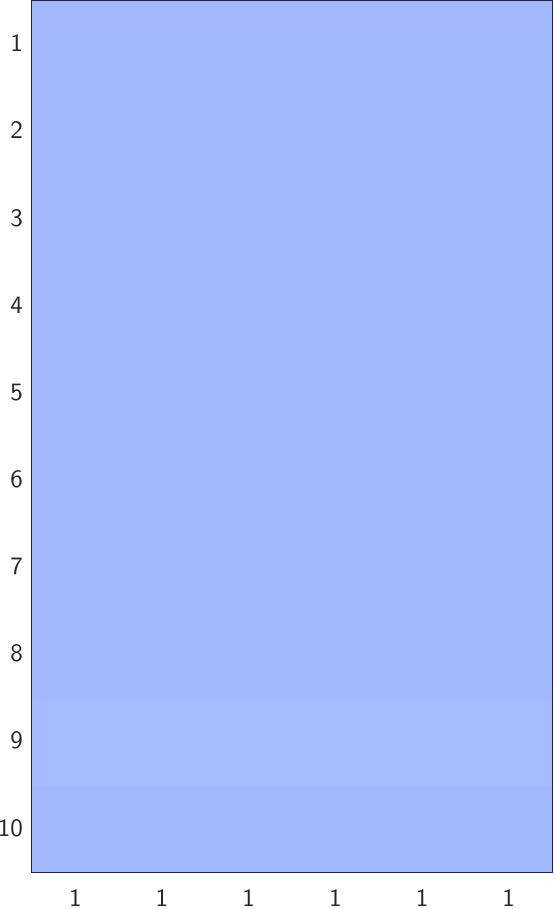}\\
    \includegraphics[width=.09\linewidth]{figUntrained/mlpRelu-biasScale1.0-5x6.png}&
    \includegraphics[width=.09\linewidth]{figUntrained/mlpGelu-biasScale1.0-5x6.png}&
    \includegraphics[width=.09\linewidth]{figUntrained/mlpSwish-biasScale1.0-5x6.png}&
    \includegraphics[width=.09\linewidth]{figUntrained/mlpSelu-biasScale1.0-5x6.png}&
    \includegraphics[width=.09\linewidth]{figUntrained/mlpTanh-biasScale1.0-5x6.png}&
    \includegraphics[width=.09\linewidth]{figUntrained/mlpGaussian-biasScale1.0-5x6.png}&
    \includegraphics[width=.09\linewidth]{figUntrained/mlpSin-biasScale1.0-5x6.png}&
    \includegraphics[width=.09\linewidth]{figUntrained/reconstructFromCoefs-biasScale1.0-5x6.png}\\

    \multicolumn{7}{l}{\textbf{With residual connections:}}\\

    \includegraphics[width=.09\linewidth]{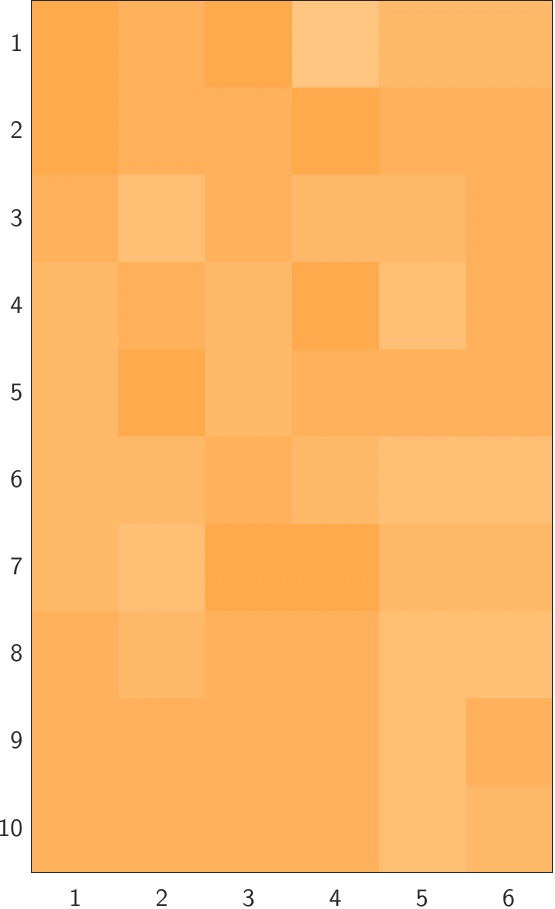}&
    \includegraphics[width=.09\linewidth]{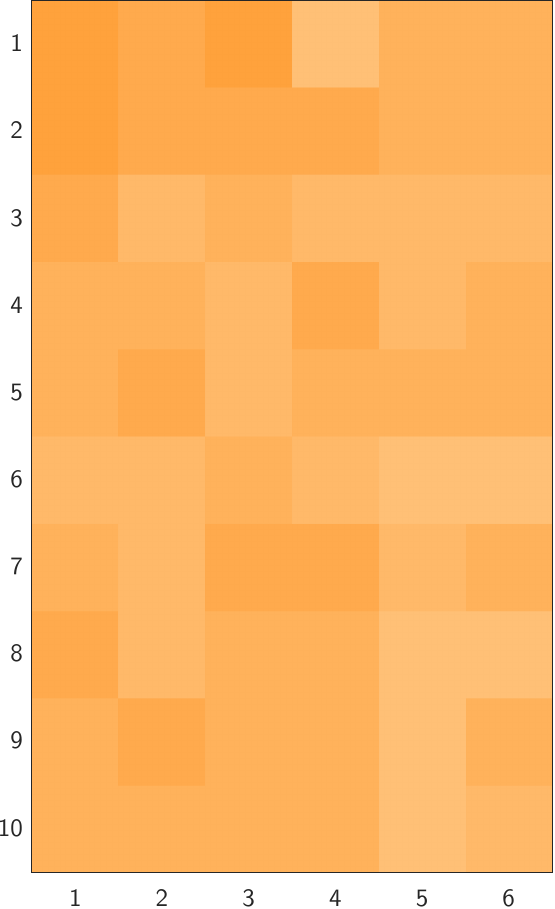}&
    \includegraphics[width=.09\linewidth]{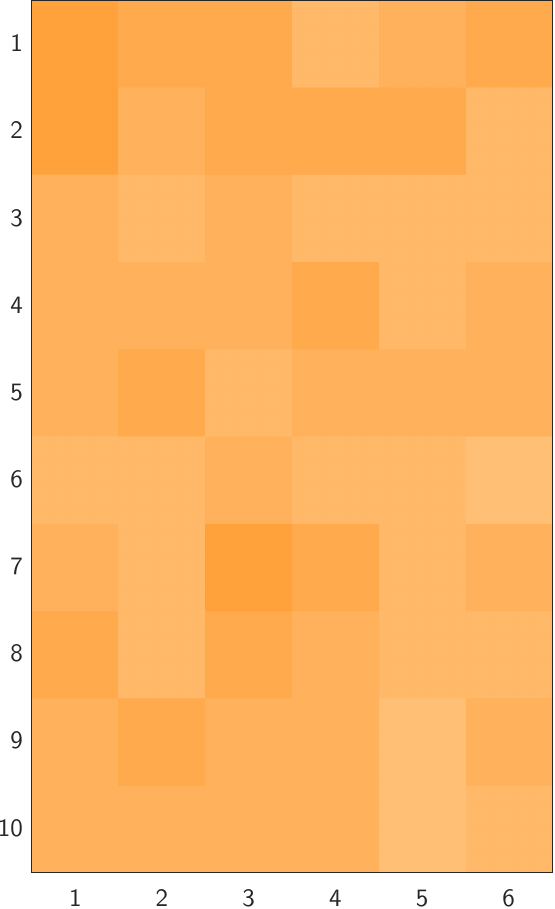}&
    \includegraphics[width=.09\linewidth]{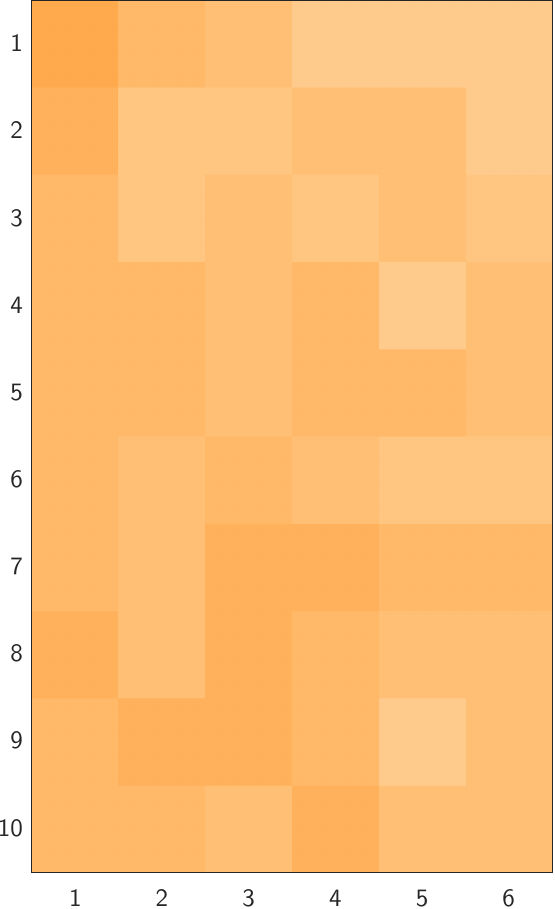}&
    \includegraphics[width=.09\linewidth]{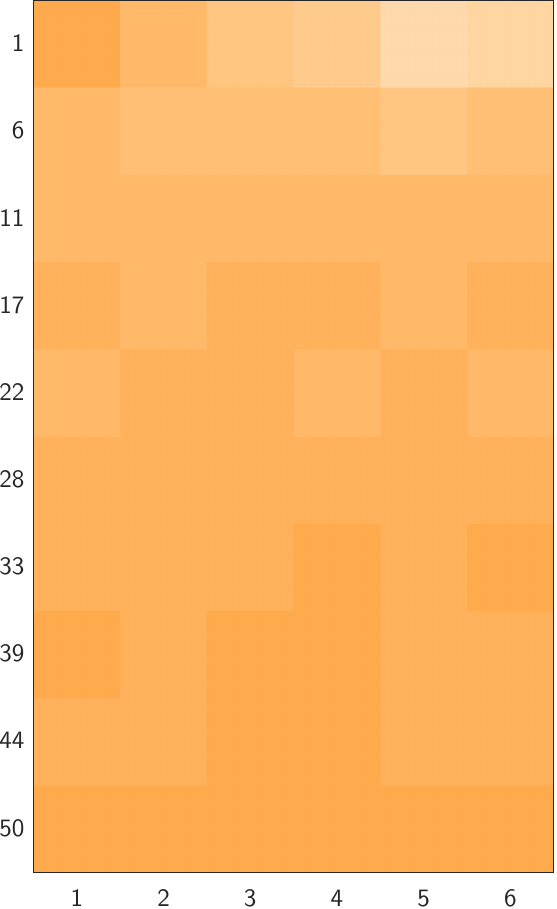}&
    \includegraphics[width=.09\linewidth]{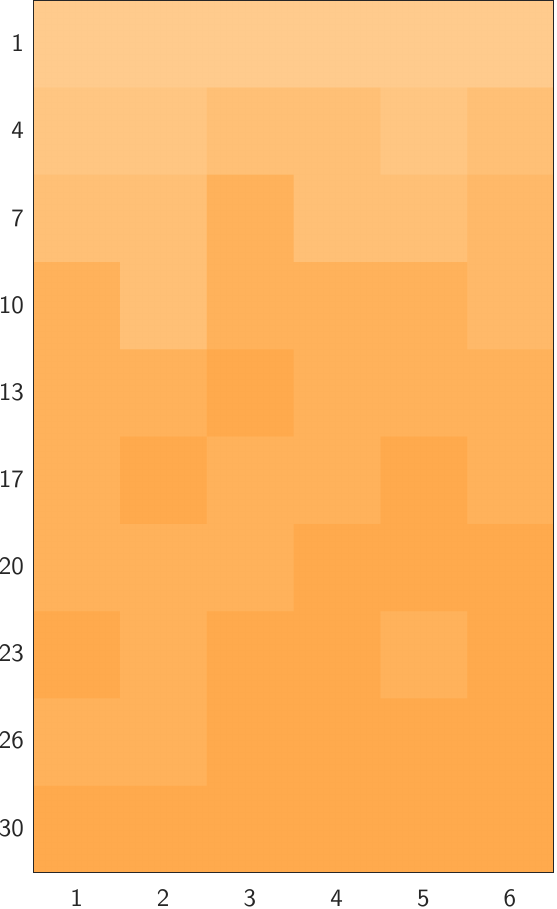}&
    \includegraphics[width=.09\linewidth]{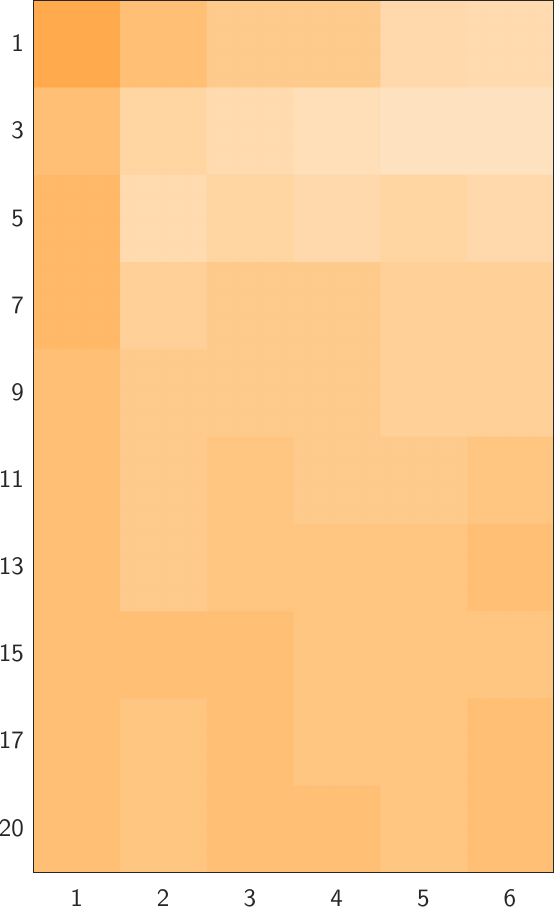}&
    \includegraphics[width=.09\linewidth]{figChebyshev/chebyshev-spectrumLowerFineFrequencies-100-signed-reconstructFromCoefs-chebyshev-heatmapWtNumLayers-10x6.pdf}\\
    \includegraphics[width=.09\linewidth]{figUntrained/mlpResRelu-biasScale1.0-5x6.png}&
    \includegraphics[width=.09\linewidth]{figUntrained/mlpResGelu-biasScale1.0-5x6.png}&
    \includegraphics[width=.09\linewidth]{figUntrained/mlpResSwish-biasScale1.0-5x6.png}&
    \includegraphics[width=.09\linewidth]{figUntrained/mlpResSelu-biasScale1.0-5x6.png}&
    \includegraphics[width=.09\linewidth]{figUntrained/mlpResTanh-biasScale1.0-5x6.png}&
    \includegraphics[width=.09\linewidth]{figUntrained/mlpResGaussian-biasScale1.0-5x6.png}&
    \includegraphics[width=.09\linewidth]{figUntrained/mlpResSin-biasScale1.0-5x6.png}&
    \includegraphics[width=.09\linewidth]{figUntrained/reconstructFromCoefs-biasScale1.0-5x6.png}\\

    \multicolumn{7}{l}{\textbf{With layer normalization:}}\\

    \includegraphics[width=.09\linewidth]{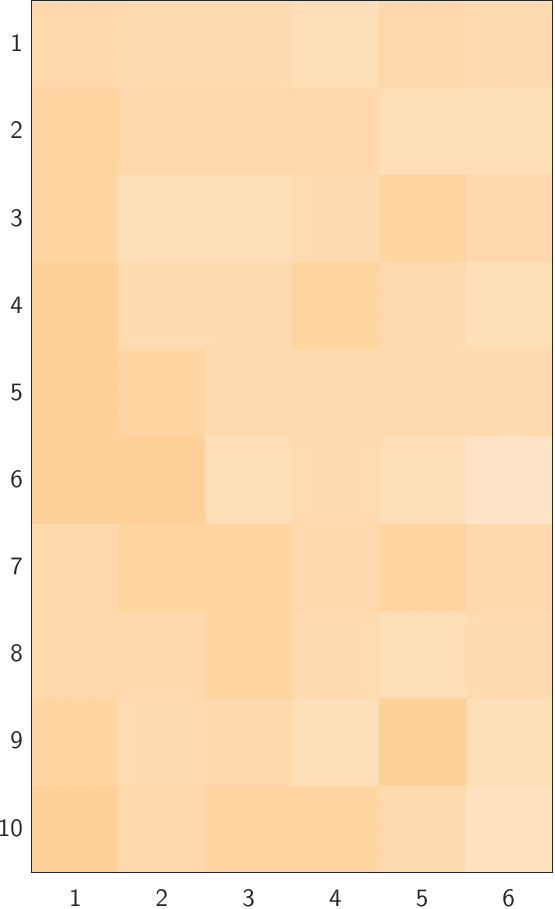}&
    \includegraphics[width=.09\linewidth]{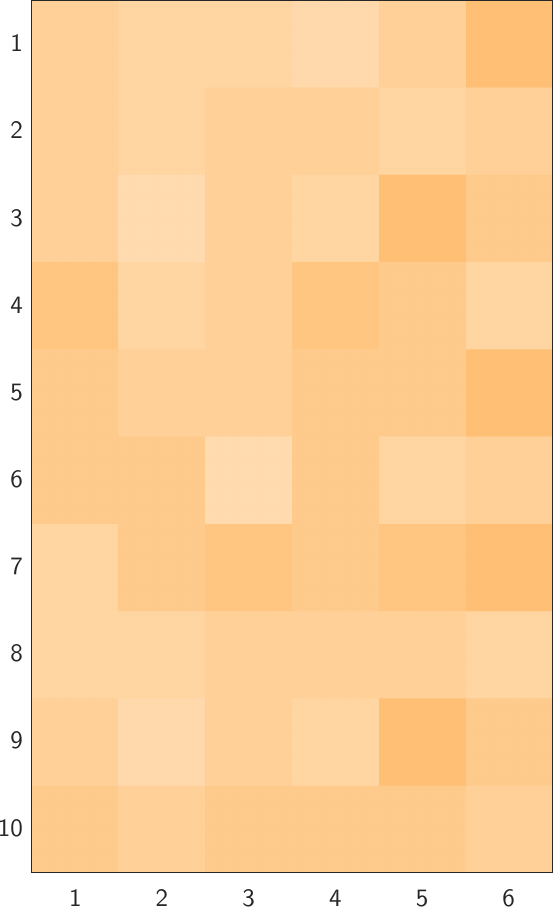}&
    \includegraphics[width=.09\linewidth]{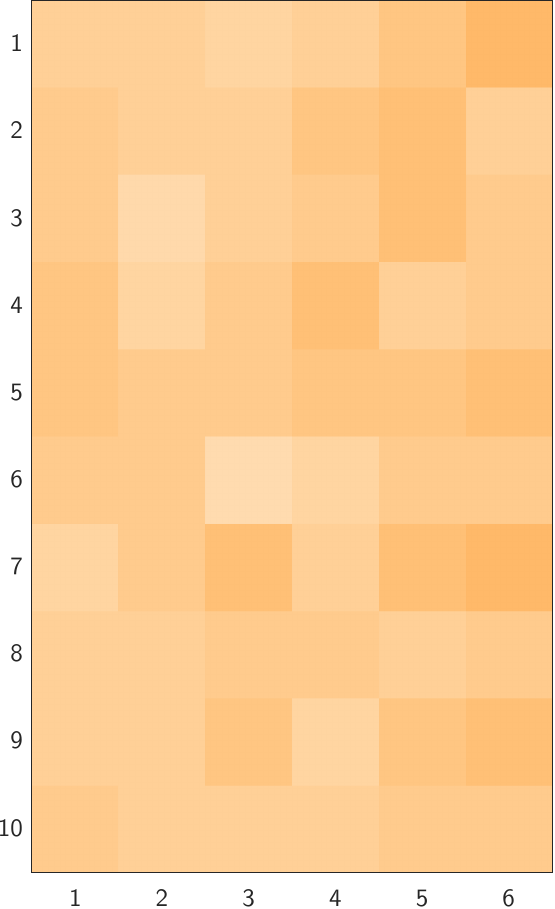}&
    \includegraphics[width=.09\linewidth]{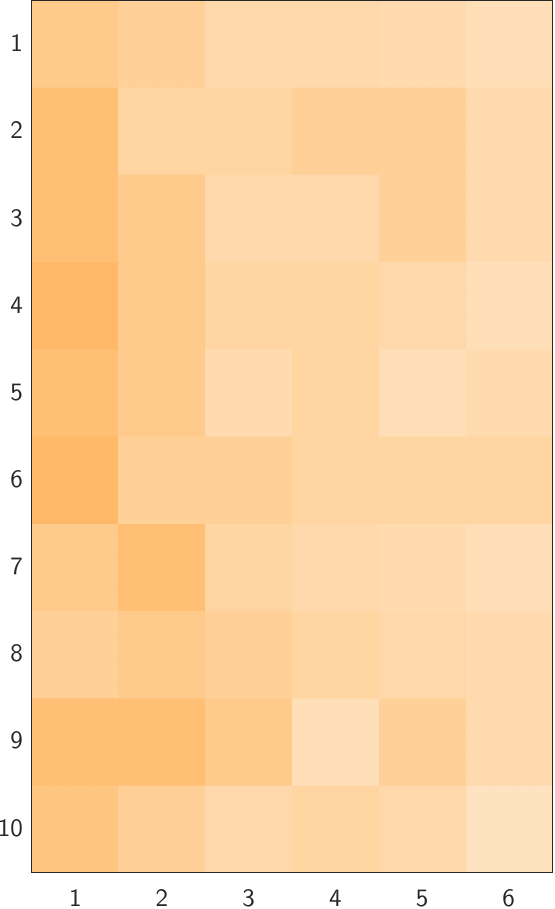}&
    \includegraphics[width=.09\linewidth]{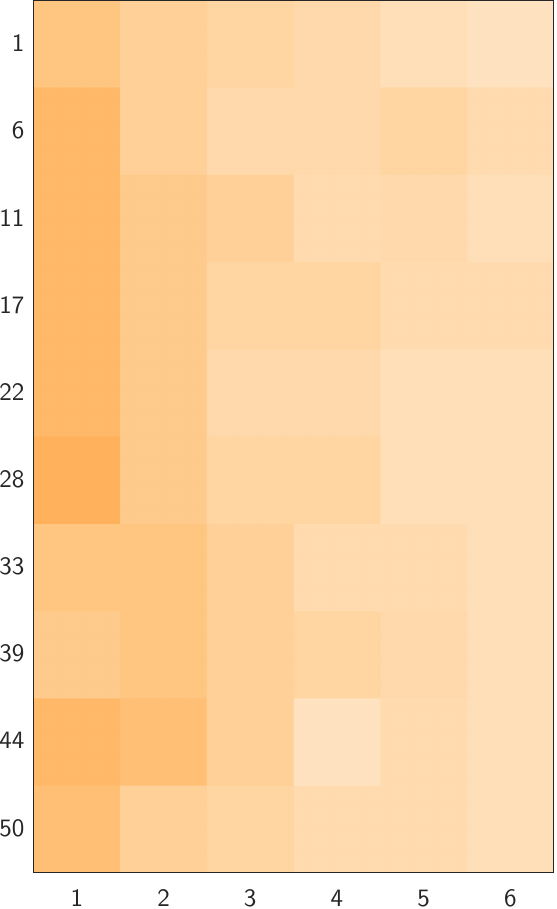}&
    \includegraphics[width=.09\linewidth]{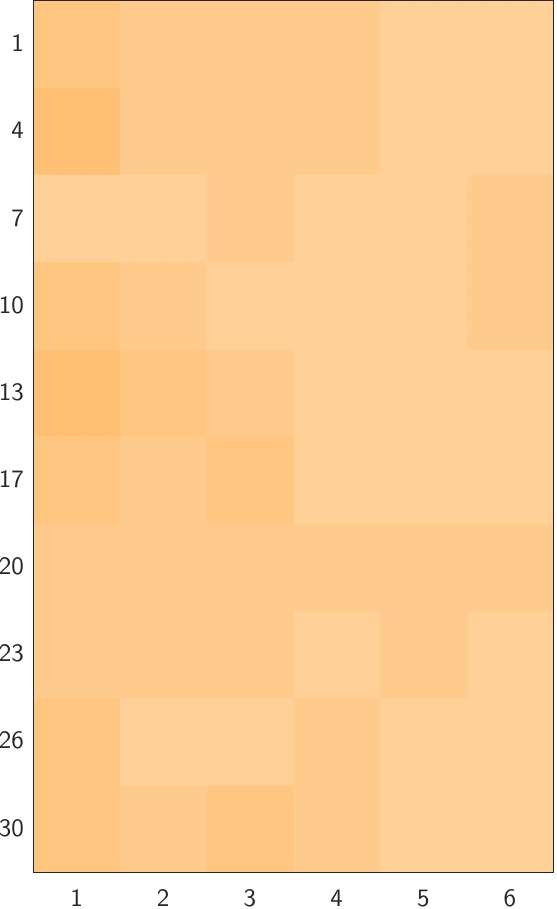}&
    \includegraphics[width=.09\linewidth]{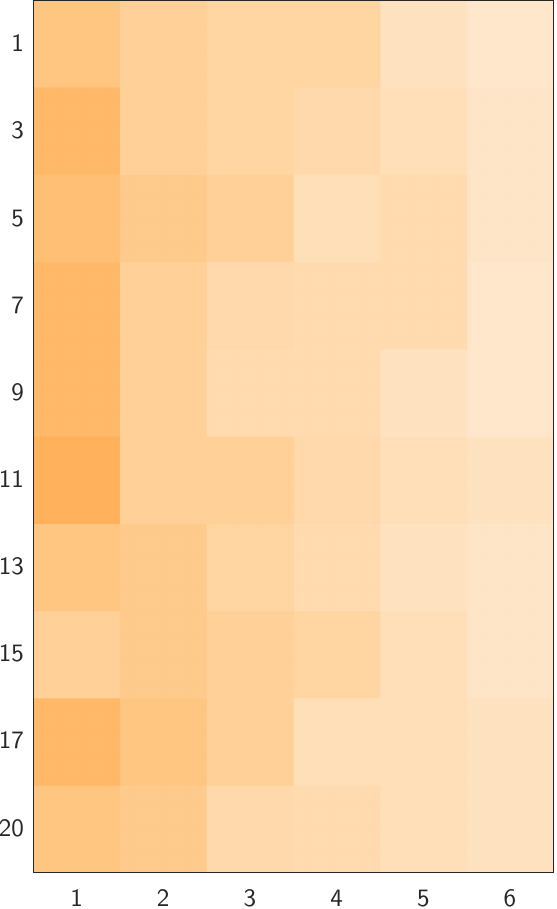}&
    \includegraphics[width=.09\linewidth]{figChebyshev/chebyshev-spectrumLowerFineFrequencies-100-signed-reconstructFromCoefs-chebyshev-heatmapWtNumLayers-10x6.pdf}\\
    \includegraphics[width=.09\linewidth]{figUntrained/mlpPostNormRelu-biasScale1.0-5x6.png}&
    \includegraphics[width=.09\linewidth]{figUntrained/mlpPostNormGelu-biasScale1.0-5x6.png}&
    \includegraphics[width=.09\linewidth]{figUntrained/mlpPostNormSwish-biasScale1.0-5x6.png}&
    \includegraphics[width=.09\linewidth]{figUntrained/mlpPostNormSelu-biasScale1.0-5x6.png}&
    \includegraphics[width=.09\linewidth]{figUntrained/mlpPostNormTanh-biasScale1.0-5x6.png}&
    \includegraphics[width=.09\linewidth]{figUntrained/mlpPostNormGaussian-biasScale1.0-5x6.png}&
    \includegraphics[width=.09\linewidth]{figUntrained/mlpPostNormSin-biasScale1.0-5x6.png}&
    \includegraphics[width=.09\linewidth]{figUntrained/reconstructFromCoefs-biasScale1.0-5x6.png}\\

    \multicolumn{7}{l}{\textbf{With gating:}}\\

    \includegraphics[width=.09\linewidth]{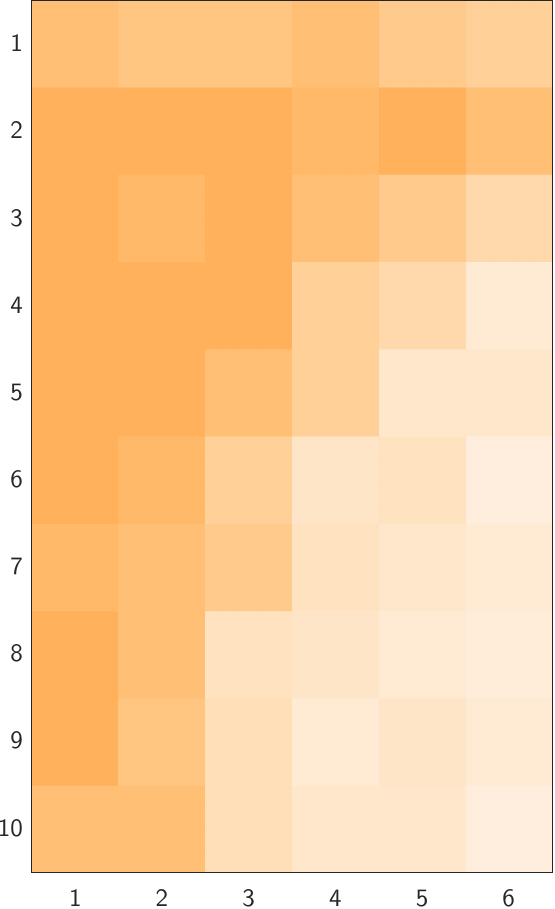}&
    \includegraphics[width=.09\linewidth]{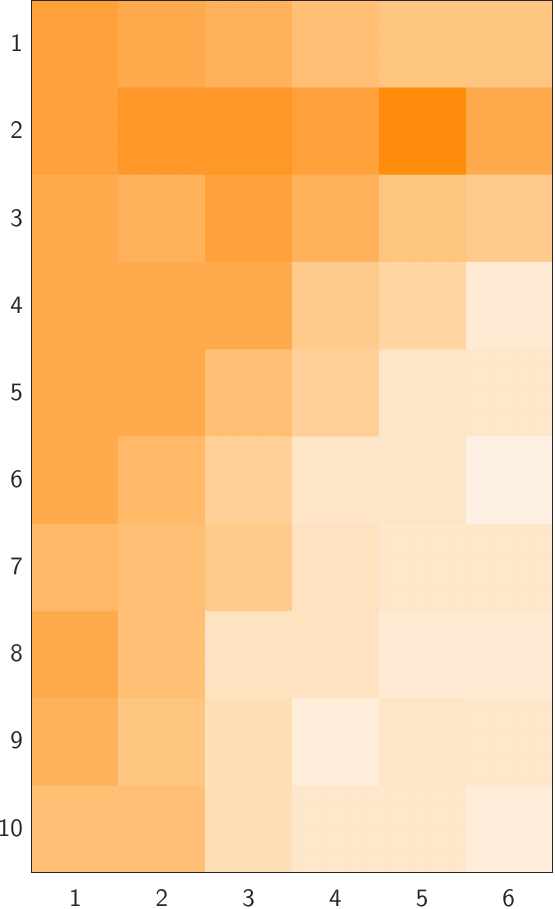}&
    \includegraphics[width=.09\linewidth]{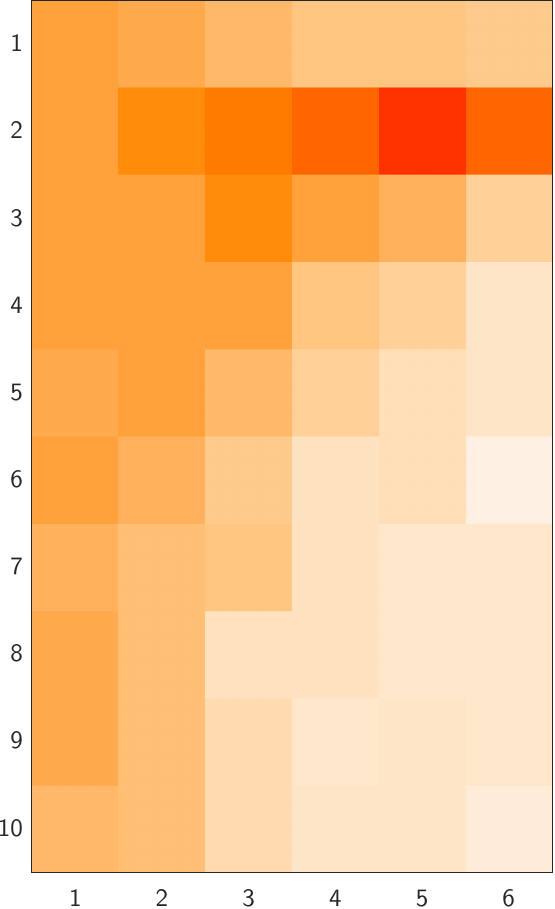}&
    \includegraphics[width=.09\linewidth]{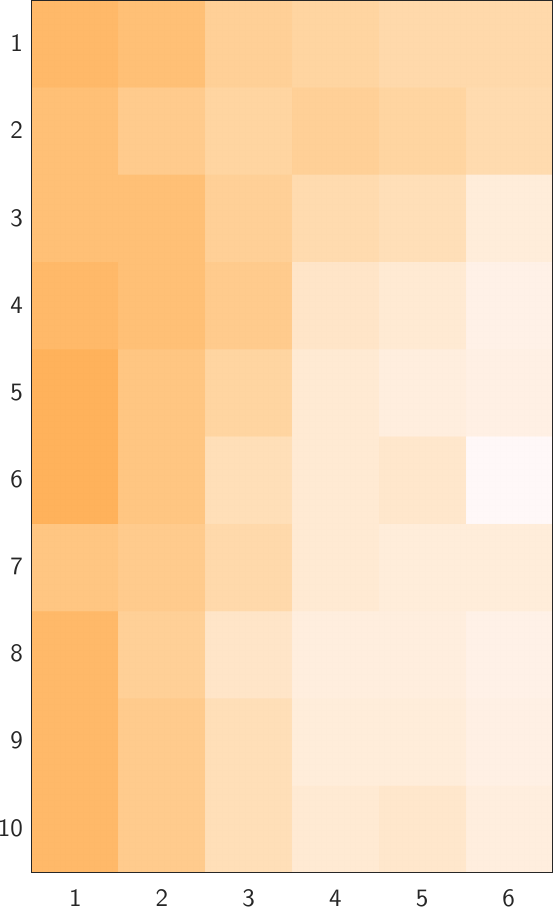}&
    \includegraphics[width=.09\linewidth]{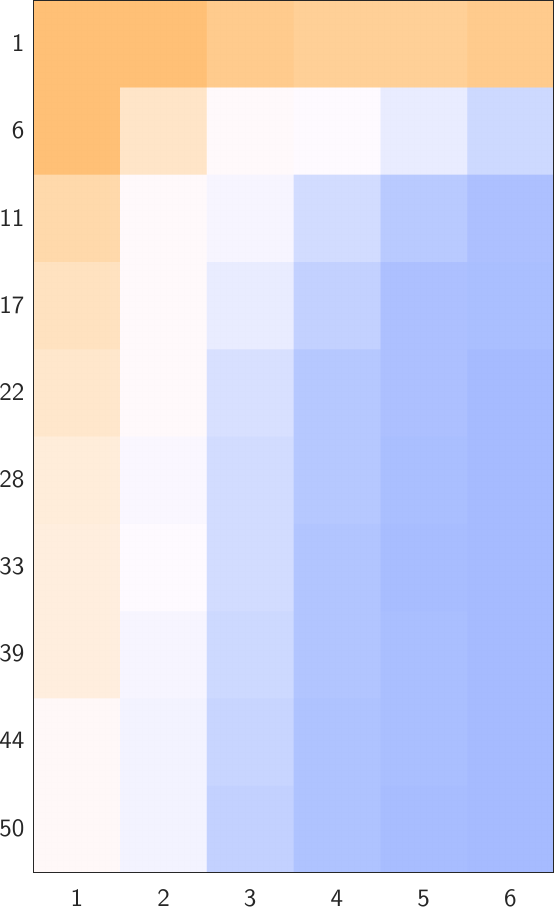}&
    \includegraphics[width=.09\linewidth]{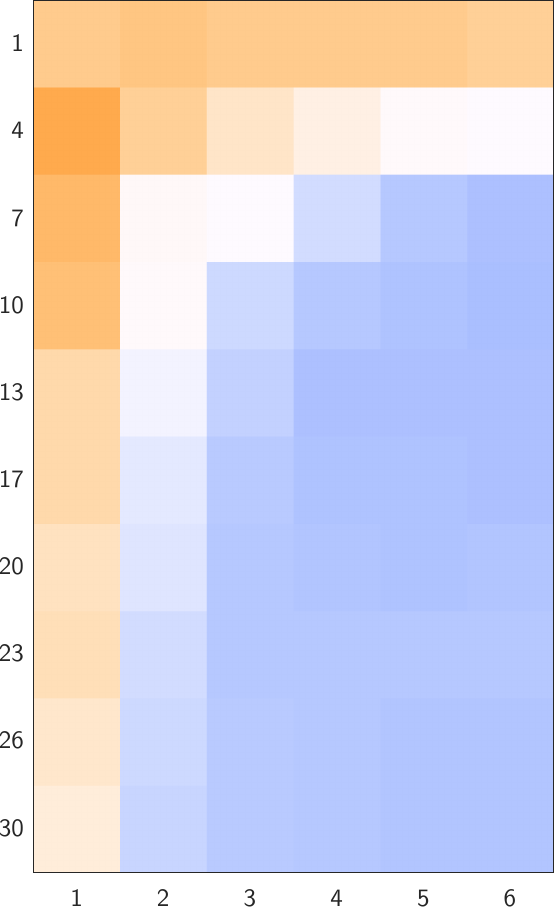}&
    \includegraphics[width=.09\linewidth]{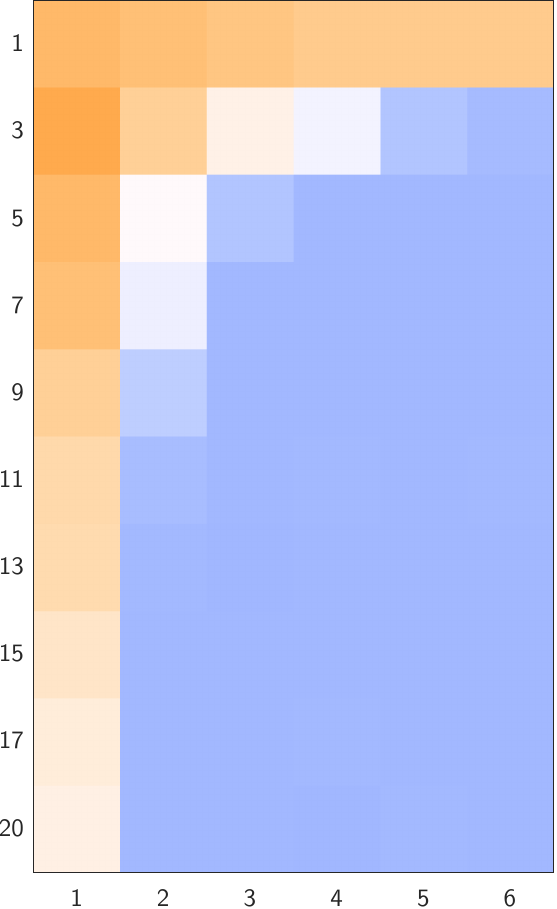}&
    \includegraphics[width=.09\linewidth]{figChebyshev/chebyshev-spectrumLowerFineFrequencies-100-signed-reconstructFromCoefs-chebyshev-heatmapWtNumLayers-10x6.pdf}\\
    \includegraphics[width=.09\linewidth]{figUntrained/mlpGatedRelu-biasScale1.0-5x6.png}&
    \includegraphics[width=.09\linewidth]{figUntrained/mlpGatedGelu-biasScale1.0-5x6.png}&
    \includegraphics[width=.09\linewidth]{figUntrained/mlpGatedSwish-biasScale1.0-5x6.png}&
    \includegraphics[width=.09\linewidth]{figUntrained/mlpGatedSelu-biasScale1.0-5x6.png}&
    \includegraphics[width=.09\linewidth]{figUntrained/mlpGatedTanh-biasScale1.0-5x6.png}&
    \includegraphics[width=.09\linewidth]{figUntrained/mlpGatedGaussian-biasScale1.0-5x6.png}&
    \includegraphics[width=.09\linewidth]{figUntrained/mlpGatedSin-biasScale1.0-5x6.png}&
    \includegraphics[width=.09\linewidth]{figUntrained/reconstructFromCoefs-biasScale1.0-5x6.png}\\
    
  \end{tabularx}\vspace{-6pt}
  \caption{\label{figHeatmapsChebyshevAppendix}
  Same as Figure~\ref{figHeatmapsFourierAppendix} with the LZ measure instead of the Fourier-based one. Results are nearly identical.
  \vspace{-8pt}}
\end{figure*}

\clearpage
\begin{figure*}[h!]
  \centering
  \renewcommand{\tabcolsep}{0.93em}
  \renewcommand{\arraystretch}{1.1}
  \small
  \begin{tabularx}{\linewidth}{cccccccc}
    \scriptsize\textsf{ReLU} & \scriptsize\textsf{GELU} & \scriptsize\textsf{Swish} & \scriptsize\textsf{SELU} & \scriptsize\textsf{TanH} & \scriptsize\textsf{Gaussian} & \scriptsize\textsf{Sin} & \scriptsize\textsf{Unbiased}\\

    \includegraphics[width=.09\linewidth]{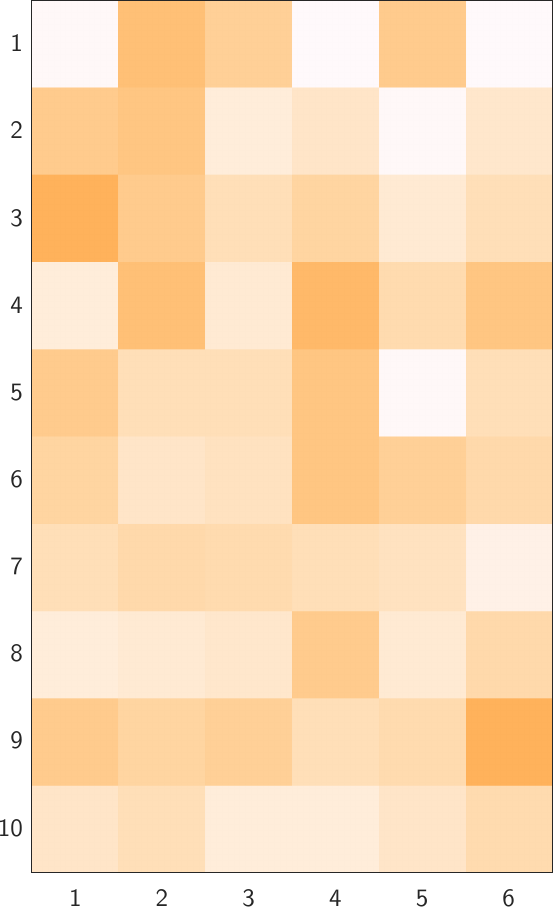}&
    \includegraphics[width=.09\linewidth]{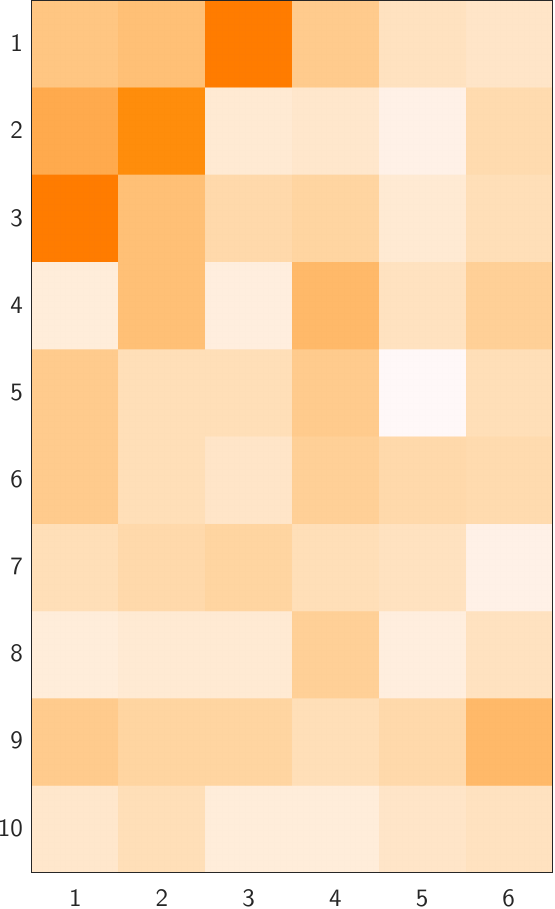}&
    \includegraphics[width=.09\linewidth]{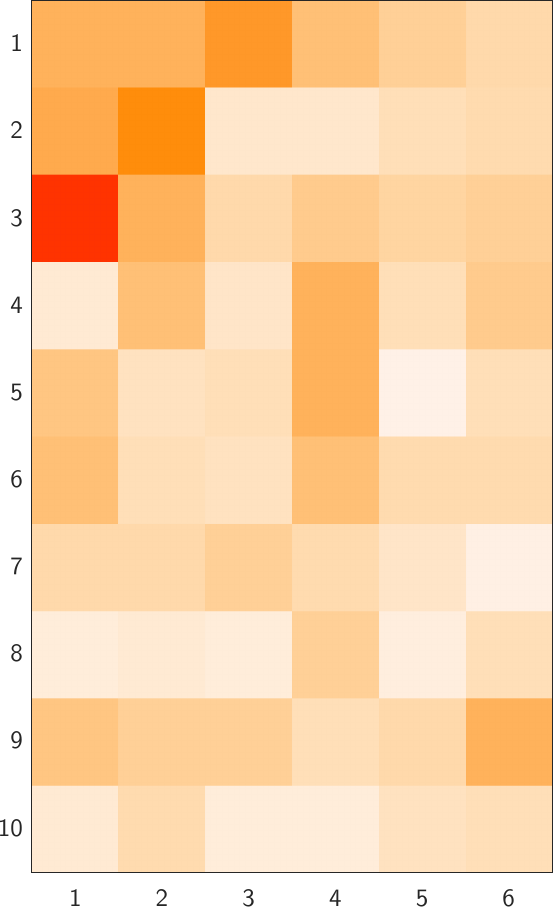}&
    \includegraphics[width=.09\linewidth]{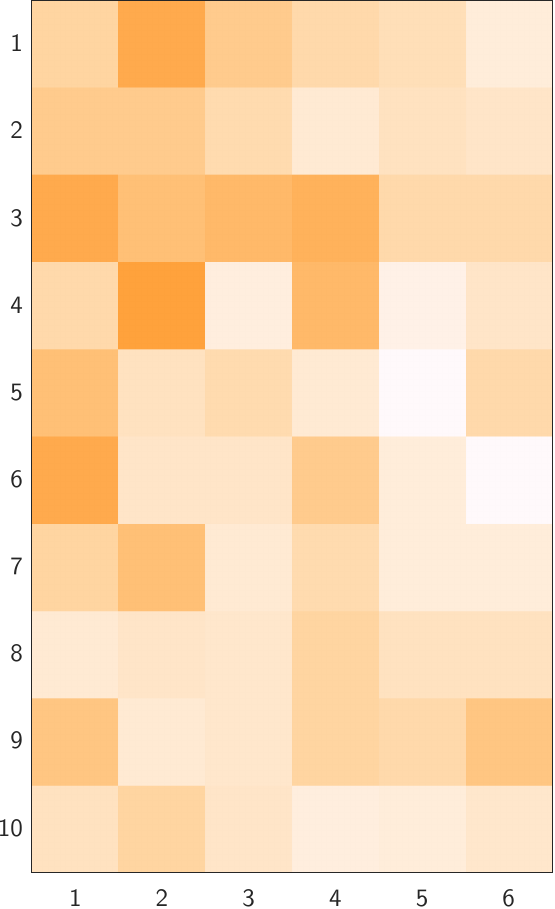}&
    \includegraphics[width=.09\linewidth]{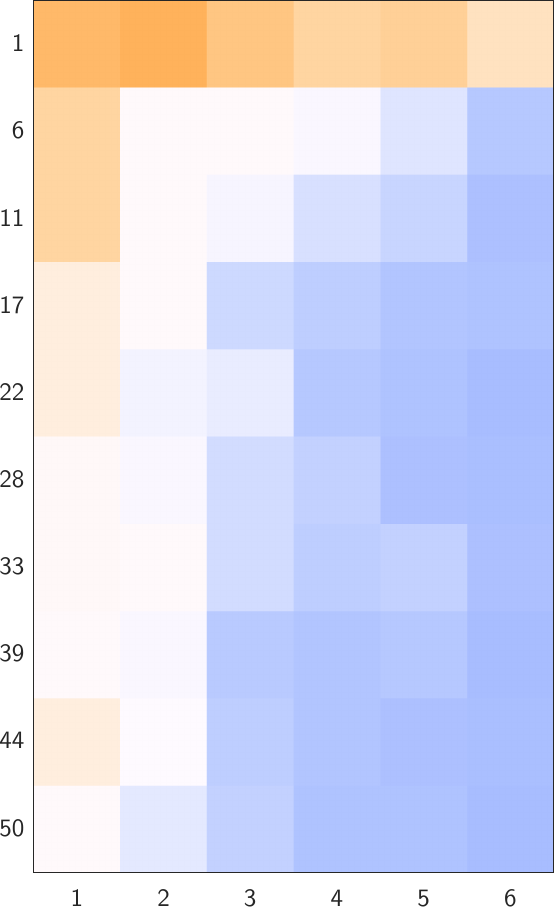}&
    \includegraphics[width=.09\linewidth]{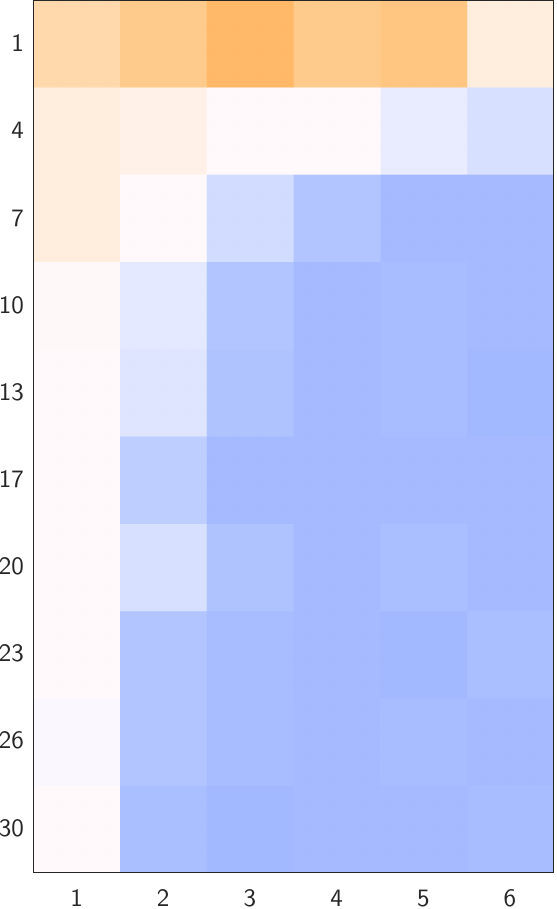}&
    \includegraphics[width=.09\linewidth]{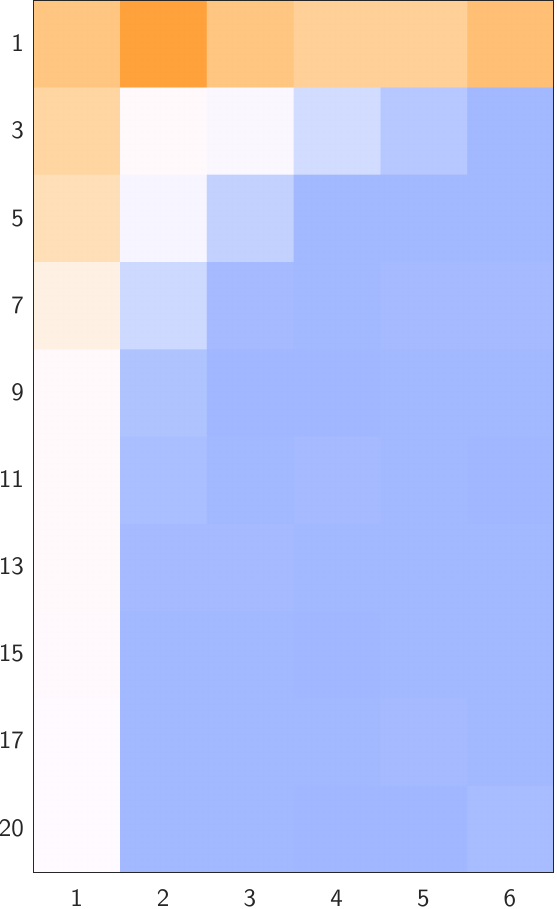}&
    \includegraphics[width=.09\linewidth]{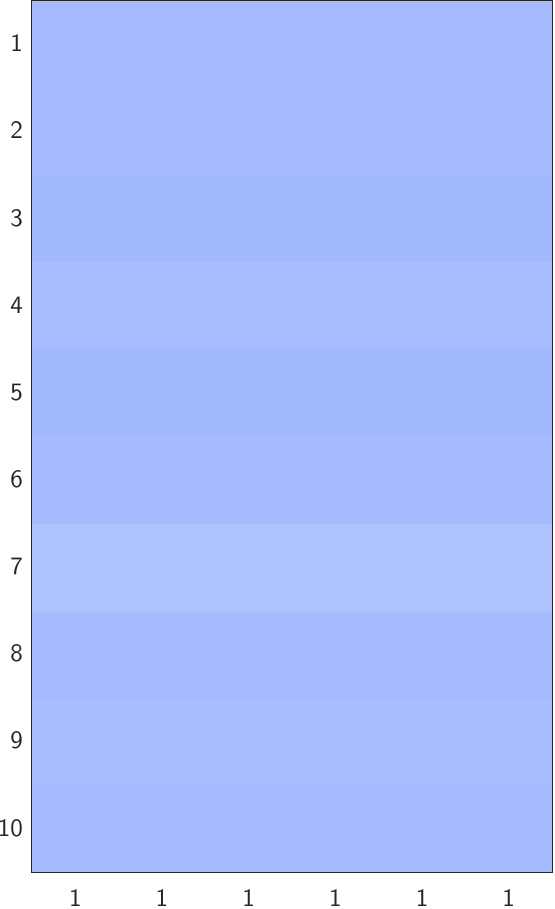}\\
    \includegraphics[width=.09\linewidth]{figUntrained/mlpRelu-biasScale1.0-5x6.png}&
    \includegraphics[width=.09\linewidth]{figUntrained/mlpGelu-biasScale1.0-5x6.png}&
    \includegraphics[width=.09\linewidth]{figUntrained/mlpSwish-biasScale1.0-5x6.png}&
    \includegraphics[width=.09\linewidth]{figUntrained/mlpSelu-biasScale1.0-5x6.png}&
    \includegraphics[width=.09\linewidth]{figUntrained/mlpTanh-biasScale1.0-5x6.png}&
    \includegraphics[width=.09\linewidth]{figUntrained/mlpGaussian-biasScale1.0-5x6.png}&
    \includegraphics[width=.09\linewidth]{figUntrained/mlpSin-biasScale1.0-5x6.png}&
    \includegraphics[width=.09\linewidth]{figUntrained/reconstructFromCoefs-biasScale1.0-5x6.png}\\

    \multicolumn{7}{l}{\textbf{With residual connections:}}\\

    \includegraphics[width=.09\linewidth]{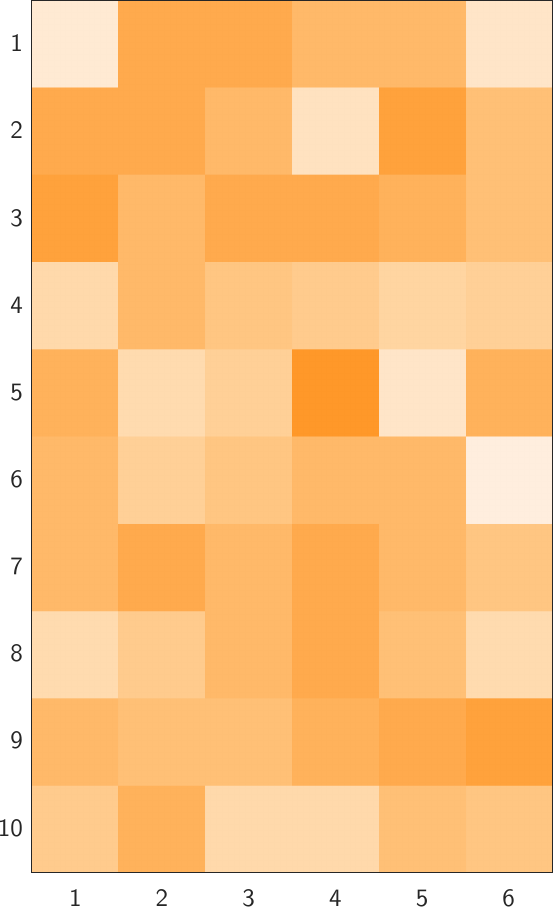}&
    \includegraphics[width=.09\linewidth]{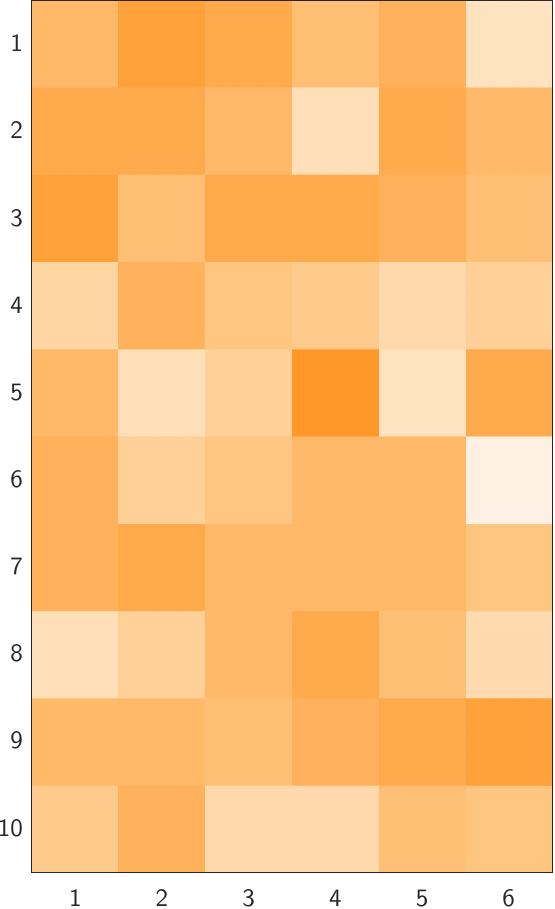}&
    \includegraphics[width=.09\linewidth]{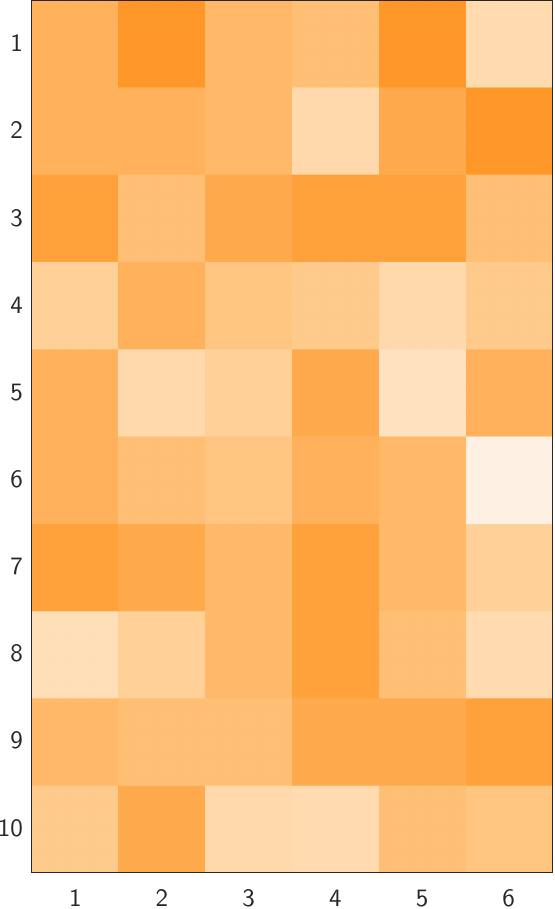}&
    \includegraphics[width=.09\linewidth]{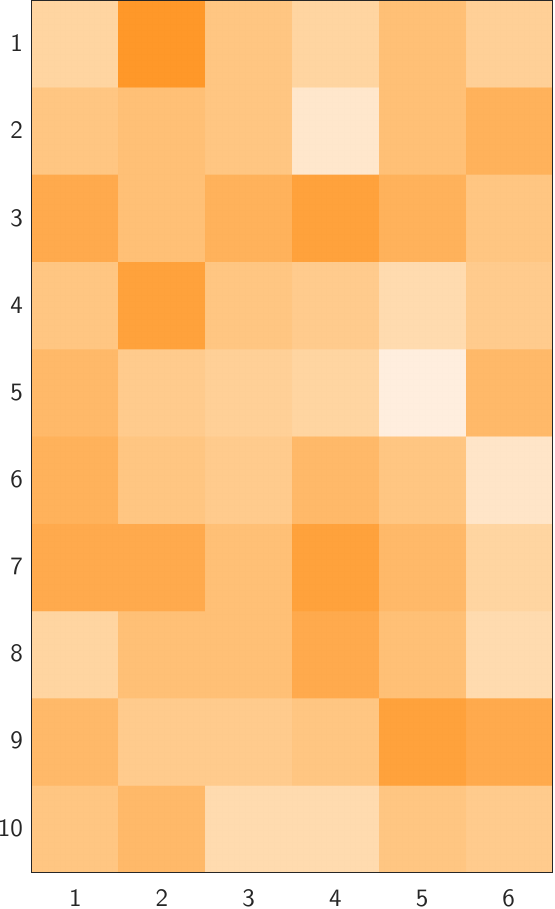}&
    \includegraphics[width=.09\linewidth]{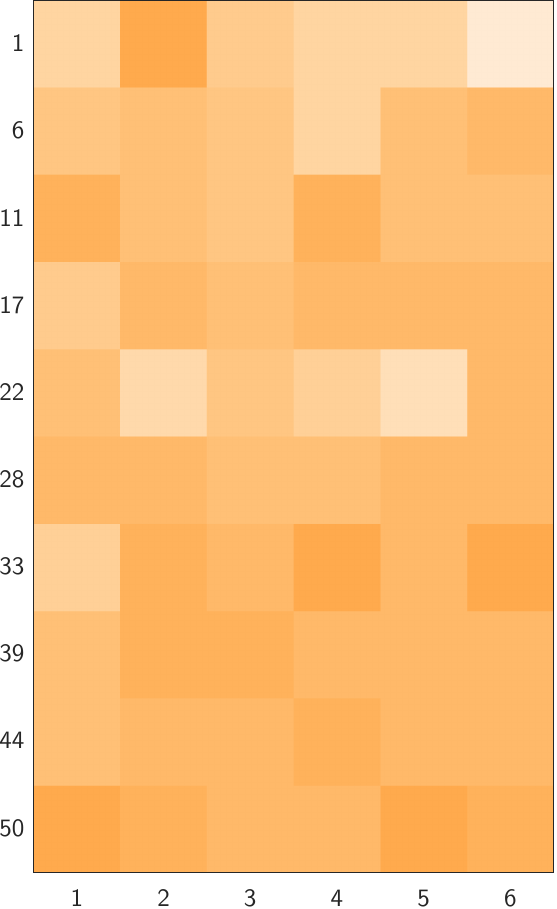}&
    \includegraphics[width=.09\linewidth]{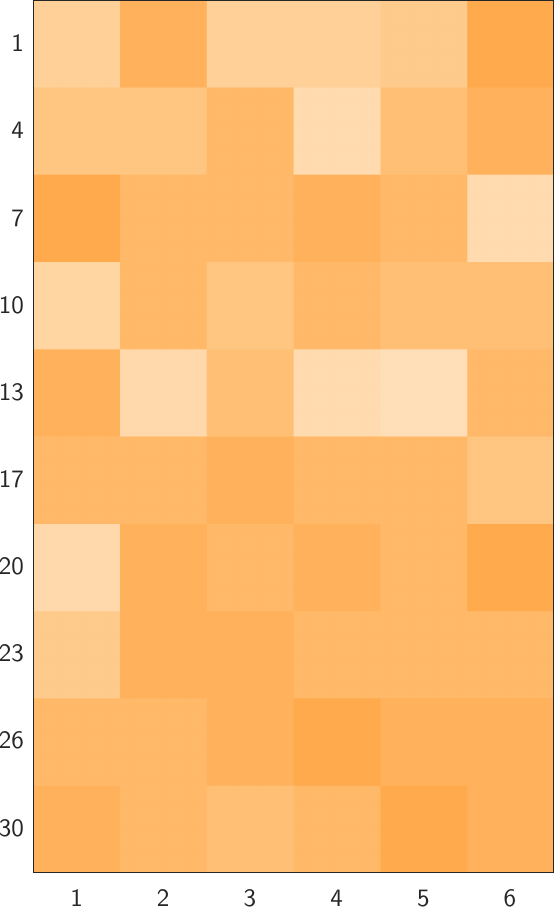}&
    \includegraphics[width=.09\linewidth]{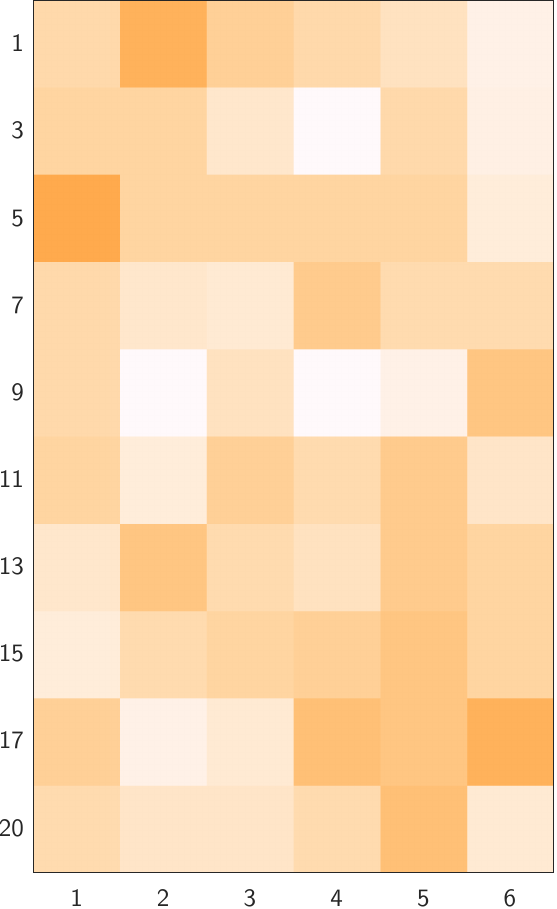}&
    \includegraphics[width=.09\linewidth]{figLz/fourier-spectrumIntegerFrequencies-lz-reconstructFromCoefs-fourier-heatmapWtNumLayers-10x6.pdf}\\
    \includegraphics[width=.09\linewidth]{figUntrained/mlpResRelu-biasScale1.0-5x6.png}&
    \includegraphics[width=.09\linewidth]{figUntrained/mlpResGelu-biasScale1.0-5x6.png}&
    \includegraphics[width=.09\linewidth]{figUntrained/mlpResSwish-biasScale1.0-5x6.png}&
    \includegraphics[width=.09\linewidth]{figUntrained/mlpResSelu-biasScale1.0-5x6.png}&
    \includegraphics[width=.09\linewidth]{figUntrained/mlpResTanh-biasScale1.0-5x6.png}&
    \includegraphics[width=.09\linewidth]{figUntrained/mlpResGaussian-biasScale1.0-5x6.png}&
    \includegraphics[width=.09\linewidth]{figUntrained/mlpResSin-biasScale1.0-5x6.png}&
    \includegraphics[width=.09\linewidth]{figUntrained/reconstructFromCoefs-biasScale1.0-5x6.png}\\

    \multicolumn{7}{l}{\textbf{With layer normalization:}}\\

    \includegraphics[width=.09\linewidth]{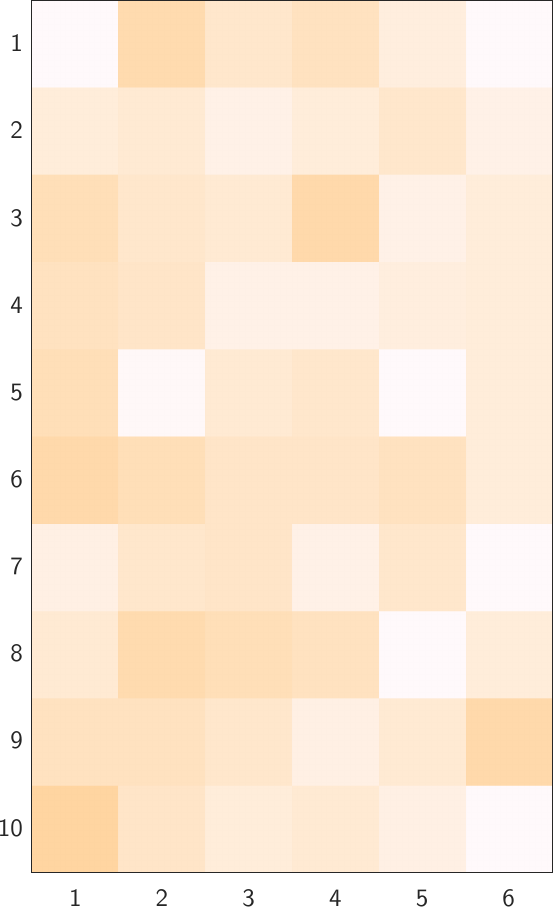}&
    \includegraphics[width=.09\linewidth]{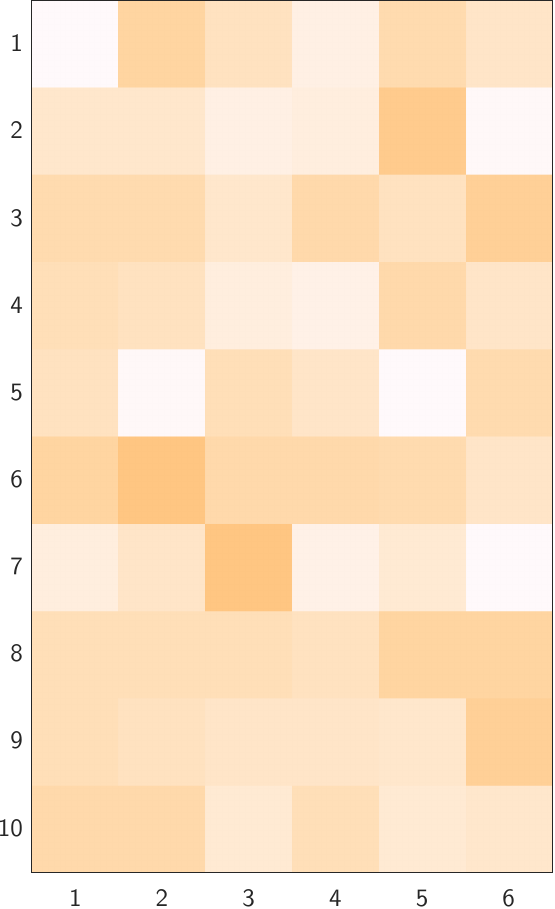}&
    \includegraphics[width=.09\linewidth]{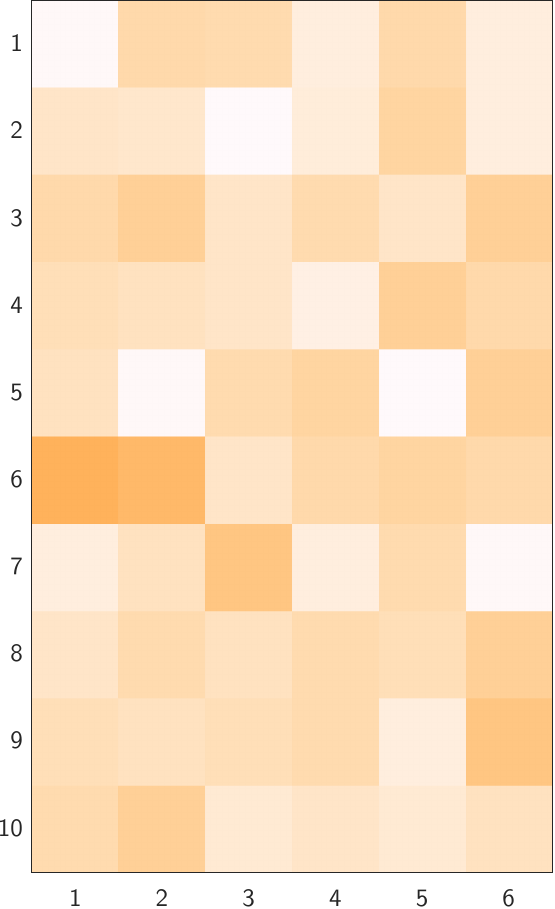}&
    \includegraphics[width=.09\linewidth]{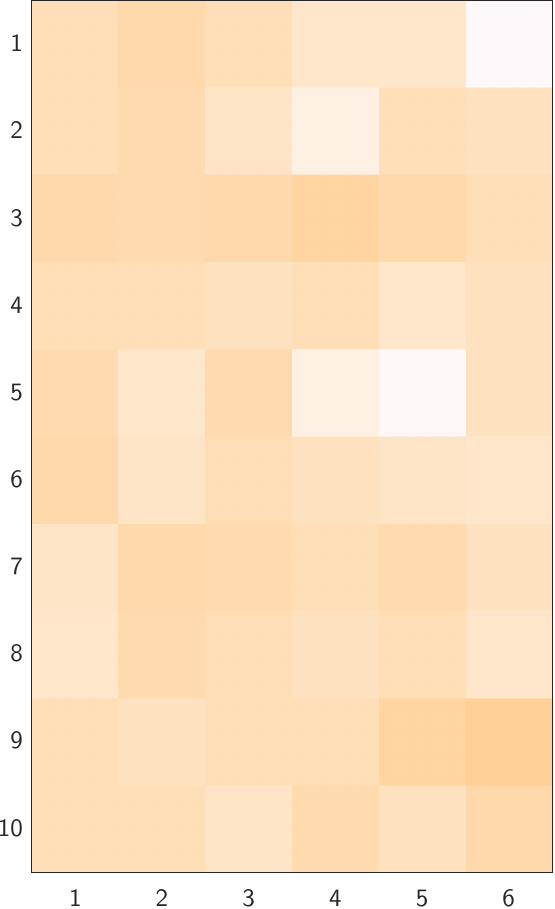}&
    \includegraphics[width=.09\linewidth]{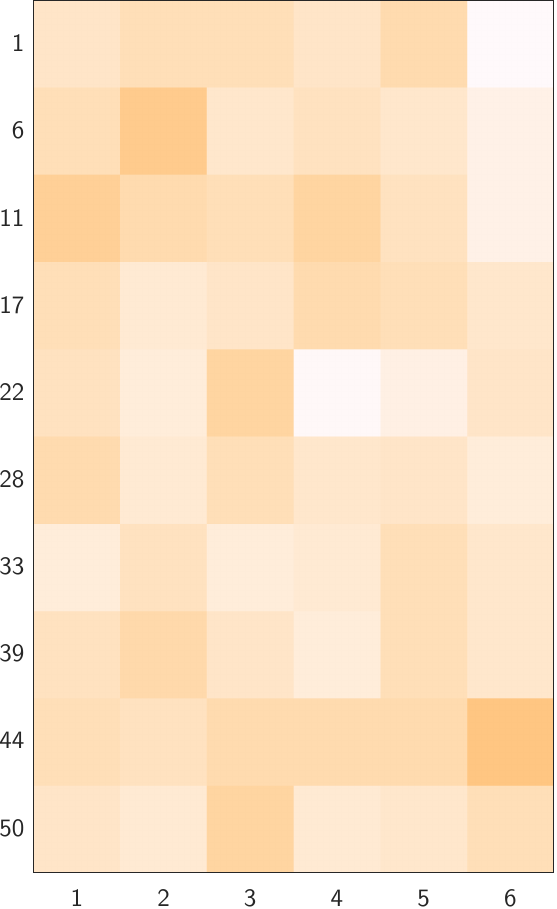}&
    \includegraphics[width=.09\linewidth]{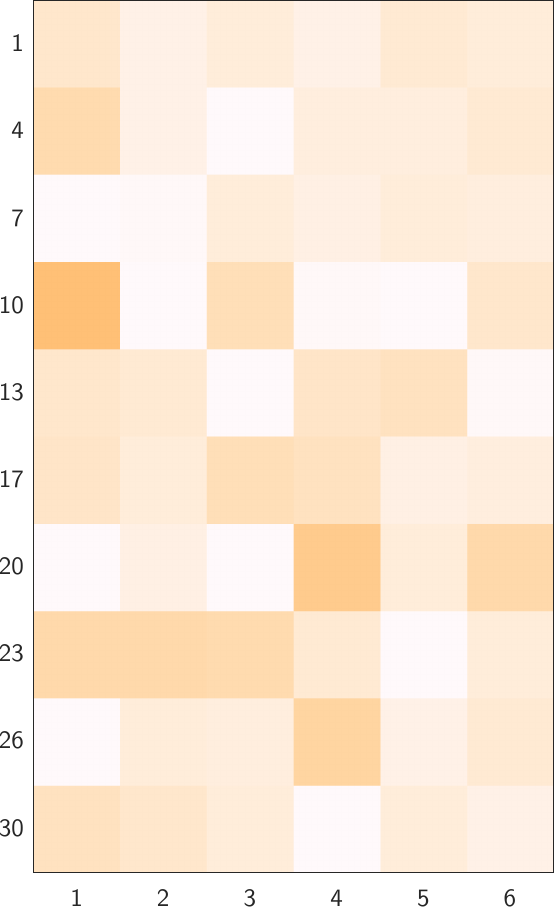}&
    \includegraphics[width=.09\linewidth]{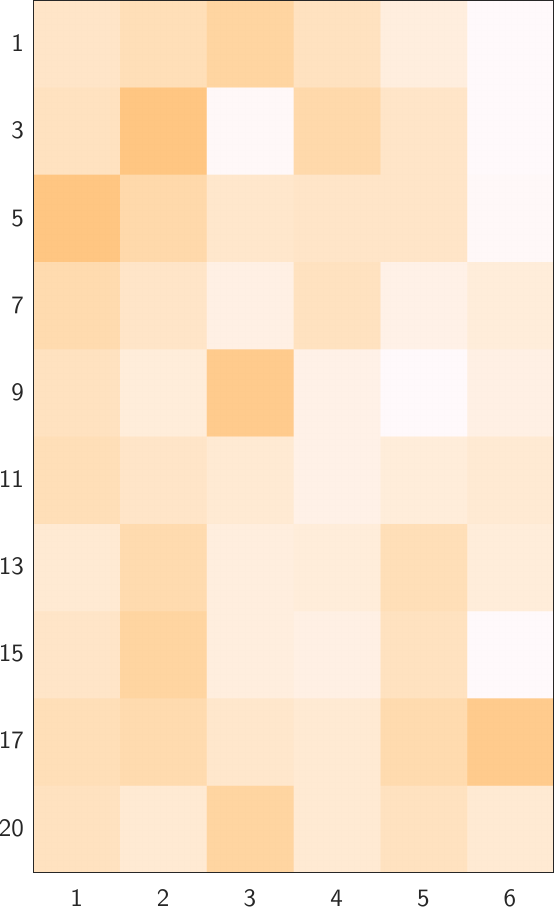}&
    \includegraphics[width=.09\linewidth]{figLz/fourier-spectrumIntegerFrequencies-lz-reconstructFromCoefs-fourier-heatmapWtNumLayers-10x6.pdf}\\
    \includegraphics[width=.09\linewidth]{figUntrained/mlpPostNormRelu-biasScale1.0-5x6.png}&
    \includegraphics[width=.09\linewidth]{figUntrained/mlpPostNormGelu-biasScale1.0-5x6.png}&
    \includegraphics[width=.09\linewidth]{figUntrained/mlpPostNormSwish-biasScale1.0-5x6.png}&
    \includegraphics[width=.09\linewidth]{figUntrained/mlpPostNormSelu-biasScale1.0-5x6.png}&
    \includegraphics[width=.09\linewidth]{figUntrained/mlpPostNormTanh-biasScale1.0-5x6.png}&
    \includegraphics[width=.09\linewidth]{figUntrained/mlpPostNormGaussian-biasScale1.0-5x6.png}&
    \includegraphics[width=.09\linewidth]{figUntrained/mlpPostNormSin-biasScale1.0-5x6.png}&
    \includegraphics[width=.09\linewidth]{figUntrained/reconstructFromCoefs-biasScale1.0-5x6.png}\\

    \multicolumn{7}{l}{\textbf{With gating:}}\\

    \includegraphics[width=.09\linewidth]{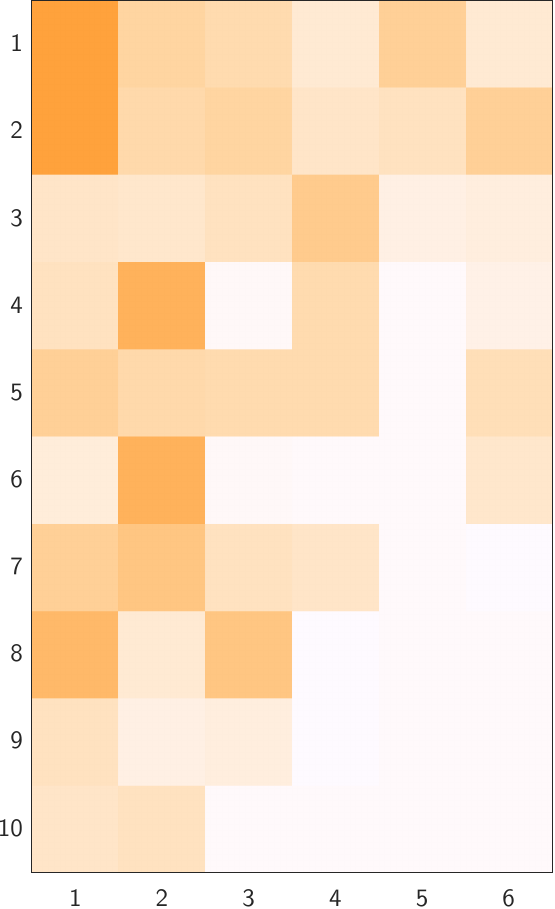}&
    \includegraphics[width=.09\linewidth]{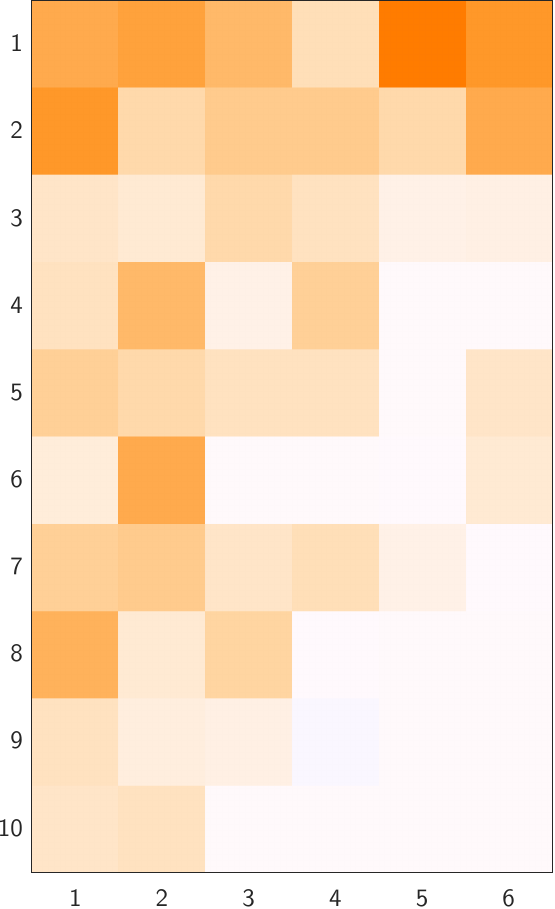}&
    \includegraphics[width=.09\linewidth]{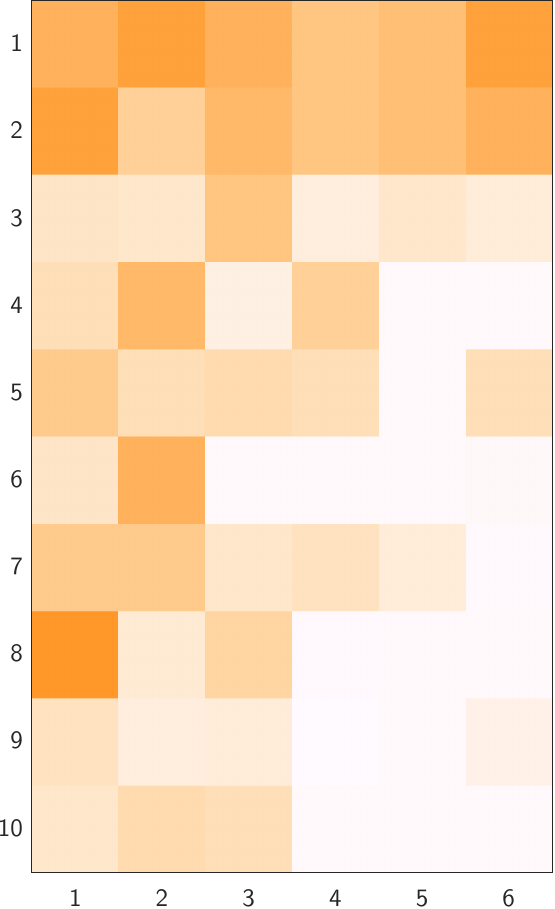}&
    \includegraphics[width=.09\linewidth]{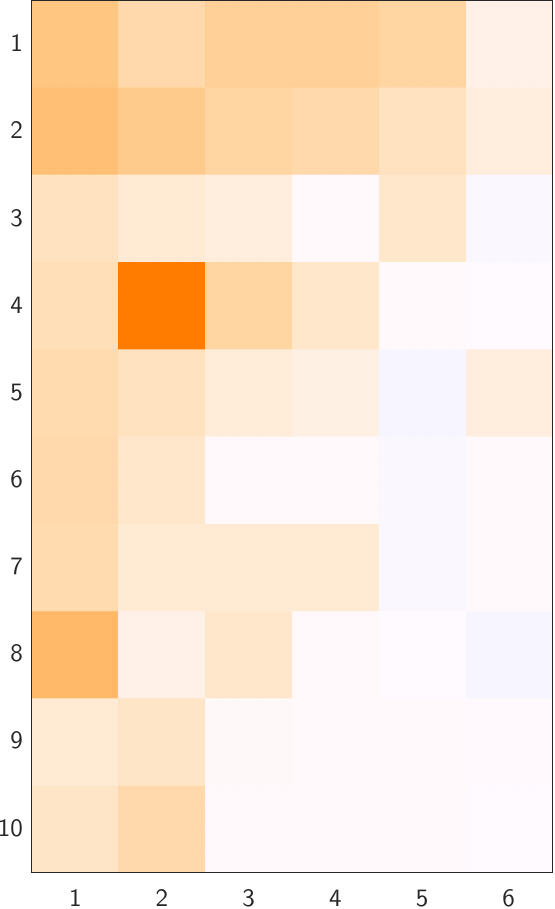}&
    \includegraphics[width=.09\linewidth]{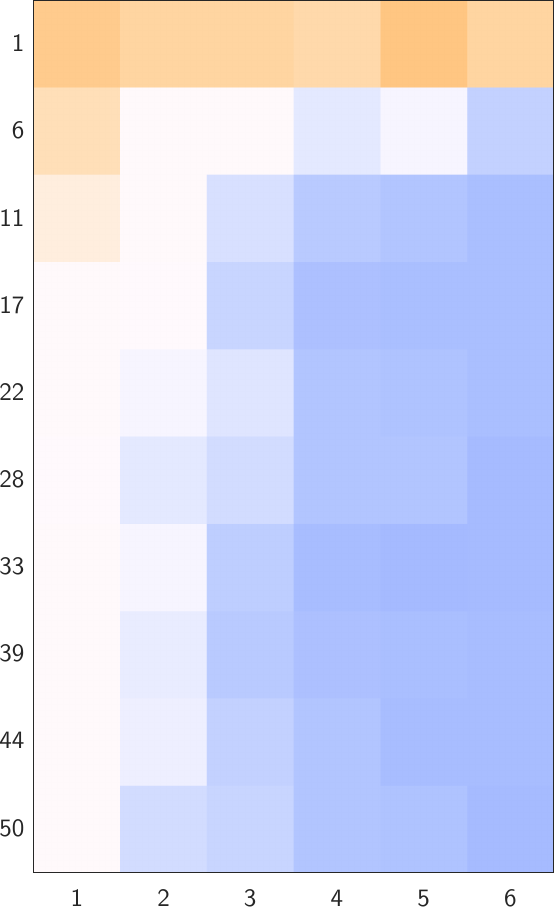}&
    \includegraphics[width=.09\linewidth]{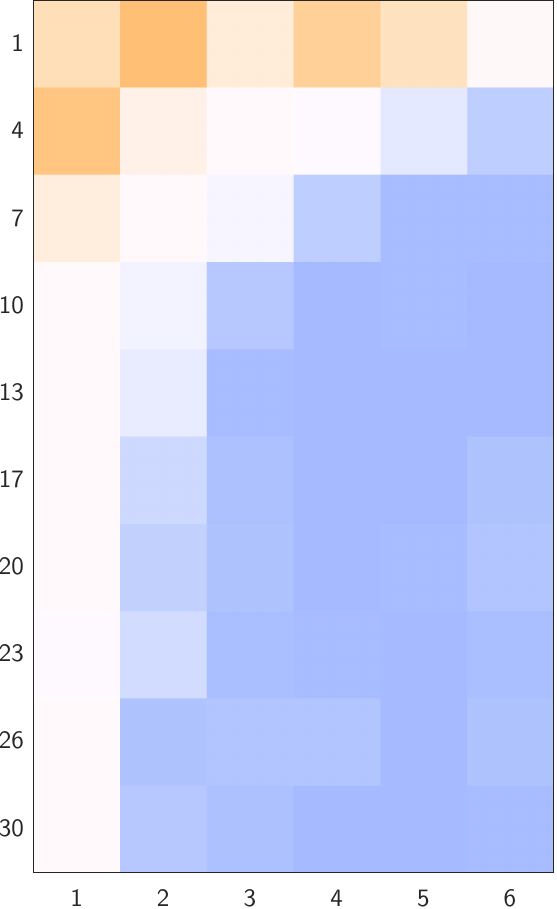}&
    \includegraphics[width=.09\linewidth]{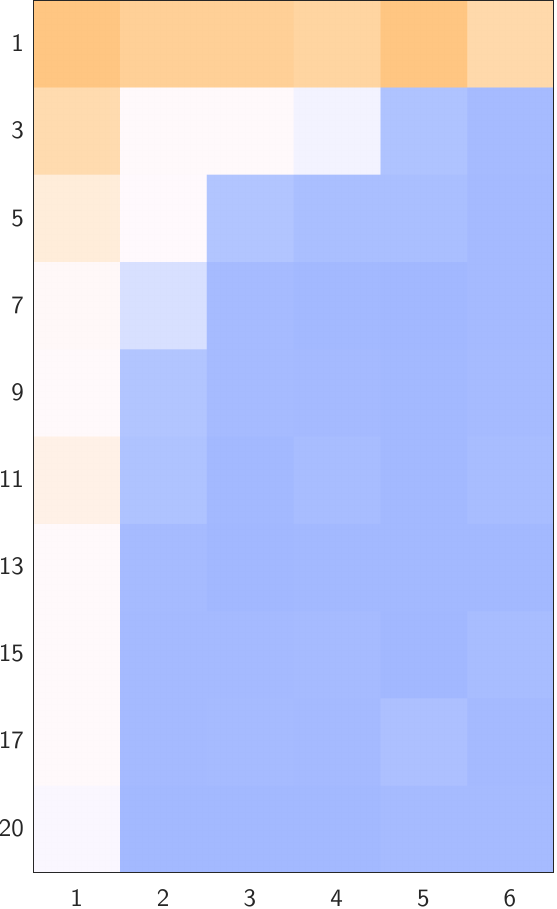}&
    \includegraphics[width=.09\linewidth]{figLz/fourier-spectrumIntegerFrequencies-lz-reconstructFromCoefs-fourier-heatmapWtNumLayers-10x6.pdf}\\
    \includegraphics[width=.09\linewidth]{figUntrained/mlpGatedRelu-biasScale1.0-5x6.png}&
    \includegraphics[width=.09\linewidth]{figUntrained/mlpGatedGelu-biasScale1.0-5x6.png}&
    \includegraphics[width=.09\linewidth]{figUntrained/mlpGatedSwish-biasScale1.0-5x6.png}&
    \includegraphics[width=.09\linewidth]{figUntrained/mlpGatedSelu-biasScale1.0-5x6.png}&
    \includegraphics[width=.09\linewidth]{figUntrained/mlpGatedTanh-biasScale1.0-5x6.png}&
    \includegraphics[width=.09\linewidth]{figUntrained/mlpGatedGaussian-biasScale1.0-5x6.png}&
    \includegraphics[width=.09\linewidth]{figUntrained/mlpGatedSin-biasScale1.0-5x6.png}&
    \includegraphics[width=.09\linewidth]{figUntrained/reconstructFromCoefs-biasScale1.0-5x6.png}\\
    
  \end{tabularx}\vspace{-6pt}
  \caption{\label{figHeatmapsLzAppendix}
  Same as Figure~\ref{figHeatmapsFourierAppendix} with the LZ measure instead of the Fourier-based one. Results are nearly identical but noisier.
  \vspace{-8pt}}
\end{figure*}

\end{document}